\documentclass[10pt,twocolumn,letterpaper]{article}

\usepackage[pagenumbers]{cvpr} 

\usepackage{graphicx}
\usepackage{amsmath}
\usepackage{amssymb}
\usepackage{booktabs}

\usepackage{multirow}
\usepackage{tabularx}
\usepackage[dvipsnames,table,xcdraw]{xcolor}
\usepackage{colortbl}
\usepackage{bbm}
\usepackage{lipsum}
\usepackage{makecell}
\usepackage[final,tracking=true,kerning=true,spacing=true,factor=1000,stretch=10,shrink=10]{microtype}

\newcommand{\ours}{SemiVL}
\newcommand*{\y}{\checkmark}
\newcommand*{\n}{\textcolor{gray}{--}}
\newcommand*{\z}{\phantom{0}}
\newcommand*{\blarrow}{\rotatebox[origin=c]{270}{$\Rsh$}}

\newcommand{\green}[1]{\textcolor{ForestGreen}{#1}}
\newcommand{\venue}[1]{{\tiny\textcolor{gray}{[#1]}}}
\newcommand{\twodots}{\mathinner {\ldotp \ldotp}}

\newcolumntype{Y}{>{\centering\arraybackslash}X}

\definecolor{cvprblue}{rgb}{0.21,0.49,0.74}
\usepackage[pagebackref,breaklinks,colorlinks,citecolor=cvprblue]{hyperref}

\def\paperID{4813} 
\def\confName{CVPR}
\def\confYear{2024}

\title{SemiVL: Semi-Supervised Semantic Segmentation with\\ Vision-Language Guidance}

\author{
   Lukas Hoyer\,\textsuperscript{1,2*} \quad
   David Joseph Tan\,\textsuperscript{2} \quad
   Muhammad Ferjad Naeem\,\textsuperscript{1} \\
   Luc Van Gool\,\textsuperscript{1,4,5} \quad
   Federico Tombari\,\textsuperscript{2,3} \\
   \textsuperscript{1}\,ETH Zurich \enskip
   \textsuperscript{2}\,Google \enskip
   \textsuperscript{3}\,TU Munich \enskip
   \textsuperscript{4}\,KU Leuven \enskip
   \textsuperscript{5}\,INSAIT Sofia\\
  {\tt\small \{lhoyer,mnaeem,vangool\}@vision.ee.ethz.ch, \{djtan,tombari\}@google.de}
}

\begin{document}
\maketitle
\begin{abstract}
In semi-supervised semantic segmentation, a model is trained with a limited number of labeled images along with a large corpus of unlabeled images to reduce the high annotation effort. While previous methods are able to learn good segmentation boundaries, they are prone to confuse classes with similar visual appearance due to the limited supervision. On the other hand, vision-language models (VLMs) are able to learn diverse semantic knowledge from image-caption datasets but produce noisy segmentation due to the image-level training. In SemiVL, we propose to integrate rich priors from VLM pre-training into semi-supervised semantic segmentation to learn better semantic decision boundaries. To adapt the VLM from global to local reasoning, we introduce a spatial fine-tuning strategy for label-efficient learning. Further, we design a language-guided decoder to jointly reason over vision and language. Finally, we propose to handle inherent ambiguities in class labels by providing the model with language guidance in the form of class definitions. We evaluate SemiVL on 4 semantic segmentation datasets, where it significantly outperforms previous semi-supervised methods. For instance, SemiVL improves the state-of-the-art by +13.5 mIoU on COCO with 232 annotated images and by +6.1 mIoU on Pascal VOC with 92 labels. Project page: \href{https://github.com/google-research/semivl}{github.com/google-research/semivl}
\end{abstract}

\vspace{-2mm}
{\let\thefootnote\relax\footnote{{$^*$This work was conducted during an internship at Google.}}} 
\vspace{-2mm}
\section{Introduction}

Semantic segmentation models predict pixel-level dense semantic labels for an image. 
They have important applications in many areas such as autonomous driving, augmented reality, robotics, medical imaging, and remote sensing.
However, training such models requires very costly pixel-wise human annotations over a large dataset.
To reduce the dependence on large labeled datasets, semi-supervised semantic segmentation aims to effectively learn from a small portion of labeled images while additionally leveraging a large set of unlabeled images.
Typical strategies include adversarial training~\cite{souly2017semi,mittal2019semi} and self-training~\cite{yang2022st++,sohn2020fixmatch,yang2023revisiting}.

\begin{figure}
    \footnotesize
    \centering
    \setlength{\tabcolsep}{0pt}
    \begin{tabularx}{\linewidth}{*{5}{Y}}
    (a) Image & (b)~G.Truth & (c)~UniMatch & (d) CLIP & \textbf{(e)~\ours\ (Ours)} \\
    \end{tabularx}
    \includegraphics[width=\linewidth]{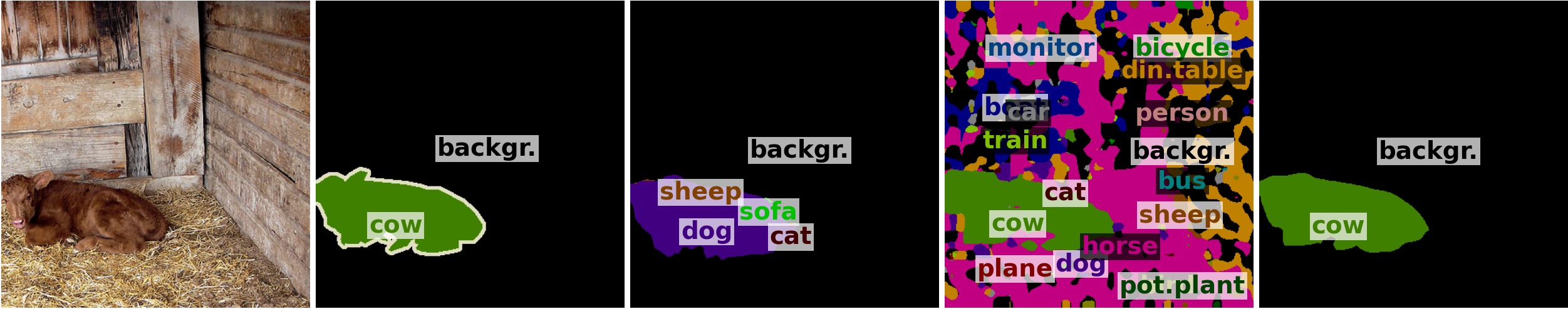}
    \includegraphics[width=\linewidth]{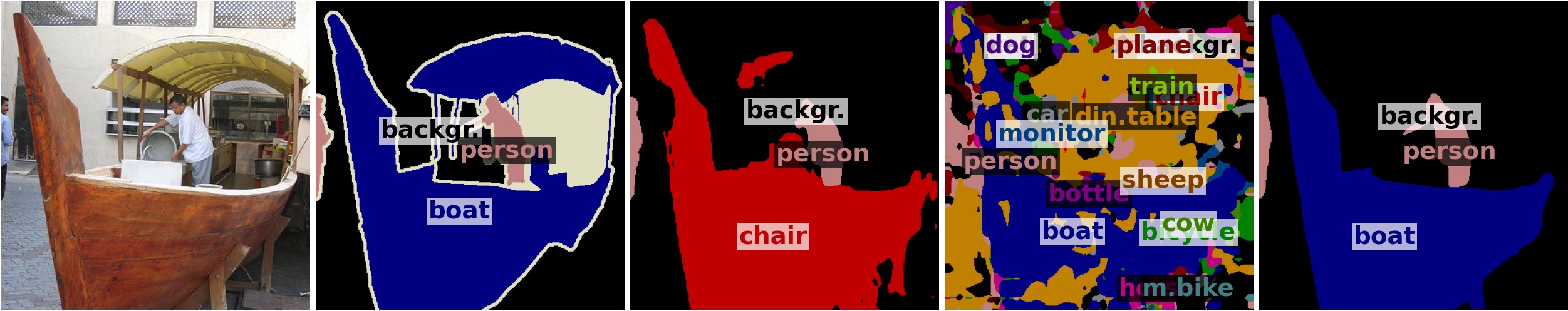}
    \caption{While previous SOTA methods such as UniMatch~\cite{yang2023revisiting} achieve good segmentation even with only 92 labeled images, they struggle to distinguish classes with similar appearance.
    In contrast, vision-language training such as CLIP~\cite{radford2021learning} learns rich semantic representations but suffers from noisy segmentation (obtained with~\cite{zhou2022extract}). Our \ours\ supplements semi-supervised training with vision-language guidance, combining good segmentation with a rich semantic understanding.}
    \label{fig:motivation}
\end{figure}

While state-of-the-art methods for semi-supervised semantic segmentation such as UniMatch~\cite{yang2023revisiting} are able to learn good segmentation masks, segments with similar visual features are prone to misclassification, due to the lack of sufficient labeled examples for learning the exact semantic decision boundaries for these classes.
This is particularly problematic for rare instances such as a calve or a boat on land as shown in Fig.~\ref{fig:motivation}c. Here, the cow (top, green) is confused with a dog (purple) and the boat (bottom, blue) with a chair (red) even though the segmentation is mostly correct.

To better capture semantics, we propose to supplement semi-supervised semantic segmentation with guidance from Vision Language Models~(VLM). 
VLMs such as CLIP~\cite{radford2021learning} are trained on a web-scale image-caption dataset. The diversity of the data and the natural language captions (instead of a fixed set of classes) enables VLMs to capture richer semantic representations.
For example, they can correctly identify the cow and the boat in Fig.~\ref{fig:motivation}d. However, as they are trained on image level, their features do not localize well and their dense predictions~\cite{maskclip} are noisy as shown in Fig.~\ref{fig:motivation}d. 

In this work, we study how to combine the good localization of semi-supervised training and the rich semantic understanding of VLMs. Based on the findings, we propose \ours, which combines both strengths to achieve a good segmentation quality and a fine-grained semantic discriminability. For example, \ours\ correctly segments and classifies the cow and boat in Fig.~\ref{fig:motivation}e. To the best of our knowledge, \ours\ is the first work to utilize vision-language guidance for semi-supervised semantic segmentation to mitigate the issue of limited dense labels.
Previous works in VLM semantic segmentation (see Sec.~\ref{ref:related_clip_segmentation}) either operate without dense labels~\cite{xu2022groupvit,zhou2022extract,cha2023learning}, which significantly limits their performance, or use large-scale annotated segmentation datasets~\cite{ding2022decoupling,zhou2023zegclip,xu2023side}, which are expensive to obtain. Instead, \ours\ can learn high-quality semantic segmentation in a semi-supervised setting with only a few labels.

Building on consistency regularization~\cite{yang2023revisiting} for semi-supervised training,
\ours\ introduces five components
to transfer the power of vision-language modeling to semantic segmentation under the constraint of limited segmentation annotations,
which are highlighted in green color in Fig.~\ref{fig:overview}:

\noindent\textbf{(1) Vision-Language Pre-Training (Sec.~\ref{sec:methods_vl_pretraining}):}
To utilize the rich semantic prior of VLMs, we initialize the semi-supervised training with vision-language pre-training. 
Compared to the previously used ImageNet pre-training, the VL pre-training on web-scale image-caption pairs provides more diversity and is not limited to a fixed set of classes.

\noindent\textbf{(2) Spatial Fine-Tuning (Sec.~\ref{sec:methods_spatial_finetuning}):}
As the VL pre-training is conducted on image level, the vision model needs to be fine-tuned to achieve good feature localization for semantic segmentation. Due to the limited annotations, the fine-tuning is prone to overfitting and forgetting the rich semantics from the pre-training. Therefore, we introduce parameter-efficient spatial fine-tuning.
It only fine-tunes network layers that model interactions between different pixels to improve the localization of features while it freezes layers that operate locally to preserve their semantic reasoning capabilities.

\noindent\textbf{(3) Language-Guided Decoder (Sec.~\ref{sec:methods_language_guided_decoder}):}
To exploit the alignment of vision and language embeddings from VL pre-training, we integrate it into the segmentation decoder.
It processes VL similarity maps using decoupled spatial and semantic reasoning. By sharing parameters across classes or pixels, the limited labels are used effectively.

\noindent\textbf{(4) Dense CLIP Guidance (Sec.~\ref{sec:methods_clip_guidance}):}
When fine-tuning the VLM towards segmentation, some of the original prior is corrupted and the network drifts to wrong predictions on the unlabeled images.
To anchor the semi-supervised training on unlabeled images, we regularize it with the predictions from a \emph{frozen} VLM, which cannot drift. As these predictions are noisy, we only use the ones with a high certainty.

\noindent\textbf{(5) Class Definitions (Sec.~\ref{sec:methods_class_definitions}):}
Depending on the use case, certain concepts can fall into different classes. For example, a person pushing a bicycle could belong to the class pedestrian or rider.
However, a small number of annotations might not be sufficient to learn all relevant dataset-specific decision boundaries between classes. 
Therefore, we utilize the novel capability of \ours\ to provide language guidance in the form of class definitions to the model. These are often part of the annotation guidelines of datasets or can be created with much less effort than mining images of relevant corner cases and labeling them with semantic segmentation.

\begin{figure}
    \centering
    \includegraphics[width=\linewidth]{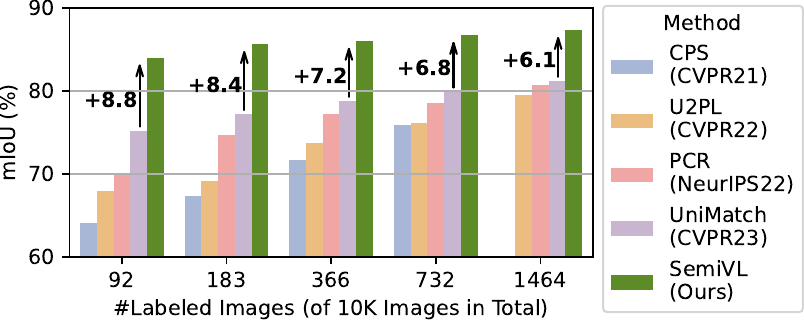}
    \caption{Semi-supervised semantic segmentation on Pascal VOC. Utilizing vision-language guidance, our \ours\ achieves major gains over previous methods, particularly for fewer labels.}
    \label{fig:sota_voc}
\end{figure}

We evaluate \ours\ on 4 common semantic segmentation datasets, where it significantly outperforms previous semi-supervised methods. For instance, it improves the state of the art by +13.5 mIoU on COCO with 232 annotated images and by +9.7 mIoU on ADE20K with 158 labels. In particular, \ours\ can maintain the performance of previous methods using \textit{only a quarter of the segmentation annotations}. The significance of the improvements is visualized in Fig.~\ref{fig:sota_voc}, showing that \ours\ achieves major gains over previous works for all labeled subsets. In particular, \ours\ improves the performance for few labels, showing its effectiveness for label-efficient learning. 
In a comprehensive ablation study on Pascal VOC, we show that all 5 components of \ours\ significantly improve the semi-supervised performance. The gains are particularly large when only a few labels are available, demonstrating \ours's effectiveness to compensate for limited annotations.

\section{Related Work}

\subsection{Semi-Supervised Semantic Segmentation}

Early works~\cite{souly2017semi,mittal2019semi} in semi-supervised segmentation use unlabeled images in a GAN framework~\cite{goodfellow2014generative} to match their predictions with the distribution of manual labels. Self-training~\cite{grandvalet2004semi, lee2013pseudo} methods generate pseudo-labels for unlabeled images and use them for iterative re-training. To mitigate pseudo-label drift and self-confirmation bias~\cite{arazo2020pseudo}, strategies such as confidence-weighting~\cite{feng2022dmt, yang2022st++},  curricula~\cite{zhang2021flexmatch, xu2021dash, ma2023enhanced}, class balancing~\cite{he2021re, hu2021semi, guan2022unbiased, hoyer2022daformer}, auxiliary self-supervised tasks~\cite{hoyer2021three,hoyer2023improving}, contrastive learning~\cite{wang2022semi}, 
symbolic reasoning~\cite{liang2023logic}, 
or soft pseudo-labels~\cite{ma2023enhanced} can be used. Another popular strategy is consistency regularization~\cite{bachman2014learning,sajjadi2016regularization}, which makes predictions invariant to perturbations such as data augmentation~\cite{sohn2020fixmatch,yang2022st++,yang2023revisiting}, mixed samples~\cite{french2019semi,hu2021semi,olsson2021classmix}, overlapping crops with different context~\cite{lai2021semi,hoyer2022hrda,hoyer2023domain}, different model initialization~\cite{chen2021semi}, masked images~\cite{hoyer2023mic}, or feature perturbations~\cite{yang2023revisiting}. Due to its success and simplicity, we base \ours\ on consistency training.

\subsection{Semantic Segmentation with VLMs}
\label{ref:related_clip_segmentation}

Vision Language Models (VLMs) such as CLIP and its variants~\cite{xclip, maskclip, xclip, naeem2023silc, naeem2023i2mvformer, naeem2022i2dformer} utilize a web-scale image-caption dataset to learn a joint embedding space between image and text.
Due to the semantically rich and diverse dataset,
these models possess great generalization to a wide range of vision tasks\cite{luo2023segclip, cho2023cat, gu2021open, khan2023introducing, wang2022clip, chen2023pali}. With prompt engineering, VLMs can be improved by ensembling prompts~\cite{radford2021learning} or generating class descriptions by LLMs~\cite{naeem2023i2mvformer, cupl}.
Recently, VLMs also gained attention in semantic segmentation.
For instance, several zero-shot methods~\cite{xu2022groupvit,zhou2022extract,shin2022reco,cha2023learning} aim to learn segmentation only from image-caption pairs without any dense annotations. MaskCLIP~\cite{zhou2022extract} shows that (noisy) segmentation emerges in the CLIP vision encoder. Further techniques include hierarchical grouping~\cite{xu2022groupvit}, retrieval and co-segmentation~\cite{shin2022reco}, or text-grounded masking~\cite{cha2023learning}.
However, the produced segmentations are noisy due to the lack of dense supervision. In contrast, open-vocabulary segmentation methods achieve better segmentation as they use a large labeled semantic segmentation dataset. Here, CLIP is often used to deal with unknown classes during test time. OpenSeg~\cite{ghiasi2022scaling} pools visual features with learned mask proposals. LSeg~\cite{li2022languagedriven} learns dense visual embeddings that align with CLIP text embeddings. ZegFormer~\cite{ding2022decoupling}, ZSseg~\cite{xu2022simple}, and OVSeg~\cite{liang2023open} predict class-agnostic segmentation masks and classify their crops with a frozen CLIP.
ZegCLIP~\cite{zhou2023zegclip} simplifies this into a one-stage approach, learning an attention decoder between CLIP text embeddings and dense vision embeddings.
CAT-Seg~\cite{cho2023cat} learns to aggregate CLIP cost maps. SAN~\cite{xu2023side} trains a side network to adapt a frozen CLIP to segmentation. To avoid overfitting to the training classes, several methods use parameter-efficient training strategies such as prompt tuning~\cite{brown2020language,jia2022visual,zhou2022learning,zhou2023zegclip}, partial fine-tuning~\cite{hu2021lora,cho2023cat}, or adapters~\cite{houlsby2019parameter,sung2022vl,xu2023side}. Due to its simplicity, SemiVL also follows a one-stage framework~\cite{zhou2023zegclip,cho2023cat,xu2023side}. 
In contrast to these methods, we do not assume access to a large-scale segmentation dataset.

\section{Methods}

\begin{figure}
    \centering
    \includegraphics[width=\linewidth]{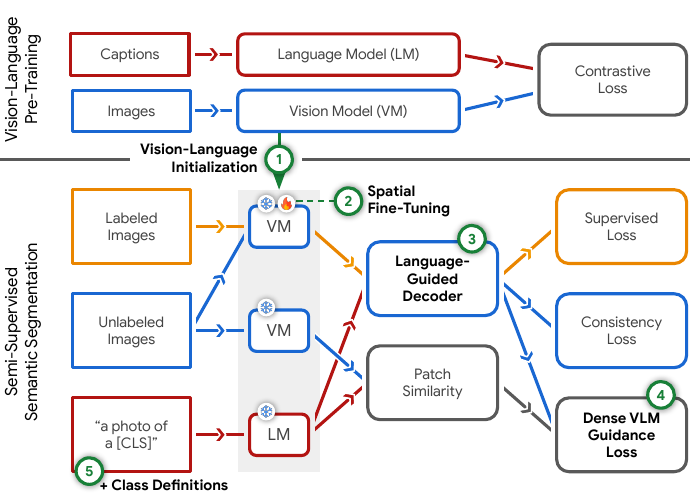}
    \caption{Overview of our \textbf{\ours\ framework}. Utilizing the rich semantic representations from vision-language (VL) pre-training (top), we propose 5 strategies (highlighted in green) to guide semi-supervised semantic segmentation (bottom): We use the rich VL prior as initialization (1) and regularization on unlabeled images (4). To adapt VL from image-level to dense reasoning, we introduce a label-efficient fine-tuning strategy (2) and a decoder architecture jointly reasoning about VL (3). Finally, we propose to steer the model with text instructions in the form of class definitions (5).}
    \label{fig:overview}
\end{figure}

In semi-supervised semantic segmentation, the training dataset consists of a set of labeled images $\mathcal{D}^l = \{(x_i^l, y_i^l)\}$ and another set of unlabeled images $\mathcal{D}^u = \{x_i^u\}$.
In the following, we present our \ours\ framework (see Fig.~\ref{fig:overview}). It is based on the popular consistency training approach to utilize unlabeled images during the semi-supervised training (Sec.~\ref{sec:methods_consistency_training}). While this already achieves good segmentation quality, it still struggles to obtain precise semantic decision boundaries from the limited supervision. To address this, we propose five components to guide the semi-supervised training with vision-language guidance, which are highlighted in green color in Fig.~\ref{fig:overview}. First, we initialize the semi-supervised training with 
a pre-trained VLM as it contains rich semantics learned from web-scale image-text data~(Sec.~\ref{sec:methods_vl_pretraining}).
Second, we only fine-tune the attention layers of the pre-trained vision model to adapt them from image-level to dense reasoning while we freeze the local MLP layers to preserve their semantic reasoning from the pre-training (Sec.~\ref{sec:methods_spatial_finetuning}).
Third, we introduce language-based reasoning in the decoder to further exploit the vision-language alignment from the pre-training (Sec.~\ref{sec:methods_language_guided_decoder}). Fourth, we regularize the training on unlabeled images with predictions from a frozen VLM to anchor the self-training and avoid drift (Sec.~\ref{sec:methods_clip_guidance}). And fifth, we provide the model with language-based class definitions (Sec.~\ref{sec:methods_class_definitions}).

\subsection{Consistency Training}
\label{sec:methods_consistency_training}

While semantic segmentation models are usually trained with a supervised loss $\mathcal{L}_s$ such as the pixel-wise cross-entropy, this is only possible on the labeled images. 
In order to additionally utilize the unlabeled images for the semi-supervised training, we resort to consistency training~\cite{sohn2020fixmatch,chen2021semi,yang2023revisiting}. It is based on the idea that the predictions of the same image should be invariant to different data augmentations or model perturbations. Specifically, the consistency loss term $\mathcal{C}$ drives the model to produce the same predictions $p^p$ under perturbations as the predictions $p^u$ without perturbations, also called pseudo-labels:

\begin{equation}
    \mathcal{C}(p^p, p^u) = \sum_{i,j} \mathbbm{1} [\text{max}(p^u_{ij}) \geq \tau] H(p^p_{ij}, p^u_{ij})\,,
\end{equation}
where $i,j$ are the pixel indices, $H$ is the cross-entropy between $p^u$ and $p^p$, and $\tau$ is a fixed confidence threshold to exclude noisy pseudo-labels $p^u$ from the consistency training. Consistency training is a variant of self-training as the pseudo-labels $p^u$ are generated by the same network as $p^p$.
The pseudo-labels $p^u$ of model $f$ are obtained from the unlabeled images: $p^u = f(x^u)$. The perturbed predictions $p^p$ are obtained under strong data augmentations $\mathcal{A}$: $p^p = f(\mathcal{A}(x^u))$. Additionally, perturbations can also achieved by perturbing the features of the model $p^{fp} = h(\mathcal{P}(g(x^w)))$, where $g$ is the encoder and $h$ the decoder of the model $f$.

In particular, we follow the state-of-the-art approach UniMatch~\cite{yang2023revisiting} and enforce the consistency over two strong data augmentations ($p^{p_1}, p^{p_2}$) and feature perturbation ($p^{fp}$)

\begin{equation}
    \mathcal{L}_u = \frac{1}{2} \mathcal{C}(p^{fp}, p^u) + \frac{1}{4} \mathcal{C}(p^{p_1}, p^u) + \frac{1}{4} \mathcal{C}(p^{p_2}, p^u)\,.
    \label{eq:L_u}
\end{equation}

The overall semi-supervised loss is $\mathcal{L} = \frac{1}{2}(\mathcal{L}_s + \mathcal{L}_u)$.

\begin{figure*}
    \centering
    \includegraphics[width=\linewidth]{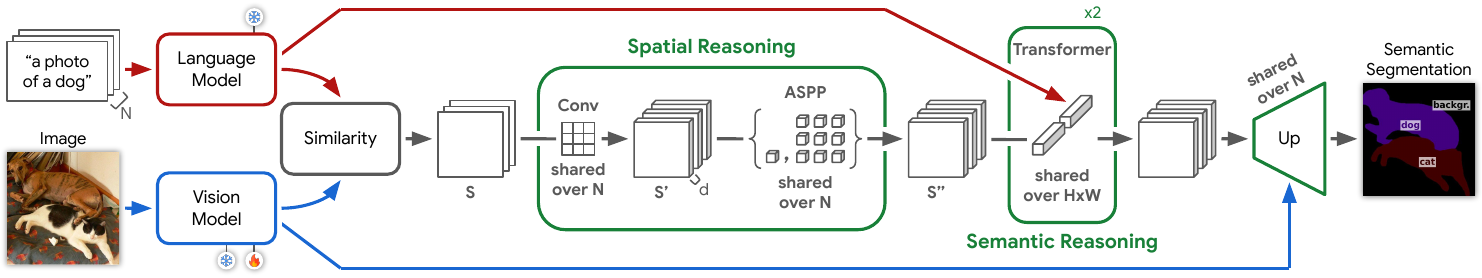}
    \caption{Overview of our \textbf{vision-language-guided architecture}. It is based on dense similarity maps of the vision and text embeddings. These are processed by spatial and semantic reasoning modules and subsequently upsampled. More details are provided in Sec.~\ref{sec:methods_language_guided_decoder}.}
    \label{fig:architecture}
\end{figure*}

\subsection{Vision-Language Pre-Training}
\label{sec:methods_vl_pretraining}
While consistency training achieves good segmentation as shown by previous works, 
it is prone to confusion of visually similar classes as can be seen in Fig.~\ref{fig:motivation}.
This is caused by the model only having a few labeled images to learn the semantic classes. 
VLMs like CLIP~\cite{radford2021learning} are trained on web-scale image-text datasets that cover almost all semantic classes a vision agent can ever come across. CLIP uses a vision encoder $\mathcal{V}$ to produce embeddings for images and a text encoder $\mathcal{T}$ to embed captions.
Both are trained jointly to map to an aligned embedding space, where corresponding images and captions have a high similarity. It is trained in a contrastive manner with ground truth image-caption pairs as positive and shuffled image-caption pairs as negatives. In that way, CLIP learns a representation with a rich semantic understanding of images.
CLIP can perform zero-shot image classification by computing the similarities of an image embedding $\mathcal{V}(x)$ with the text embeddings $\mathcal{T}(t_n)$ of multiple class prompts $t_n$ in the form of ``a photo of a [CLS$_n$]". 

To address the limitations of consistency training, we propose to initialize the semantic segmentation encoder $g$ with the pre-trained VLM vision encoder $\mathcal{V}$ to utilize its rich semantic prior.
Our CLIP initialization replaces the prevailing ImageNet initialization. Compared to image classification pre-training on ImageNet, the VLM pre-training does not require a manually annotated dataset but can be trained on web-crawled image-caption pairs. Further, it can learn richer semantic representations due to the versatile captions, which are not restricted to a specific set of classes.

\subsection{Spatial Fine-Tuning}
\label{sec:methods_spatial_finetuning}

While the vision-language pre-training of CLIP has learned to distinguish fine-grained semantic concepts, it was only trained to optimize image-level features. Therefore, the semantic features are not necessarily spatially aligned with the image content, resulting in noisy segmentations as shown in Fig.~\ref{fig:motivation}d. To mitigate this issue, it is necessary to adapt the backbone towards semantic segmentation.

However, due to the limited number of annotated images, the fine-tuning is prone to overfitting and forgetting the rich semantics from the vision-language pre-training. This process can be further reinforced by the self-confirmation bias of the self-training. Inspired by parameter-efficient fine-tuning~\cite{hu2021lora, jia2022visual, cho2023cat, sung2022vl}, we introduce spatial fine-tuning to semi-supervised semantic segmentation.
A ViT~\cite{dosovitskiy2020image} block consists of a multi-head attention layer and a subsequent MLP.
Only the attention layer models interactions between different patches, while the MLP operates locally on each patch (similar to a $1{\times}1$ convolution).
Spatial fine-tuning only fine-tunes the attention layers, which are responsible for spatial reasoning. In that way, the alignment of semantic features and their corresponding image content can be refined for dense predictions. On the other side, the MLP layers are frozen as they do not perform spatial reasoning to preserve the semantic reasoning capabilities of CLIP pre-training.
We further fine-tune the position embeddings to attribute for the shift from global to dense reasoning as well as input size change from pre-training to fine-tuning.

\subsection{Language-Guided Decoder}
\label{sec:methods_language_guided_decoder}

A major advantage of the vision-language pre-training is the alignment of the vision and language embeddings, which enables the reasoning about both modalities and their semantic relations. To utilize this capability in the model architecture, we integrate the vision-language alignment in a language-guided decoder architecture, which is visualized in Fig.~\ref{fig:architecture}.

For that purpose, we obtain a dense vision-language similarity map $S \in \mathbb{R}^{H\times W \times N}$ with height $H$, width $W$ and number of classes $N$ as proposed in \cite{zhou2022extract} by computing the patch-wise cosine similarities of the vision embeddings $g(x)$ and the text embeddings $\mathcal{T}(t_n)$ over all classes $n \in [0 \twodots N]$ with their text prompt $t_n$
\begin{equation}
    S_{ijn} = \frac{g(x)_{ij} \cdot \mathcal{T}(t_n)}{\|g(x)_{ij}\| \|\mathcal{T}(t_n)\|}\,.
    \label{eq:S}
\end{equation}
Here, $i,j$ denote the patch indices. The embeddings of $g$ and $\mathcal{T}$ are aligned as $g$ is initialized from $\mathcal{V}$ (see Sec.~\ref{sec:methods_vl_pretraining}). While the original localization of semantics in the similarity map is noisy at the beginning of the training, it is improved during the spatial fine-tuning.
The similarity maps are further refined by spatial and semantic reasoning modules.

The spatial reasoning module operates on each class similarity map independently (i.e. $S_n \in \mathbb{R}^{H \times W \times 1}$) and models no inter-class relations. Thereby, the learned spatial reasoning is shared across classes.
First, each $S_n$ is processed by a $7{\times}7$ convolution to learn local spatial structures and embed them to similarity volumes $S'_n \in \mathbb{R}^{H \times W \times d}$ of $d$ dimensions. Subsequently, a residual ASPP~\cite{chen2018encoder} processes the obtained similarity volumes to model long-range context relations, resulting in a combined similarity volume for all classes $S'' \in \mathbb{R}^{H \times W \times N \times d}$.

The semantic reasoning module models the relationship between classes. Each pixel in the similarity volumes $S''_{ij} \in \mathbb{R}^{1 \times 1 \times N \times d}$ is processed independently (no spatial interaction) to share the learned relations between spatial locations.
It is implemented as two Transformer~\cite{vaswani2017attention} blocks. Further, the features are supplemented with the $N$ text embeddings of the class names, which act as semantic anchors.

By decoupling spatial and semantic reasoning, the learned weights can be shared over different classes for spatial reasoning and shared over different locations for semantic reasoning. In that way, the limited annotations can be utilized more effectively and overfitting is reduced.

The common vision transformers operate on a 16 times smaller feature resolution than the input. This limits precise segmentation boundaries. Therefore, we add 2 upsampling blocks~\cite{ronneberger2015u}. They learn the upsampling using a transpose convolution. Further, skip connections from earlier encoder layers are used. The upsampled features and the skip features are concatenated and fused by two convolutions. For label-efficient learning, the upsampling blocks operate on each class independently. A final convolution maps the $d$ dimensions to one channel to obtain the logits for a class.

\subsection{Dense CLIP Guidance}
\label{sec:methods_clip_guidance}

While the previous strategies are designed to best transfer the knowledge from vision-language pre-training to semantic segmentation, the self-training on the unlabeled images can cause a drift of the training to erroneous predictions. Incorrect predictions in $p^u$ are used for the self-training and result in a self-confirmation bias. To anchor the self-training on the unlabeled images and reduce this issue, we guide the consistency training with predictions from a frozen auxiliary CLIP model, which cannot drift.

For this purpose, we extract dense vision-language similarity maps $S$ (see Eq.~\ref{eq:S}) with a frozen $\mathcal{V}$ from the unlabeled images $x_u$. The similarity map is processed by a softmax and can be used as a semantic segmentation $p^{DC}$ pseudo-label. As $p^{DC}$ tends to be noisy (see Fig.~\ref{fig:overview}), we only utilize its high confidence predictions exceeding a threshold $\zeta$. To guide the consistency training on the unlabeled images with dense CLIP guidance, we supplement the consistency loss terms $\mathcal{C}$ in Eq.~\ref{eq:L_u} by dense CLIP guidance:
    
\begin{equation}
\begin{aligned}
    \mathcal{C}(p^p, p^u, p^{DC}) &= \mathcal{C}(p^p, p^u) + \\ 
    \lambda_{DC} \sum &\mathbbm{1} [\text{max}(p^{DC}) \geq \zeta] H(p^p, p^{DC})\,,
\end{aligned}
\end{equation}
where $\lambda_{DC}$ weighs the dense CLIP guidance loss term. As the guidance is picked up by the pseudo-labels $p^u$ and spatially refined over the course of the self-training, the guidance is more important at the beginning of the training. Therefore, $\lambda^{DC}$ is linearly decayed during the training.

\begin{figure*}
\footnotesize
\centering
\setlength{\tabcolsep}{3pt}
\begin{tabular}{ccc}
&
\begin{tabularx}{.45\linewidth}{*{4}{Y}}
Image & G. Truth & UniMatch~\cite{yang2023revisiting} & \ours\ \\
\end{tabularx} &
\begin{tabularx}{.45\linewidth}{*{4}{Y}}
Image & G. Truth & UniMatch~\cite{yang2023revisiting} & \ours\ \\
\end{tabularx} \\

\parbox[t]{2mm}{\multirow{2}{*}{\rotatebox[origin=c]{90}{VOC$_{92}$}}} &
\includegraphics[width=.45\linewidth]{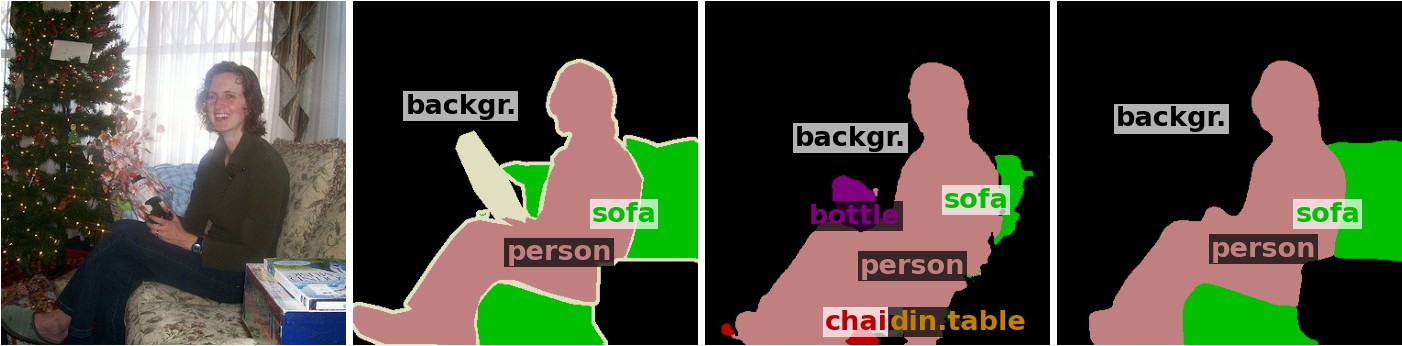} &
\includegraphics[width=.45\linewidth]{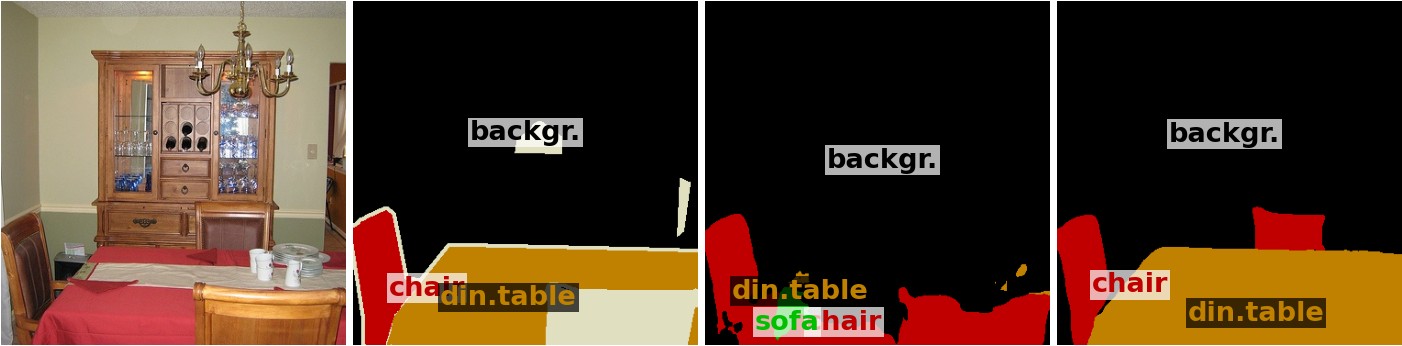} \\[-3pt]
&
\includegraphics[width=.45\linewidth]{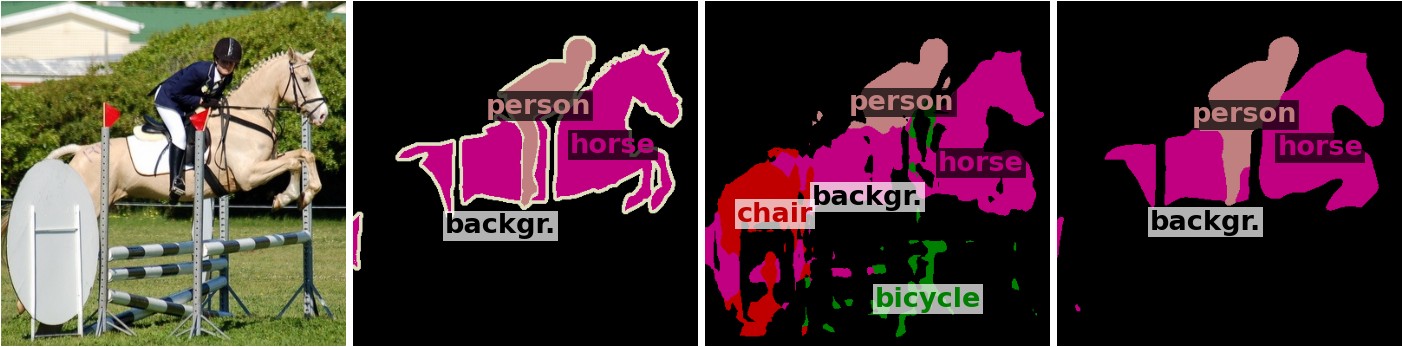} &
\includegraphics[width=.45\linewidth]{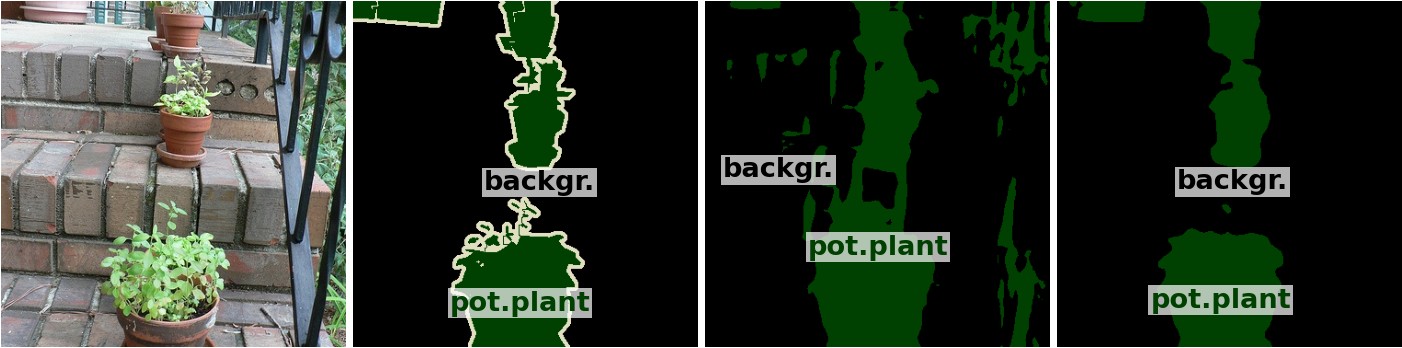} \\[+3pt]

\parbox[t]{2mm}{\multirow{1}{*}{\rotatebox[origin=c]{90}{ADE$_{158}$}}} &
\includegraphics[width=.45\linewidth]{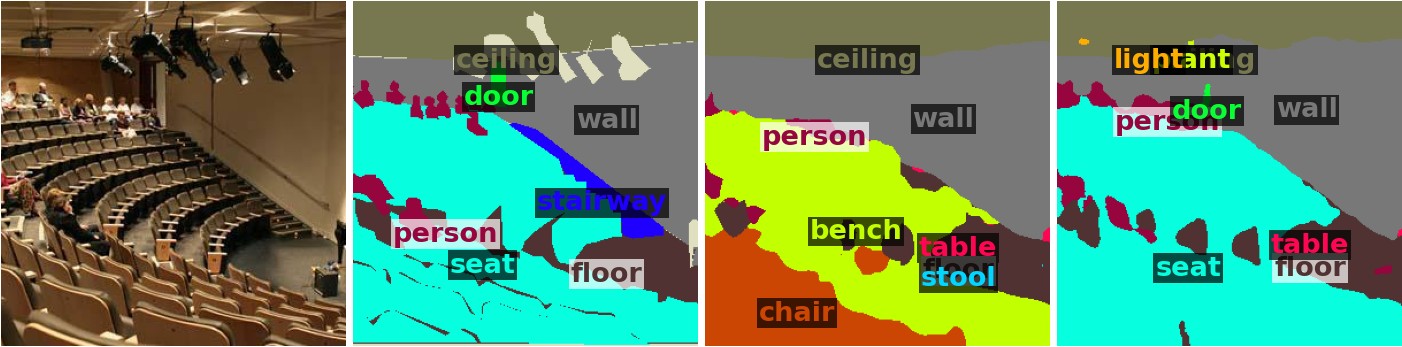} &
\includegraphics[width=.45\linewidth]{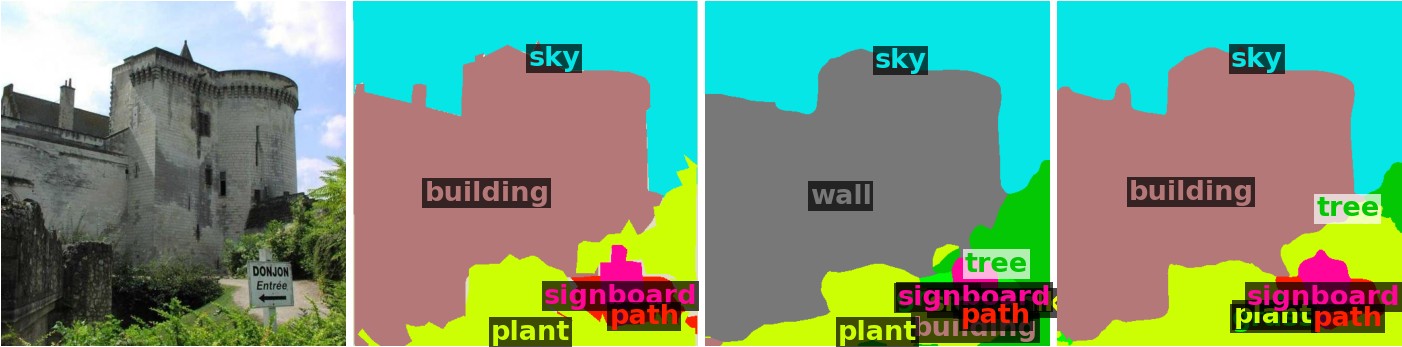} \\[-3pt]
&
\includegraphics[width=.45\linewidth]{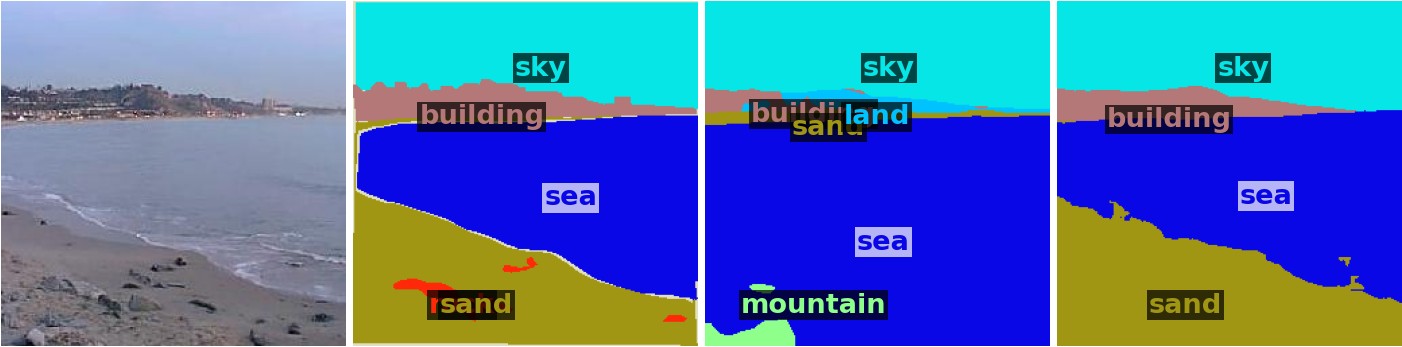} &
\includegraphics[width=.45\linewidth]{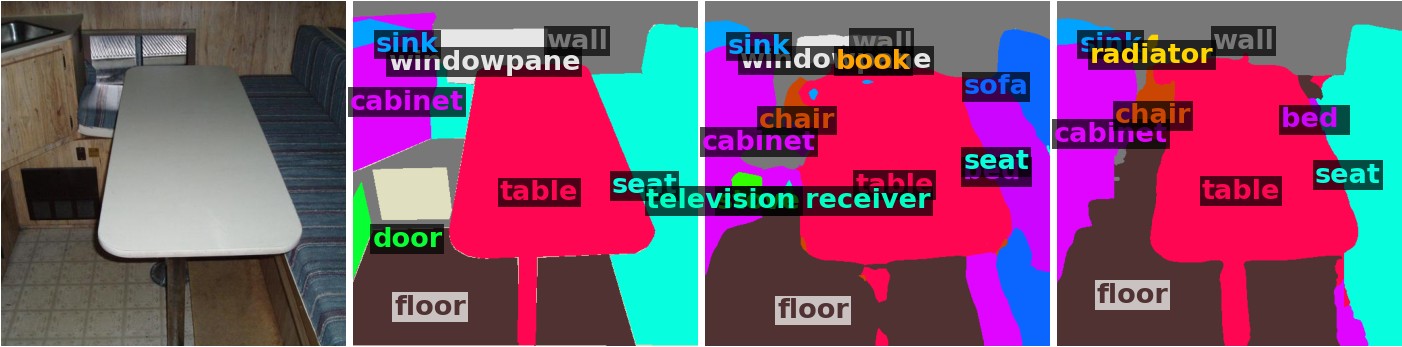}
\end{tabular}

\caption{\textbf{Example predictions} on VOC (92 labels) and ADE20K (158 labels) showing the improved semantic understanding of \ours.}
\label{fig:example_predictions}
\end{figure*}

\subsection{Class Definition Guidance}
\label{sec:methods_class_definitions}

When the number of labeled samples in the semi-supervised setting is small, they might not be sufficient to capture all relevant decision boundaries between classes for the given dataset. This is particularly problematic for edge cases that are not part of the labeled subset or subcategories that could belong to different classes: Does an armchair belong to the \emph{chair} or \emph{sofa} class? Does a bench belong to the \emph{chair} or \emph{background} class? As similar ambiguities happen when humans annotate a dataset, there are usually annotation guidelines with class definitions for a dataset. For example, Pascal VOC defines: ``\emph{chair} includes armchairs, deckchairs but not stools or benches".
As \ours\ can process arbitrary text, we propose to provide these annotation guidelines to the model to better capture the decision boundaries for a dataset.
Importantly, providing the class descriptions is much less effort than mining images of all relevant corner cases and labeling them with semantic segmentation. Compared to generic class descriptions from LLMs~\cite{naeem2023i2mvformer,cupl} or synonyms~\cite{lin2023clip}, this can better capture dataset-specific decision boundaries, e.g. a stool belongs to the \emph{background} class in Pascal VOC.

Even though it would be possible to provide the text encoder $\mathcal{T}$ directly with the class definitions, CLIP was trained with image captions that do not match such definitions. In particular, they do not handle negations well. 
Therefore, we build a set of concepts $b_a$ for each class $a$ that contain additional subcategories or descriptions extracted from the class definitions.
For example, the concept \emph{armchair} belongs to the class \emph{chair} (i.e.  $\textit{armchair} \in b_\mathit{chair}$) while \emph{stool} belongs to \emph{background} according to the VOC definition.

To utilize the concepts, we integrate them into the CLIP guidance training. With the frozen CLIP, we infer the vision-text similarity $p^{concept}$ over all concepts $\cup_a b_a$. We aggregate the concepts back to classes using the concept-class assignment from the class definitions. The concept with the highest score determines the predicted class at position $ij$

\begin{equation}
    p^{DC}_{ija} = \max_{b \in b_a} p^{concept}_{ijb}\,.
\end{equation}

\section{Experiments}

\subsection{Implementation Details}

\paragraph{Architecture:} \ours\ uses a ViT-B/16~\cite{dosovitskiy2020image} vision encoder and a Transformer~\cite{vaswani2017attention} text encoder, both pre-trained by CLIP~\cite{radford2021learning}.
The dense vision embeddings are extracted following~\cite{zhou2022extract}.
The language-guided decoder uses an initial $7{\times}7$ convolution to map the similarity maps to $d{=}128$ channels, a residual ASPP~\cite{chen2018encoder} with dilation rates \{6,12,18\} for spatial reasoning, and 2 Transformer blocks~\cite{vaswani2017attention} with 4 heads for semantic reasoning. For compute and memory efficiency, the semantic reasoning operates on $4{\times}4$ average-pooled features maps. The 2 upsampling blocks use a $2{\times}2$ transpose convolution, skip connections with 16/32 channels from the first/fourth ViT blocks, and 2 convolutions with GroupNorm~\cite{wu2018group} to fuse both into 64/32 channels. For Cityscapes, we use the feature map of the first ResNet block as skip connection to handle particularly small segments.

\noindent\textbf{Training:}
\ours\ is benchmarked on Pascal VOC 2012~\cite{everingham2010pascal}, COCO~\cite{caesar2018coco}, ADE20K~\cite{zhou2017scene}, and Cityscapes~\cite{cordts2016cityscapes} using the same labeled subsets as common protocol~\cite{zou2020pseudoseg,chen2021semi,wang2022semi,yang2023revisiting}.
It is trained with a batch of 8 labeled and 8 unlabeled images using AdamW~\cite{loshchilov2017decoupled} for 80/10/40/240 epochs on VOC/COCO/ADE20K/Cityscapes. The inital learning rate is $1{\times}10^{-4}$/$4{\times}10^{-4}$/$4{\times}10^{-4}$/$5{\times}10^{-5}$ with a $0.9$ polynomial decay. Spatial fine-tuning uses a learning rate multiplier of $0.01$/$0.001$/$0.001$/$0.1$. We use $801{\times}801$ random crops for Cityscapes following~\cite{yang2023revisiting} and $512{\times}512$ for the others. For inference, sliding window evaluation is used.
Following UniMatch~\cite{yang2023revisiting}, we set $\tau{=}0.95$, use 50\% channel dropout for $\mathcal{P}$, and use random color jitter, grayscale, CutMix, scale, and crop for $\mathcal{A}$.
For \ours, we set $\zeta{=}0.9$ and use a linear schedule from 0.1 to 0 for $\lambda_{DC}$. The class definitions are provided in the supplement. The training is conducted on 4 (for VOC) or 8 (for others) A100 GPUs.

\begin{table}
\caption{State-of-the-art comparison on \textbf{Pascal VOC}. The mIoU (\%) is compared across different splits for the labeled subset $\mathcal{D}^l$.
$^\dagger$~denotes re-produced results in the same setting as \ours.
}
\label{tab:sota_voc}
\centering
\scriptsize
\setlength{\tabcolsep}{2pt}
\begin{tabular}{lllccccc}
\toprule
   \multirow{2}{*}{Method} &                &     \multirow{2}{*}{Net} &     1/115 &    1/58 &    1/29 &    1/14 &    1/7 \\
   & & & (92) & (183) & (366) & (732) & (1464) \\ 
\midrule
PseudoSeg~\cite{zou2020pseudoseg} & \venue{ICLR'21} & R101 & 57.6 & 65.5 & 69.1 & 72.4 & -- \\
                     CPS~\cite{chen2021semi} &                        \venue{CVPR'21} &                      R101 &           64.1 &           67.4 &           71.7 &           75.9 &             -- \\
                    ST++~\cite{yang2022st++} &                        \venue{CVPR'22} &                      R101 &           65.2 &           71.0 &           74.6 &           77.3 &           79.1 \\
                 U$^2$PL~\cite{wang2022semi} &                        \venue{CVPR'22} &                      R101 &           68.0 &           69.2 &           73.7 &           76.2 &           79.5 \\
                      PCR~\cite{xu2022semi} &                     \venue{NeurIPS'22} &                      R101 &           70.1 &           74.7 &           77.2 &           78.5 &           80.7 \\
                   ESL~\cite{ma2023enhanced} &                        \venue{ICCV'23} &                      R101 &           71.0 &           74.0 &           78.1 &           79.5 &           81.8 \\
             LogicDiag~\cite{liang2023logic} &                        \venue{ICCV'23} &                      R101 &           73.3 &           76.7 &           77.9 &           79.4 &             -- \\
          UniMatch~\cite{yang2023revisiting} &                        \venue{CVPR'23} &                      R101 &           75.2 &           77.2 &           78.8 &           79.9 &           81.2 \\
                  3-CPS~\cite{li2023diverse} &                        \venue{ICCV'23} &                      R101 &           75.7 &           77.7 &           80.1 &           80.9 &           82.0 \\
\midrule
    ZegCLIP$^\dagger$~\cite{zhou2023zegclip} &                        \venue{CVPR'23} &                  ViT-B/16 &           69.3 &           74.2 &           78.7 &           81.0 &           82.0 \\
                  ZegCLIP+UniMatch$^\dagger$ &                  \textcolor{gray}{---} &                  ViT-B/16 &           78.0 &           80.3 &           80.9 &           82.8 &           83.6 \\
\midrule
UniMatch$^\dagger$~\cite{yang2023revisiting} &                        \venue{CVPR'23} &                  ViT-B/16 &           77.9 &           80.1 &           82.0 &           83.3 &           84.0 \\
                       \multirow{2}{*}{Ours} & \multirow{2}{*}{\textcolor{gray}{---}} & \multirow{2}{*}{ViT-B/16} &           \textbf{84.0} &           \textbf{85.6} &           \textbf{86.0} &           \textbf{86.7} &           \textbf{87.3} \\
                                             &                                        &                           & \green{(+6.1)} & \green{(+5.5)} & \green{(+4.0)} & \green{(+3.4)} & \green{(+3.3)} \\
\bottomrule
\end{tabular}
\end{table}
\begin{table}
    \centering
    \caption{\textbf{Class-Wise IoU} on Pascal VOC with 92 Labels.}
    \label{fig:classwise}
    \includegraphics[width=\linewidth]{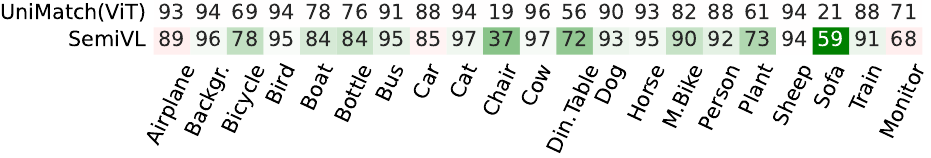}
\end{table}
\begin{table}
\centering
\caption{Comparison with state-of-the-art methods on \textbf{COCO}.}
\label{tab:sota_coco}
\scriptsize
\setlength{\tabcolsep}{3pt}
\begin{tabular}{lllccccc}
\toprule
   \multirow{2}{*}{Method} &                &     \multirow{2}{*}{Net} &     1/512 &    1/256 &    1/128 &    1/64 &    1/32 \\
   & & & (232) & (463) & (925) & (1849) & (3697) \\ 
\midrule
           PseudoSeg~\cite{zou2020pseudoseg} &                        \venue{ICLR'21} &                      XC65 &            29.8 &           37.1 &           39.1 &           41.8 &           43.6 \\
             PC$^2$Seg~\cite{zhong2021pixel} &                        \venue{ICCV'21} &                      XC65 &            29.9 &           37.5 &           40.1 &           43.7 &           46.1 \\
          UniMatch~\cite{yang2023revisiting} &                        \venue{CVPR'23} &                      XC65 &            31.9 &           38.9 &           44.4 &           48.2 &           49.8 \\
             LogicDiag~\cite{liang2023logic} &                        \venue{ICCV'23} &                      XC65 &            33.1 &           40.3 &           45.4 &           48.8 &           50.5 \\
\midrule
UniMatch$^\dagger$~\cite{yang2023revisiting} &                        \venue{CVPR'23} &                  ViT-B/16 &            36.6 &           44.1 &           49.1 &           53.5 &           55.0 \\
                       \multirow{2}{*}{Ours} & \multirow{2}{*}{\textcolor{gray}{---}} & \multirow{2}{*}{ViT-B/16} &            \textbf{50.1} &           \textbf{52.8} &           \textbf{53.6} &           \textbf{55.4} &           \textbf{56.5} \\
                                             &                                        &                           & \green{(+13.5)} & \green{(+8.7)} & \green{(+4.5)} & \green{(+1.9)} & \green{(+1.5)} \\
\bottomrule
\end{tabular}
\end{table}

\subsection{Comparison with the State of the Art}

We compare \ours\ with previous semi-supervised methods on four popular semantic segmentation datasets in Tab.~\ref{tab:sota_voc}-\ref{tab:sota_cityscapes}.
We additionally reproduce the SOTA method UniMatch~\cite{yang2023revisiting} with a ViT~\cite{dosovitskiy2020image} backbone and the same training parameters as \ours\ for a fair comparison.
On all four datasets, \ours\ achieves consistent and significant gains over previous works as detailed in the following.

\noindent\textbf{Pascal VOC:} The Pascal VOC dataset~\cite{everingham2010pascal} contains 10582 training images of which 1464 images have segmentation annotations for 21 classes. 
Tab.~\ref{tab:sota_voc} compares \ours\ with previous methods for different portions of labeled images ranging from 1/115 of the training set (92 labels) up to 1/7 (1464 labels).
\ours\ significantly improves the performance from +3.3 mIoU (1464 labels) up to +6.1 (92 labels) over UniMatch(ViT). The larger improvement for fewer annotations shows the efficiency of \ours\ to learn from limited annotations. \ours\ can reduce the need for labels by a factor of 8 compared to UniMatch(ViT). 
As a further strong baseline using CLIP, we integrate the open-vocabulary method ZegCLIP~\cite{zhou2023zegclip} into UniMatch. Even though it improves the performance over UniMatch, \ours\ still significantly outperforms it. The class-wise analysis in Tab.~\ref{fig:classwise} shows that \ours\ particularly improves classes, where UniMatch struggles (low IoU), such as chair, dining table, or sofa.
These are classes that are often confused with the background class or semantically similar classes.
This is also reflected in the example predictions in Fig.~\ref{fig:example_predictions}.

\begin{table}
\centering
\caption{Comparison with state-of-the-art methods on \textbf{ADE20K}.}
\label{tab:sota_ade}
\scriptsize
\setlength{\tabcolsep}{3pt}
\begin{tabular}{lllccccc}
\toprule
\multirow{2}{*}{Method} &                &     \multirow{2}{*}{Net} &     1/128 &    1/64 &    1/32 &    1/16 & 1/8 \\
   & & & (158) & (316) & (632) & (1263) & (2526) \\ 
\midrule
                CutMix~\cite{french2019semi} &                        \venue{BMVC'20} &                      R101 &             -- &             -- &           26.2 &           29.8 &           35.6 \\
                       AEL~\cite{hu2021semi} &                     \venue{NeurIPS'21} &                      R101 &             -- &             -- &           28.4 &           33.2 &           38.0 \\
          UniMatch~\cite{yang2023revisiting} &                        \venue{CVPR'23} &                      R101 &           15.6 &           21.6 &           28.1 &           31.5 &           34.6 \\
\midrule
UniMatch$^\dagger$~\cite{yang2023revisiting} &                        \venue{CVPR'23} &                  ViT-B/16 &           18.4 &           25.3 &           31.2 &           34.4 &           38.0 \\
\multirow{2}{*}{Ours} & \multirow{2}{*}{\textcolor{gray}{---}} & \multirow{2}{*}{ViT-B/16} &           \textbf{28.1} &           \textbf{33.7} &           \textbf{35.1} &           \textbf{37.2} &           \textbf{39.4} \\
                                             &                                        &                           & \green{(+9.7)} & \green{(+8.4)} & \green{(+3.9)} & \green{(+2.8)} & \green{(+1.4)} \\

\bottomrule
\end{tabular}
\end{table}
\begin{table}
\centering
\caption{Comparison with state-of-the-art methods on \textbf{Cityscapes}.}
\label{tab:sota_cityscapes}
\scriptsize
\setlength{\tabcolsep}{3pt}
\begin{tabular}{lllccccc}
\toprule
   \multirow{2}{*}{Method} &                &     \multirow{2}{*}{Net} & 1/30 &     1/16 &            1/8 &            1/4 &            1/2 \\
   & & & (100) & (186) & (372) & (744) & (1488) \\ 
\midrule
PseudoSeg~\cite{zou2020pseudoseg} & \venue{ICLR'21} & R101 & 61.0 & -- & 69.8 & 72.4 & -- \\
                     CPS~\cite{chen2021semi} &                        \venue{CVPR'21} &                      R101 & -- &           69.8 &           74.3 &           74.6 &           76.8 \\
                 U$^2$PL~\cite{wang2022semi} &                        \venue{CVPR'22} &                      R101 & -- &           74.9 &           76.5 &           78.5 &           79.1 \\
                      PCR~\cite{xu2022semi} &                     \venue{NeurIPS'22} &                      R101 & -- &           73.4 &           76.3 &           78.4 &           79.1 \\
                       AEL~\cite{hu2021semi} &                     \venue{NeurIPS'21} &                      R101 & -- &           75.8 &           77.9 &           79.0 &           80.3 \\
                   ESL~\cite{ma2023enhanced} &                        \venue{ICCV'23} &                      R101 & -- &           75.1 &           77.2 &           78.9 &           80.5 \\
                  3-CPS~\cite{li2023diverse} &                        \venue{ICCV'23} &                      R101 & -- &           75.7 &           77.4 &           78.5 &             -- \\
          UniMatch~\cite{yang2023revisiting} &                        \venue{CVPR'23} &                      R101 & 73.0 &          76.6 &           77.9 &           79.2 &           79.5 \\
             LogicDiag~\cite{liang2023logic} &                        \venue{ICCV'23} &                      R101 & -- &          76.8 &           78.9 &           80.2 &           \textbf{81.0} \\
\midrule
UniMatch$^\dagger$~\cite{yang2023revisiting} &                        \venue{CVPR'23} &                  ViT-B/16 & 73.8 &          76.6 &           78.2 &           79.1 &           79.6 \\
                       \multirow{2}{*}{Ours} & \multirow{2}{*}{\textcolor{gray}{---}} & \multirow{2}{*}{ViT-B/16} & \textbf{76.2} &           \textbf{77.9} &           \textbf{79.4} &           \textbf{80.3} &           80.6 \\
                                             &                                        &                           & \green{(+2.4)} & \green{(+1.3)} & \green{(+1.2)} & \green{(+1.2)} & \green{(+1.0)} \\

\bottomrule
\end{tabular}
\end{table}

\noindent\textbf{COCO:} The COCO dataset~\cite{caesar2018coco} has 118k training images with segmentation annotations for 81 classes. The higher number of classes makes it more challenging for semi-supervised learning. Here, the power of vision-language guidance to distinguish semantic concepts particularly shines. \ours\ achieves major mIoU gains of up to +13.5 (for 232 labels) over UniMatch(ViT) as shown in Tab.~\ref{tab:sota_coco}. 

\noindent\textbf{ADE20K:} The ADE dataset~\cite{zhou2017scene} has 20210 training images with segmentation annotations for 150 classes and is even more challenging than COCO. Tab.~\ref{tab:sota_ade} shows that \ours\ achieves gains of up to +9.7 mIoU (for 158 labels).

\noindent\textbf{Cityscapes:} The Cityscapes dataset~\cite{cordts2016cityscapes} has 2975 training images of street scenes with fine annotations for 19 classes. Tab.~\ref{tab:sota_cityscapes} shows that \ours\ outperforms UniMatch+ViT by up to +2.4 mIoU (100 labels). We hypothesize that the improvements are smaller than for the other benchmarks due to many challenging small classes in Cityscapes (e.g. distant poles or traffic signs), which are usually not part of the captions used in vision-language pre-training, limiting the efficiency of vision-language guidance.

\begin{table}
\centering
\caption{\textbf{Ablation study} of \ours's components on Pascal VOC: vision-language pre-training (VL Pretr.), spatial fine-tuning (SFT), language-guided decoder (Lang.Dec.), dense CLIP guidance (CLIP Guid.), and Class Definition Guidance (Cls.Def.).}
\label{tab:ablation}
\scriptsize
\setlength{\tabcolsep}{3pt}
\begin{tabular}{ccccccccc}
\toprule
VL Pretr. & SFT & Lang.Dec. & CLIP Guid. & Cls.Def. & \multicolumn{2}{c}{mIoU$_{92}$} & \multicolumn{2}{c}{mIoU$_{1464}$} \\
\midrule
\n & \n & \n & \n & \n & 77.9 & \blarrow & 84.0 & \blarrow\\
\y & \n & \n & \n & \n & 80.6 & +2.7 & 86.0 & +2.0\\
\y & \y & \n & \n & \n & 81.7 & +3.8 & 86.7 & +2.7\\
\y & \y & \y & \n & \n & 82.7 & +4.8 & 87.3 & +3.3\\
\y & \y & \y & \y & \n & 83.2 & +5.3 & \textbf{87.4} & \textbf{+3.4}\\
\y & \y & \y & \y & \y & \textbf{84.0} & \textbf{+6.1} & 87.3 & +3.3 \\
\bottomrule
\end{tabular}
\end{table}

\begin{table}\centering
\caption{Comparison of \textbf{fine-tuning strategies} on Pascal VOC.}
\label{tab:finetuning}
\scriptsize
\setlength{\tabcolsep}{3pt}
\begin{tabular}{lccc}
\toprule
Fine-Tuning Method  & \#Updated Params. (Encoder) & mIoU$_{92}$ & mIoU$_{1464}$ \\
\midrule
Frozen              & -- & 75.9 & 81.7 \\
Deep Prompt Tuning~\cite{zhou2023zegclip}  & \z0.7 M & 79.5 & 84.3 \\
LoRA~\cite{hu2021lora}                & \z1.4 M  & 80.5 & 85.8 \\
Full Fine-Tuning    & 86.8 M  & 80.6 & 86.0 \\
Spatial Fine-Tuning & 29.1 M & \textbf{81.7} & \textbf{86.7} \\
\bottomrule
\end{tabular}
\end{table}

\begin{table}
\parbox[t][][t]{.45\linewidth}{
\centering
\caption{\textbf{Decoder ablation study} on Pascal VOC.}
\label{tab:decoder_study}
\scriptsize
\setlength{\tabcolsep}{1pt}
\begin{tabular}{lcc}
\toprule
Decoder Variant                      & mIoU$_{92}$ & mIoU$_{1464}$ \\
\midrule
Joint Reasoning              & 64.9 & 86.3\\
VL Similarity                          & 68.6 & 84.5\\
VL S. + Up                      & 74.1 & 86.6\\
VL S. + Up + Spatial            & 80.3 & 86.9\\
\makecell{VL S. + Up + Spatial\\\hspace{6.0mm} + Semantic} & \textbf{82.7} & \textbf{87.3}\\
\bottomrule
\end{tabular}

}%
\hfill
\parbox[t][][t]{.45\linewidth}{
\centering
\caption{Study on \textbf{class definition guidance} on VOC.}
\label{tab:class_definition_study}
\scriptsize
\setlength{\tabcolsep}{3pt}
\begin{tabular}{lc}
\toprule
Class Definition                      & mIoU$_{92}$ \\
\midrule
Dictionary Def. (Text)    & 82.7 \\
GPT Concepts (Max)    & 82.9 \\
Class Name                       & 83.2 \\
Guidelines  (Text)                    & 83.4 \\
Guidel. (Avg)  & 83.4 \\
Ours: Guidel. (Max) & \textbf{84.0}   \\
\bottomrule
\end{tabular}

}
\end{table}

\subsection{Analysis}

\noindent\textbf{Ablation Study:}
Tab.~\ref{tab:ablation} shows the ablation of \ours\ on Pascal VOC. 
The first row shows the UniMatch baseline  with an ImageNet pre-trained ViT encoder and DeepLabv3+ decoder. UniMatch uses consistency training (see Sec.~\ref{sec:methods_consistency_training}).
All components of \ours\ contribute to the major performance gain over UniMatch. The largest improvement originates from the VL pre-training, improving the semantic understanding. But also the spatial fine-tuning and the language-guided decoder significantly improve the performance for both 92 and 1464 labels. The CLIP guidance and the class definitions improve the mIoU for 92 labels, while they do not affect the mIoU for 1464 labels. Probably, the $16{\times}$ more labels provide sufficient information about the class definitions of the dataset so that no additional CLIP guidance is necessary in this case. All components provide stronger improvements for fewer labels, showing their suitability for training with limited annotations.

\noindent\textbf{Fine-Tuning Strategies:}
Tab.~\ref{tab:finetuning} compares different methods to adapt the VLM to semantic segmentation. While a frozen backbone does not overfit to the small labeled subset, it also cannot adapt to segmentation. Deep prompt tuning (DPT)~\cite{zhou2022learning,zhou2023zegclip} learns additional input tokens to manipulate a frozen ViT. We use the same parameters as~\cite{zhou2023zegclip}. It significantly improves over a frozen backbone. LoRA~\cite{hu2021lora} learns low-rank ($r{=}8$) residuals for the attention weights ($W_{qkvo}$).
It improves over DPT achieving a similar performance as full model fine-tuning. Our spatial fine-tuning can further improve over both. On the one side, it preserves the semantic reasoning of the local MLPs and avoids their overfitting to the limited labels. On the other side, its full rank fine-tuning of the attention layers might be necessary to sufficiently adapt the ViT from image-level to dense reasoning.
For all variants, we tuned the backbone learning rate.

\noindent\textbf{Decoder Study:}
\ours's decoder is designed to decouple semantic from spatial reasoning for label-efficient learning.
As a baseline for joint reasoning, we input all classes together into the spatial reasoning ($N$ times more convolution channels) instead of processing each class separately (and sharing the parameters across classes).
Tab.~\ref{tab:decoder_study} shows that decoupled reasoning (82.7 mIoU) significantly outperforms joint reasoning (64.9 mIoU) for 92 labels as the joint reasoning overfits to the small labeled subset.
Further, Tab.~\ref{tab:decoder_study} provides an ablation study over \ours's decoder modules. It shows that each module (upsampling, spatial reasoning, and semantic reasoning) is crucial for its performance.

\noindent\textbf{Class Definitions:}
Tab.~\ref{tab:class_definition_study} compares different strategies for text prompts in CLIP inference.
It shows that class definitions from the dataset annotation guidelines achieve the best performance. Here, taking the maximum score over concepts (Sec.~\ref{sec:methods_class_definitions}) works better than using an averaged text embedding over the concepts or the guidelines as raw text. As baselines, we also benchmark definitions from the Oxford Languages dictionary and class concepts queried from GPT3.5. However, they perform worse, probably because they are dataset agnostic and do not represent the dataset-specific class boundaries as well as the annotation guidelines.

\section{Conclusions}

In this work, we have investigated the impact of vision-language guidance on semi-supervised semantic segmentation. First, we have shown that VL pre-training is a particularly suited strategy with fine-grained semantic understanding. Second, we introduced a spatial fine-tuning strategy to semi-supervised learning to efficiently adapt VL pre-training from image-level to dense understanding. Third, we demonstrated the power of vision-language alignment in a specifically designed decoder architecture. And fourth, we regularized the training with guidance from a frozen CLIP model with class definition guidelines. Our \ours\ framework combines the strategies and provides major performance improvements over previous state-of-the-art methods on multiple benchmarks. Overall, \ours\ preserves performance using only a quarter of the annotated images.

{
    \small
    \bibliographystyle{ieeenat_fullname}
    \bibliography{main}
}

\clearpage

\makeatletter
\renewcommand{\theHsection}{papersection.\number\value{section}} 
\renewcommand{\thesection}{\Alph{section}}
\renewcommand{\thefigure}{S\arabic{figure}}
\renewcommand{\thetable}{S\arabic{table}}
\setcounter{section}{0}
\setcounter{table}{0}
\setcounter{figure}{0}
\makeatother

\maketitlesupplementary

\section{Overview}

The supplementary material of SemiVL provides the source code (Sec.~\ref{sec:supp_code}), studies the influence of the hyperparameters of the dense CLIP guidance (Sec.~\ref{sec:supp_dc_hp}), provides details on the used class definitions (Sec.~\ref{sec:supp_class_definitions}), and qualitatively analyzes SemiVL on four semantic segmentation datasets (Pascal VOC, COCO, ADE20K, and Cityscapes) including a component ablation on Pascal VOC (Sec.~\ref{sec:supp_qualitatives}).

\section{Source Code}
\label{sec:supp_code}

\ifdefined\ANONYM
The source code of SemiVL is provided in the \texttt{code/} folder besides this document.
\else
The source code of SemiVL is provided at \href{https://github.com/google-research/semivl}{https://github.com/google-research/semivl}.
\fi
For further information on the environment setup and experiment execution, please refer to \texttt{README.md}.
SemiVL is implemented in PyTorch.
It is based on the implementations of UniMatch~\cite{yang2023revisiting}, MaskCLIP~\cite{zhou2022extract}, ZegCLIP~\cite{zhou2023zegclip}, and MMSegmentation~\cite{mmseg2020}.

\begin{table}[b]
\centering
\caption{Study on the \textbf{hyperparameters of CLIP guidance} on Pascal VOC.}
\label{tab:clip_guidance_params}
\scriptsize
\setlength{\tabcolsep}{3pt}
\begin{tabular}{ccc}
\toprule
$\lambda_{DC,0}$ & $\zeta$ & mIoU$_{92}$ \\
\midrule
0\phantom{.00}                          & 0.9                            & 82.7                       \\
0.01                       & 0.9                            & 83.6                       \\
0.1\z                        & 0.9                            & \textbf{84.0}                       \\
1\phantom{.00}                          & 0.9                            & 75.7                       \\
\midrule
0.1\z                        & \z0.7                            & 81.5                       \\
0.1\z                        & \z0.8                            & 83.2                       \\
0.1\z                        & \z0.9                            & \textbf{84.0}                       \\
0.1\z                        & 0.99                           & 83.9  \\
\bottomrule
\end{tabular}
\end{table}
\begin{table}
\centering
\caption{Class definitions from \textbf{annotation guidelines as class concepts (ours)} on Pascal VOC.}
\label{tab:class_def_voc_concepts}
\scriptsize
\setlength{\tabcolsep}{3pt}
\begin{tabular}{p{0.17\linewidth} p{0.82\linewidth}}
\toprule
Class $c$ & Concepts $b_c$ \\
\midrule
background & ``background'', ``bed'', ``building'', ``cabinet'', ``ceiling'', ``curtain'', ``door'', ``fence'', ``floor'', ``grass'', ``ground'', ``mountain'', ``road'', ``rock'', ``shelves'', ``sidewalk'', ``sky'', ``snow'', ``tree'', ``wall'', ``water'', ``window'', ``hang glider'', ``helicopter'', ``jet ski'', ``go-cart'', ``tractor'', ``emergency vehicle'', ``lorry'', ``truck'', ``lion'', ``stool'', ``bench'', ``wheelchair'', ``coffee table'', ``desk'', ``side table'', ``picnic bench'', ``wolve'', ``flowers in a vase'', ``goat'', ``tram'', ``laptop'', ``advertising display'', ``vehicle interior'' \\
aeroplane & ``aeroplane'', ``airplane'', ``glider'' \\
bicycle & ``bicycle'', ``tricycle'', ``unicycle'' \\
bird & ``bird'' \\
boat & ``boat'', ``ship'', ``rowing boat'', ``pedalo'' \\
bottle & ``bottle'', ``plastic bottle'', ``glass bottle'', ``feeding bottle'' \\
bus & ``bus'', ``minibus'' \\
car & ``car'', ``van'', ``large family car'', ``realistic toy car'' \\
cat & ``cat'', ``domestic cat'' \\
chair & ``chair'', ``armchair'', ``deckchair'' \\
cow & ``cow'' \\
dining table & ``dining table'', ``table for eating at'' \\
dog & ``dog'', ``domestic dog'' \\
horse & ``horse'', ``pony'', ``donkey'', ``mule'' \\
motorbike & ``motorbike'', ``moped'', ``scooter'', ``sidecar'' \\
person & ``person'', ``people'', ``baby'', ``face'' \\
potted plant & ``potted plant'', ``indoor plant in a pot'', ``outdoor plant in a pot'' \\
sheep & ``sheep'' \\
sofa & ``sofa'' \\
train & ``train'', ``train carriage'' \\
tv/monitor & ``tv'', ``monitor'', ``standalone screen'' \\
\bottomrule
\end{tabular}
\end{table}
\begin{table} [b]
\vspace{20px}
\centering
\caption{Class definitions from \textbf{annotation guidelines as class concepts (ours)} on Cityscapes.}
\label{tab:class_def_cityscapes_concepts}
\scriptsize
\setlength{\tabcolsep}{3pt}
\begin{tabular}{p{0.17\linewidth} p{0.82\linewidth}}
\toprule
Class $c$ & Concepts $b_c$ \\
\midrule
road & ``road'', ``street'', ``parking space'' \\
sidewalk & ``sidewalk'' \\
building & ``building'', ``skyscaper'', ``house'', ``bus stop building'', ``garage'', ``car port'', ``scaffolding'' \\
wall & ``individual standing wall, which is not part of a building'' \\
fence & ``fence'', ``hole in fence'' \\
pole & ``pole'', ``sign pole'', ``traffic light pole'' \\
traffic light & ``traffic light'' \\
traffic sign & ``traffic sign'', ``parking sign'', ``direction sign'' \\
vegetation & ``vegetation'', ``tree'', ``hedge'' \\
terrain & ``terrain'', ``grass'', ``soil'', ``sand'' \\
sky & ``sky'' \\
person & ``person'', ``pedestrian'', ``walking person'', ``standing person'', ``person sitting on the ground'', ``person sitting on a bench'', ``person sitting on a chair'' \\
rider & ``rider'', ``cyclist'', ``motorcyclist'' \\
car & ``car'', ``jeep'', ``SUV'', ``van'' \\
truck & ``truck'', ``box truck'', ``pickup truck'', ``truck trailer'' \\
bus & ``bus'' \\
train & ``train'', ``tram'' \\
motorcycle & ``motorcycle'', ``moped'', ``scooter'' \\
bicycle & ``bicycle'' \\
\bottomrule
\end{tabular}
\end{table}

\section{Dense CLIP Guidance Hyperparameters}
\label{sec:supp_dc_hp}

Tab.~\ref{tab:clip_guidance_params} shows the influence of the hyperparameters of the dense CLIP guidance loss on unlabeled images, i.e. the initial loss weight $\lambda_{DC,0}$ and the confidence threshold $\zeta$. The optimal values are $\lambda_{DC,0}=0.1$ and $\zeta=0.9$. When $\lambda_{DC,0}$ is chosen smaller, the performance gradually decreases. However, a larger $\lambda_{DC,0}$ causes a considerable drop. Probably, the induced gradients from erroneous dense CLIP pseudo-labels become too strong and corrupt the default consistency training. Considering the confidence threshold $\zeta$, the performance gradually degrades around the optimal value. Only when it is chosen too small and less confident CLIP predictions are included, the performance drops below the baseline.

\section{Class Definitions}
\label{sec:supp_class_definitions}

\begin{table}
\centering
\caption{Class definitions from \textbf{annotation guidelines (raw text)} on Pascal VOC.}
\label{tab:class_def_voc_guidelines}
\scriptsize
\setlength{\tabcolsep}{3pt}
\begin{tabular}{p{0.17\linewidth} p{0.82\linewidth}}
\toprule
Class $c$ & Annotation Guidelines \\
\midrule
background & ``background'' \\
aeroplane & ``aeoroplane including gliders but not hang gliders or helicopters'' \\
bicycle & ``bicycle including tricycles, unicycles'' \\
bird & ``bird'' \\
boat & ``boat including ships, rowing boats, pedaloes but not jet skis'' \\
bottle & ``bottle including plastic, glass or feeding bottles'' \\
bus & ``bus including minibus but not trams'' \\
car & ``car including vans, large family cars for 6-8 people, toy cars but not go-carts, tractors, emergency vehicles, lorries, trucks, or the vehicle interior'' \\
cat & ``domestic cat'' \\
chair & ``chair including armchairs, deckchairs, but not stools, wheelchairs, seats in buses or cars'' \\
cow & ``cow'' \\
dining table & ``table for eating at but not coffee tables, desks, side tables or picnic benches'' \\
dog & ``domestic dog (not wolves etc.)'' \\
horse & ``horse including ponies, donkeys, mules etc.'' \\
motorbike & ``motorbike including mopeds, scooters, sidecars'' \\
person & ``person including babies, faces (i.e. truncated people)'' \\
potted plant & ```indoor plants or outdoor plants clearly in a pot but not flowers in vases'' \\
sheep & ``sheep but not a goat'' \\
sofa & ``sofa excluding sofas made up as sofa-beds'' \\
train & ``train including train carriages, excluding trams'' \\
tv/monitor & ``tv/monitor including standalone screens but not laptops nor advertising displays'' \\
\bottomrule
\end{tabular}
\end{table}
\begin{table}
\centering
\caption{Class definitions as \textbf{concepts from GPT} on Pascal VOC.}
\label{tab:class_def_voc_gpt}
\scriptsize
\setlength{\tabcolsep}{3pt}
\begin{tabular}{p{0.17\linewidth} p{0.82\linewidth}}
\toprule
Class $c$ & GPT Concepts \\
\midrule
background & ``background'', ``scene'', ``environment'', ``setting'', ``context'' \\
aeroplane & ``aeroplane'', ``aircraft'', ``plane'', ``jet'', ``aviation'' \\
bicycle & ``bicycle'', ``bike'', ``cycle'', ``pedal'', ``two-wheeler'' \\
bird & ``bird'', ``avian'', ``feathered creature'', ``fowl'', ``winged animal'' \\
boat & ``boat'', ``vessel'', ``watercraft'', ``ship'', ``sailboat'' \\
bottle & ``bottle'', ``flask'', ``container'', ``jar'', ``vial'' \\
bus & ``bus'', ``coach'', ``transit'', ``shuttle'', ``public transport'' \\
car & ``car'', ``automobile'', ``vehicle'', ``motorcar'', ``sedan'' \\
cat & ``cat'', ``feline'', ``kitty'', ``kitten'', ``pussycat'' \\
chair & ``chair'', ``seat'', ``furniture'', ``stool'', ``armchair'' \\
cow & ``cow'', ``bovine'', ``cattle'', ``ox'', ``livestock'' \\
diningtable & ``diningtable'', ``table'', ``dining furniture'', ``dinner table'', ``kitchen table'' \\
dog & ``dog'', ``canine'', ``pooch'', ``puppy'', ``man's best friend'' \\
horse & ``horse'', ``equine'', ``stallion'', ``pony'', ``mare'' \\
motorbike & ``motorbike'', ``motorcycle'', ``bike'', ``motor'', ``two-wheeled vehicle'' \\
person & ``person'', ``human'', ``individual'', ``human being'', ``someone'' \\
pottedplant & ``pottedplant'', ``pot plant'', ``houseplant'', ``potted flower'', ``indoor plant'' \\
sheep & ``sheep'', ``lamb'', ``ewe'', ``ram'', ``woolly animal'' \\
sofa & ``sofa'', ``couch'', ``settee'', ``divan'', ``lounge'' \\
train & ``train'', ``locomotive'', ``railway vehicle'', ``railroad train'', ``engine'' \\
tv/monitor & ``tv/monitor'', ``television'', ``screen'', ``display'', ``monitor'' \\
\bottomrule
\end{tabular}
\end{table}
\begin{table}
\centering
\caption{Class definition as \textbf{definitions from Oxford Languages} on Pascal VOC.}
\label{tab:class_def_voc_oxford}
\scriptsize
\setlength{\tabcolsep}{3pt}
\begin{tabular}{p{0.17\linewidth} p{0.82\linewidth}}
\toprule
Class $c$ & Oxford Languages Definition \\
\midrule
background & ``background'' \\
aeroplane & ``aeroplane'', ``a flying vehicle with fixed wings'' \\
bicycle & ``bicycle'', ``a vehicle consisting of two wheels held in a frame one behind the other, propelled by pedals and steered with handlebars attached to the front wheel'' \\
bird & ``bird'', ``a warm-blooded egg-laying vertebrate animal distinguished by the possession of feathers, wings, a beak, and typically by being able to fly'' \\
boat & ``boat'', ``a vessel for travelling over water, propelled by oars, sails, or an engine'' \\
bottle & ``bottle'', ``a glass or plastic container with a narrow neck, used for storing drinks or other liquids'' \\
bus & ``bus'', ``a large motor vehicle carrying passengers by road'' \\
car & ``car'', ``a four-wheeled road vehicle that is powered by an engine and is able to carry a small number of people'' \\
cat & ``cat'', ``a small domesticated carnivorous mammal with soft fur, a short snout, and retractable claws'' \\
chair & ``chair'', ``a separate seat for one person, typically with a back and four legs'' \\
cow & ``cow'', ``a fully grown female animal of a domesticated breed of ox, kept to produce milk or beef'' \\
dining table & ``dining table'', ``a table on which meals are served in a dining room'' \\
dog & ``dog'', ``a domesticated carnivorous mammal that typically has a long snout and non-retractable claws'' \\
horse & ``horse'', ``a large plant-eating domesticated mammal with solid hoofs and a flowing mane and tail, used for riding, racing, and to carry and pull loads'' \\
motorbike & ``motorbike'', ``a two-wheeled vehicle that is powered by a motor and has no pedals'' \\
person & ``person'', ``a human being regarded as an individual'' \\
potted plant & ``potted plant'', ``a plant in a pot'' \\
sheep & ``sheep'', ``a domesticated ruminant mammal with a thick woolly coat'' \\
sofa & ``sofa'', ``a long upholstered seat with a back and arms, for two or more people'' \\
train & ``train'', ``a series of connected railway carriages or wagons moved by a locomotive or by integral motors'' \\
tv/monitor & ``tv/monitor'', ``a device for watching television'' \\
\bottomrule
\end{tabular}
\end{table}

In the following, we provide the used class definitions and their split into concepts. The class definitions are based on the annotation guidelines of Pascal VOC\footnote{\href{http://host.robots.ox.ac.uk/pascal/VOC/voc2011/guidelines.html}{http://host.robots.ox.ac.uk/pascal/VOC/voc2011/guidelines.html}} and Cityscapes\footnote{\href{https://www.cityscapes-dataset.com/dataset-overview/\#class-definitions}{https://www.cityscapes-dataset.com/dataset-overview/\#class-definitions}}. For COCO and ADE20K, we only use the class names as there are no annotation guidelines publicly available. For SemiVL, these free text class definitions are split into concepts $b$, which are assigned to the corresponding class $c$ as described in the main paper. The resulting class concepts are shown in Tab.~\ref{tab:class_def_voc_concepts} for Pascal VOC and in Tab.~\ref{tab:class_def_cityscapes_concepts} for Cityscapes. For, Pascal VOC we further add the background class names from the Pascal Context~\cite{mottaghi2014role} set to $b_\mathit{background}$.

For the class definition study in the main paper, we further use the annotation guidelines as raw text (see Tab.~\ref{tab:class_def_voc_guidelines}), the definitions from the Oxford Languages dictionary (Tab.~\ref{tab:class_def_voc_oxford}), and class concepts obtained from GTP3.5\footnote{Prompt: ``I have the following classes from a semantic segmentation dataset: [`background', `aeroplane', `bicycle', `bird', `boat', `bottle',  `bus', `car', `cat', `chair', `cow', `dining table', `dog',  `horse', `motorbike', `person', `potted plant', `sheep', `sofa',  `train', `tv/monitor']. Please, provide 5 distinctive subcategories or synonyms for each class so that a vision language model can distinguish it from the other classes. None of these words should cause confusion with other classes. These words will be used in the prompt ``a photo of a X". Please, provide the output as a nested python list.''} (Tab.~\ref{tab:class_def_voc_gpt}). In particular, it can be seen that by not knowing the dataset-specific decision boundaries, GPT concepts are not well aligned with the used dataset. For example, GPT produces ``stool'' as a concept for the class \emph{chair}, while ``stool'' is considered \emph{background} in Pascal VOC. Similar problems apply for ``container'', ``jar'', ``furniture'', and ``table'' (too generic).

\section{Extended Qualitative Comparison}
\label{sec:supp_qualitatives}

\begin{figure*}
\footnotesize
\centering
\begin{tabularx}{.48\linewidth}{*{4}{Y}}
Image & G. Truth & UniMatch~\cite{yang2023revisiting} & \ours\ \\
\end{tabularx} 
\begin{tabularx}{.48\linewidth}{*{4}{Y}}
Image & G. Truth & UniMatch~\cite{yang2023revisiting} & \ours\ \\
\end{tabularx} \\
\includegraphics[width=.48\linewidth]{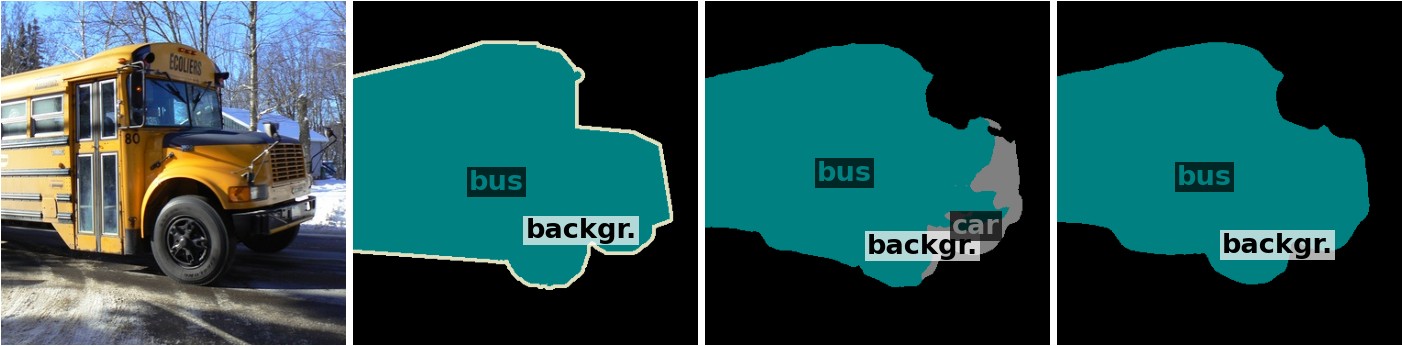}
\includegraphics[width=.48\linewidth]{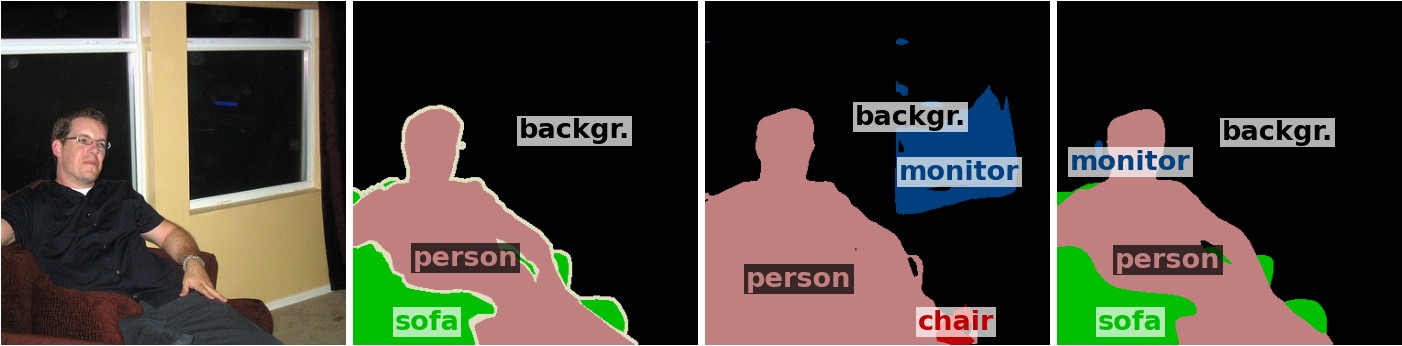}
\includegraphics[width=.48\linewidth]{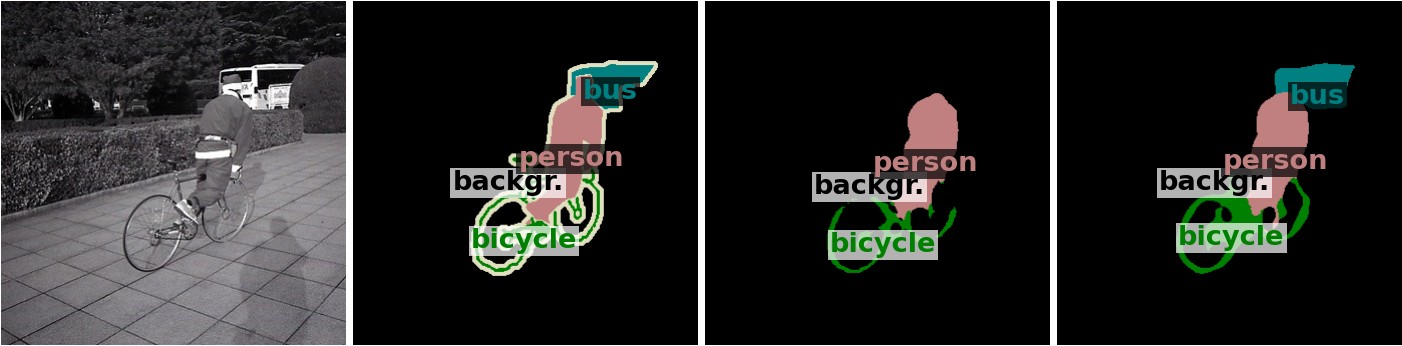}
\includegraphics[width=.48\linewidth]{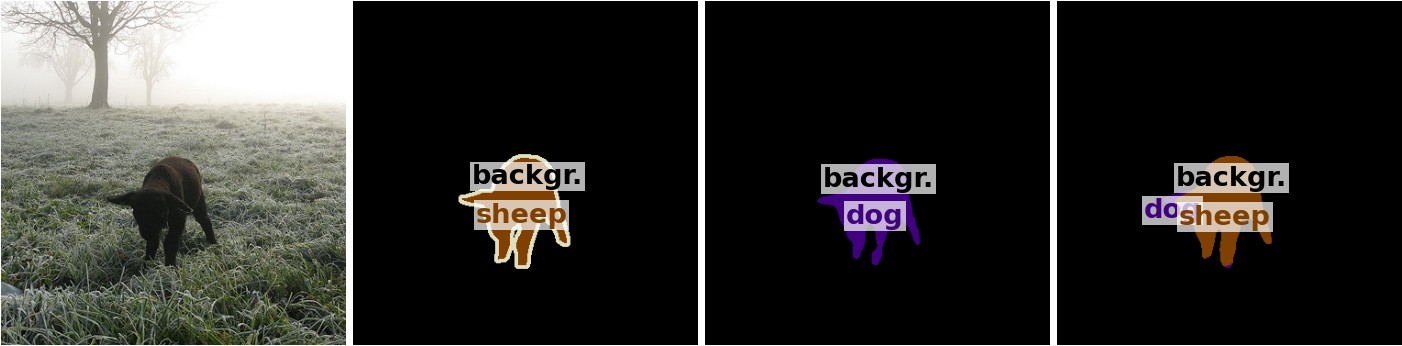}
\includegraphics[width=.48\linewidth]{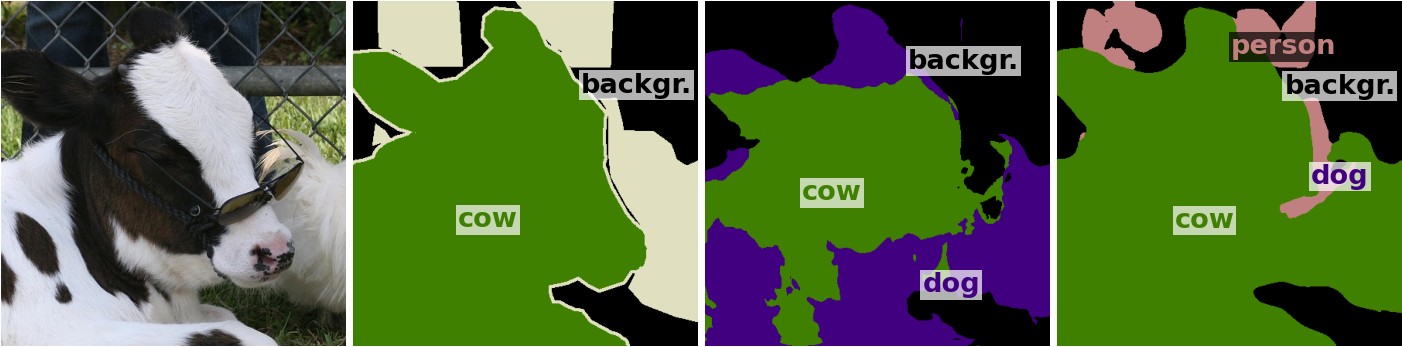}
\includegraphics[width=.48\linewidth]{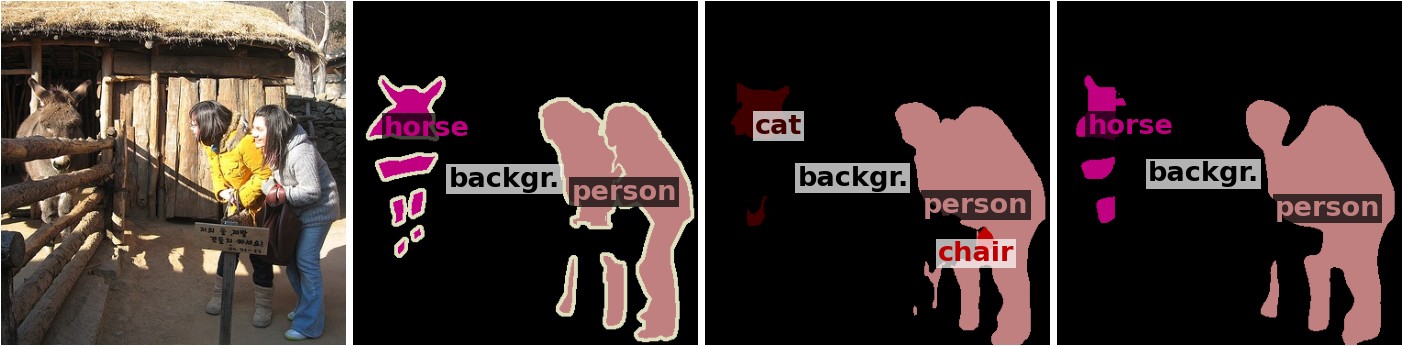}
\includegraphics[width=.48\linewidth]{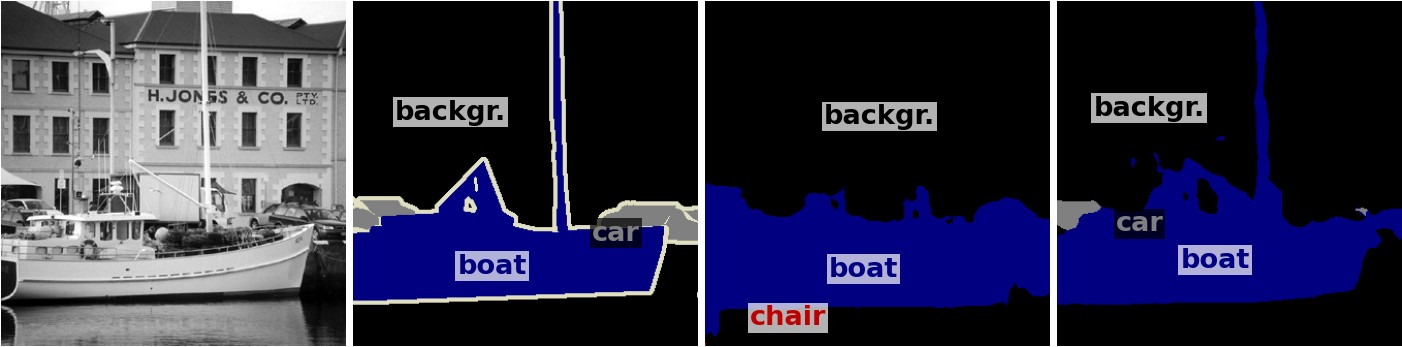}
\includegraphics[width=.48\linewidth]{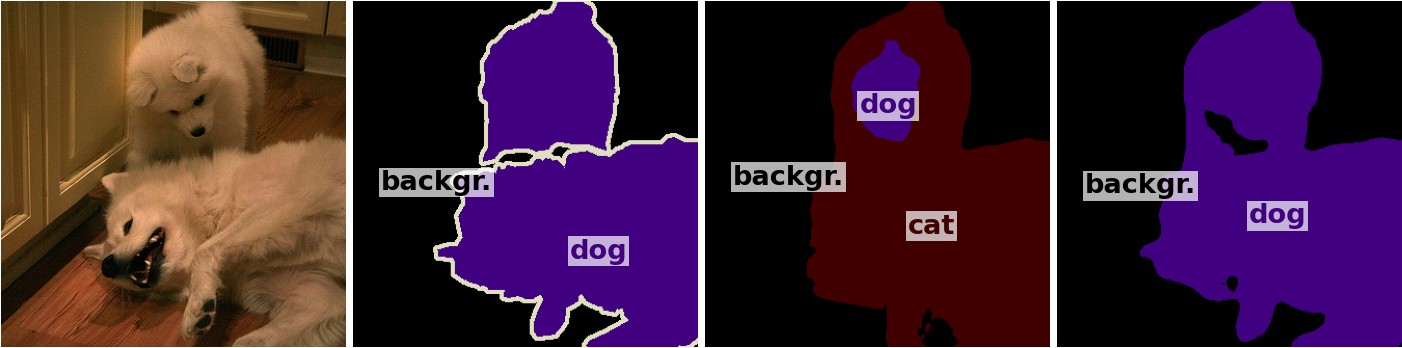}
\includegraphics[width=.48\linewidth]{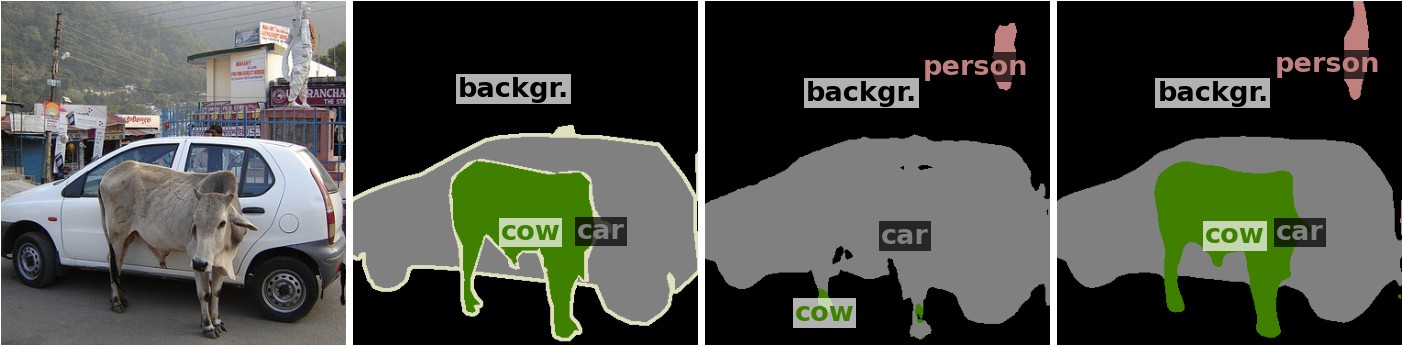}
\includegraphics[width=.48\linewidth]{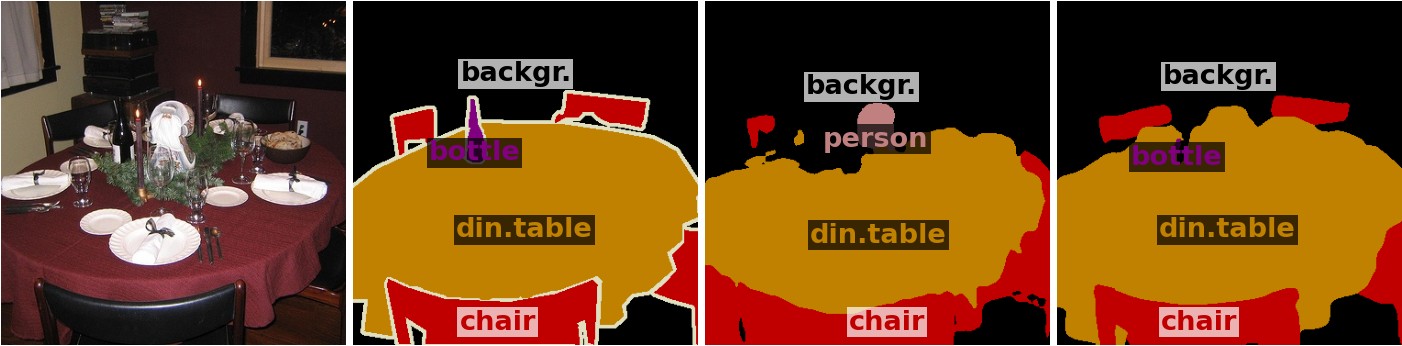}
\includegraphics[width=.48\linewidth]{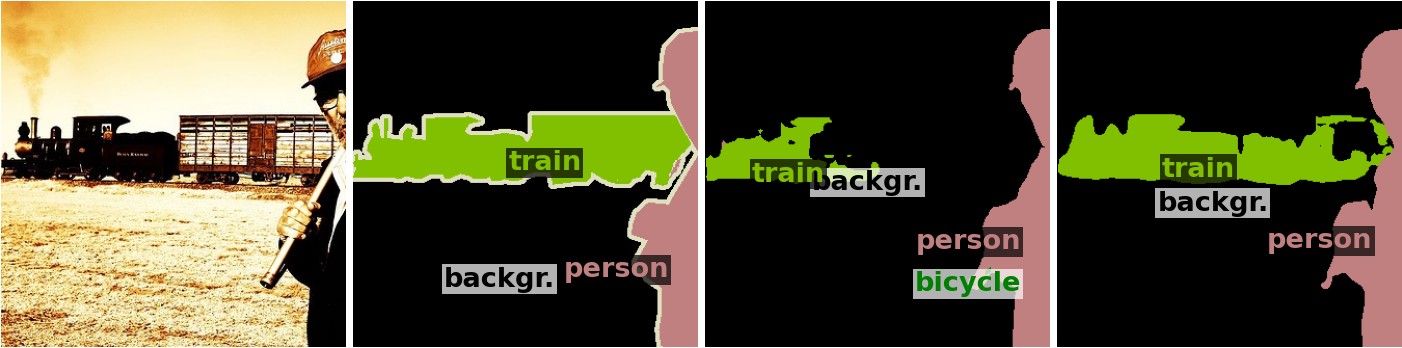}
\includegraphics[width=.48\linewidth]{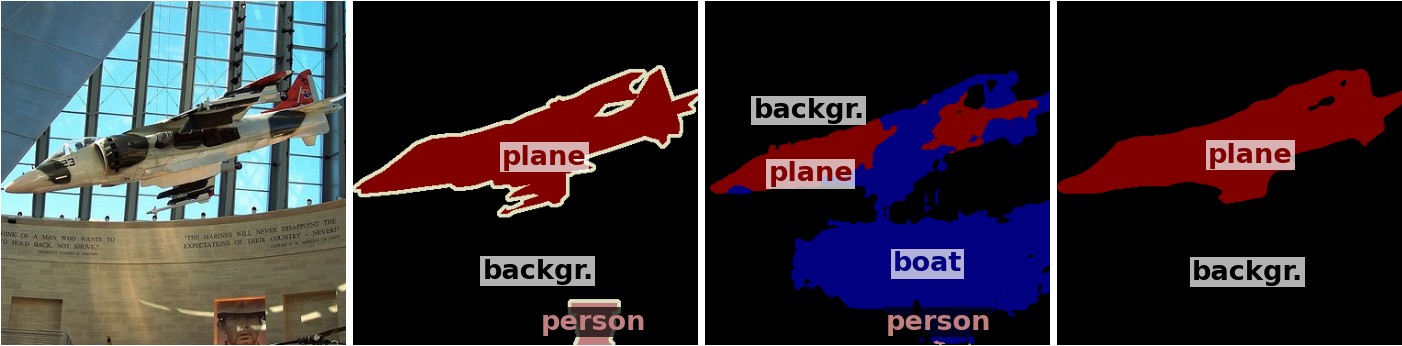}
\includegraphics[width=.48\linewidth]{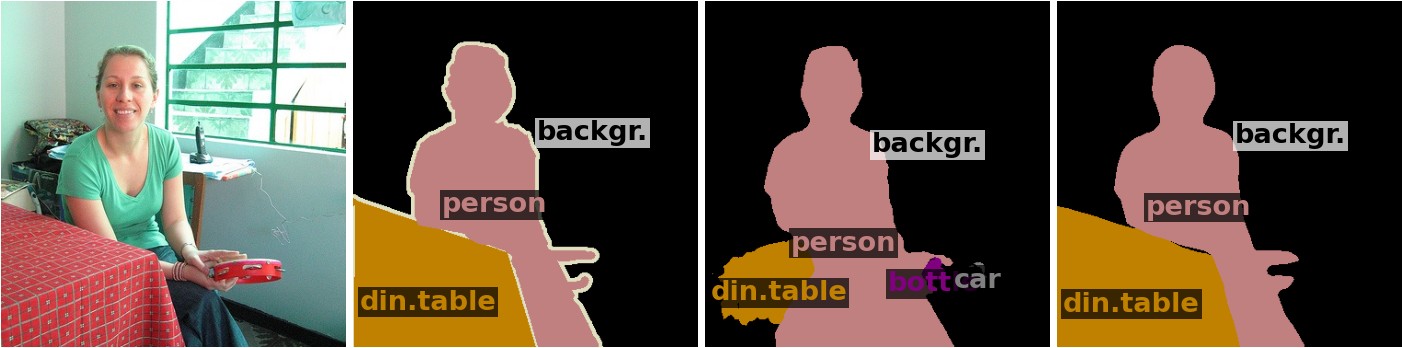}
\includegraphics[width=.48\linewidth]{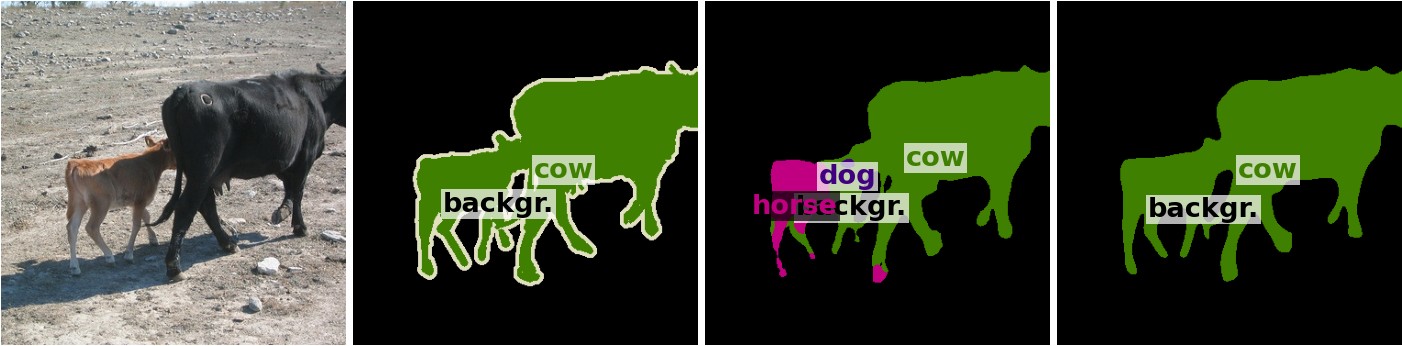}
\includegraphics[width=.48\linewidth]{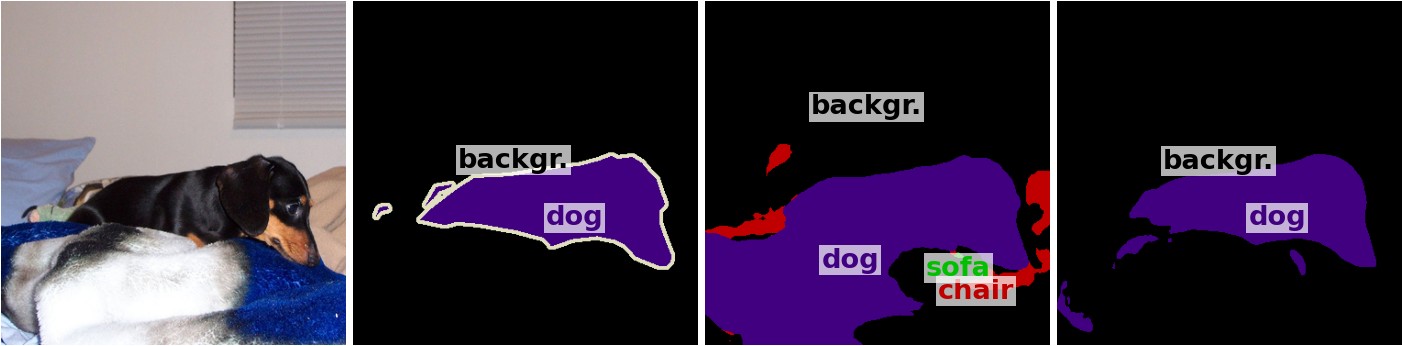}
\includegraphics[width=.48\linewidth]{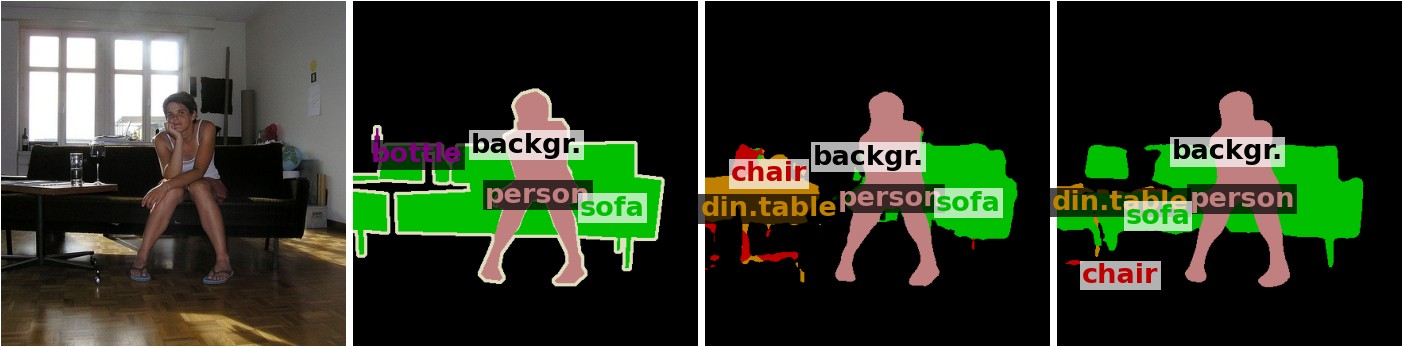}
\includegraphics[width=.48\linewidth]{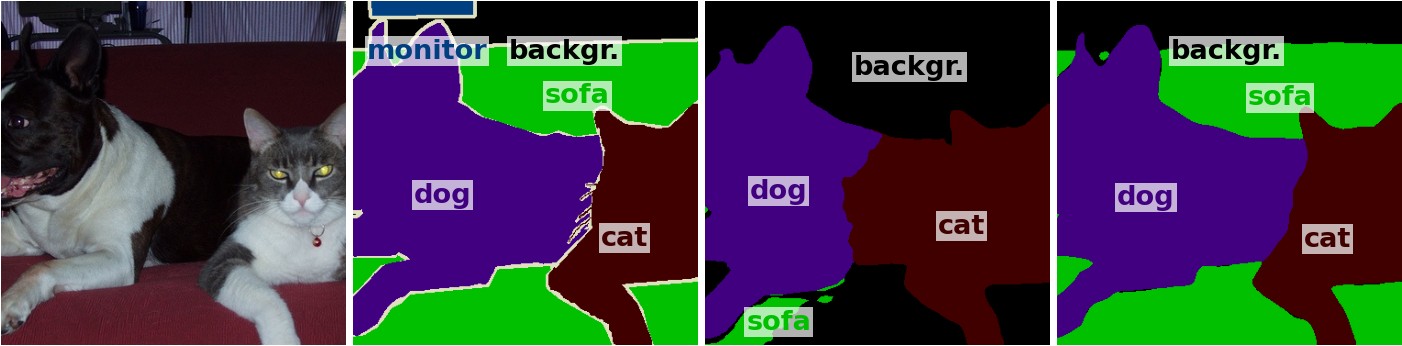}
\includegraphics[width=.48\linewidth]{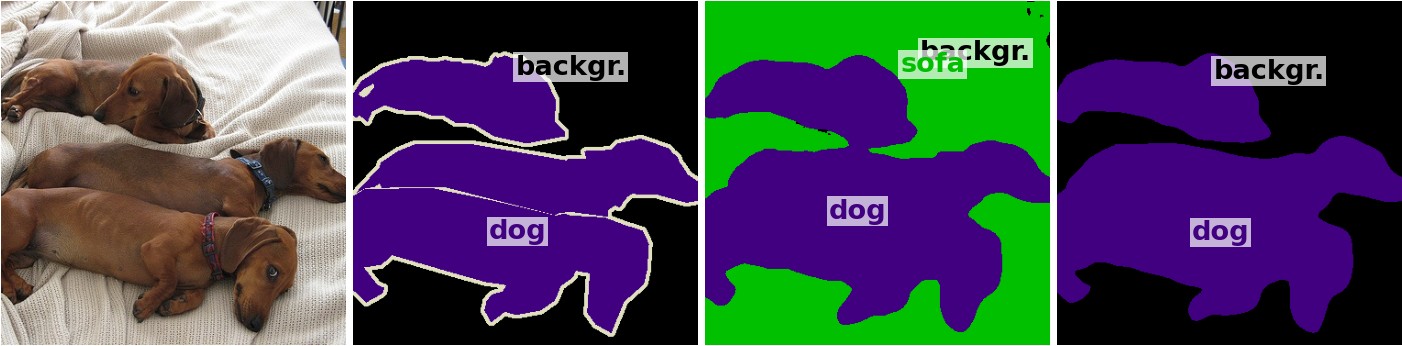}
\caption{\textbf{Example predictions on VOC} (92 labels) showing the improved semantic understanding of \ours. In particular, SemiVL better distinguishes classes with similar visual appearance such as different animals (sheep, cat, dog, cow, and horse), different furniture (chair, dining table, and sofa), and different vehicles (bus, car, boat, airplane, and train). It also improves the distinction of foreground classes versus the background class.}
\label{fig:suppl_examples_voc}
\end{figure*}

\begin{figure*}
\footnotesize
\centering
\begin{tabularx}{.48\linewidth}{*{4}{Y}}
Image & G. Truth & UniMatch~\cite{yang2023revisiting} & \ours\ \\
\end{tabularx} 
\begin{tabularx}{.48\linewidth}{*{4}{Y}}
Image & G. Truth & UniMatch~\cite{yang2023revisiting} & \ours\ \\
\end{tabularx} \\
\includegraphics[width=.48\linewidth]{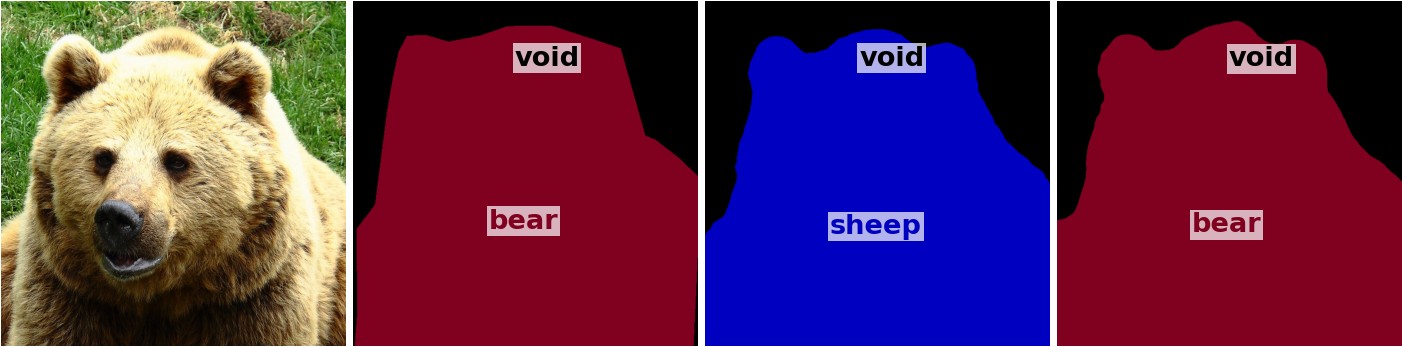}
\includegraphics[width=.48\linewidth]{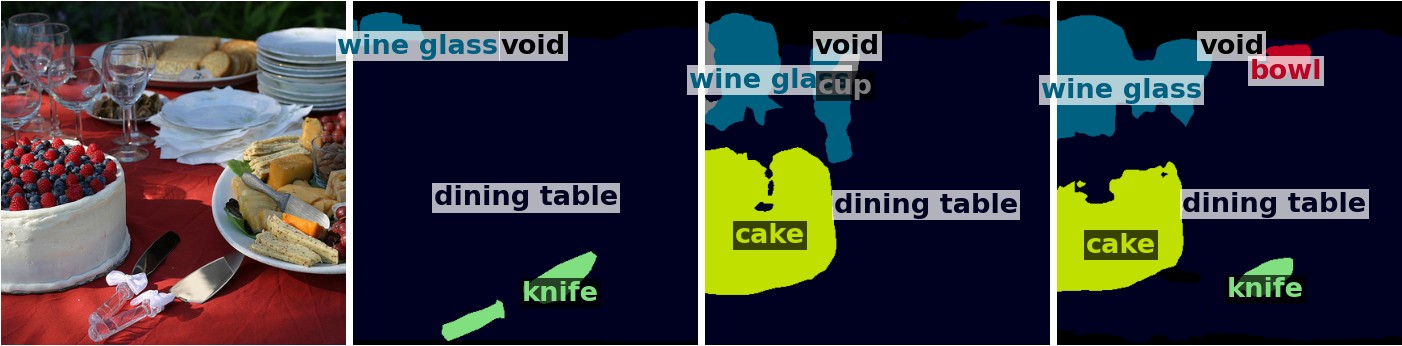}
\includegraphics[width=.48\linewidth]{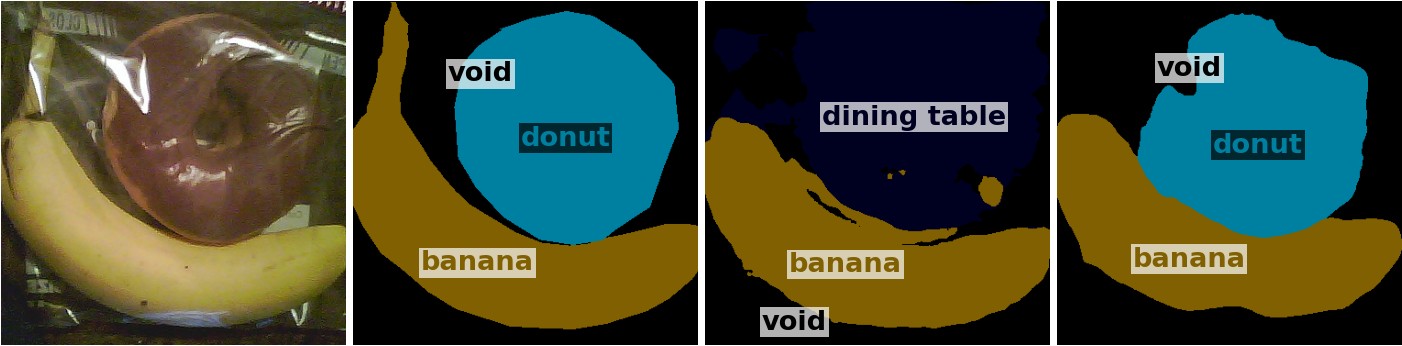}
\includegraphics[width=.48\linewidth]{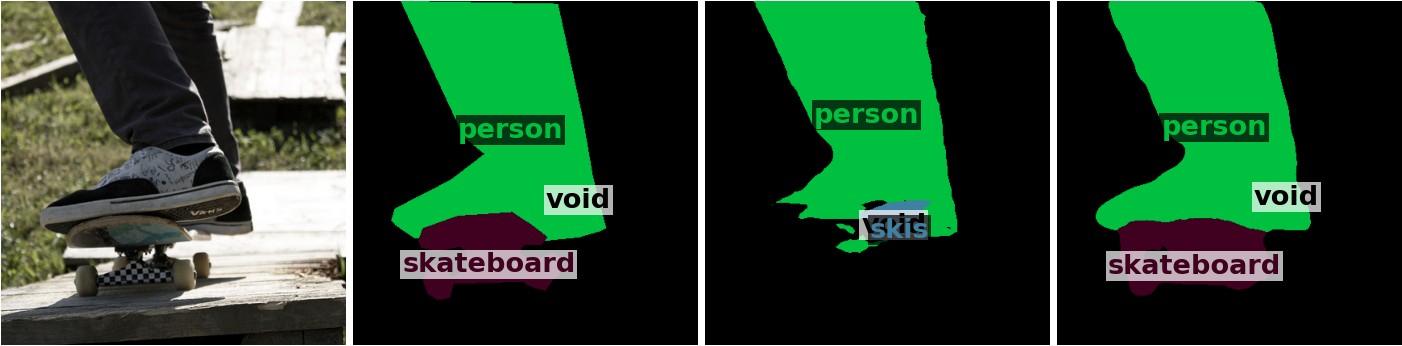}
\includegraphics[width=.48\linewidth]{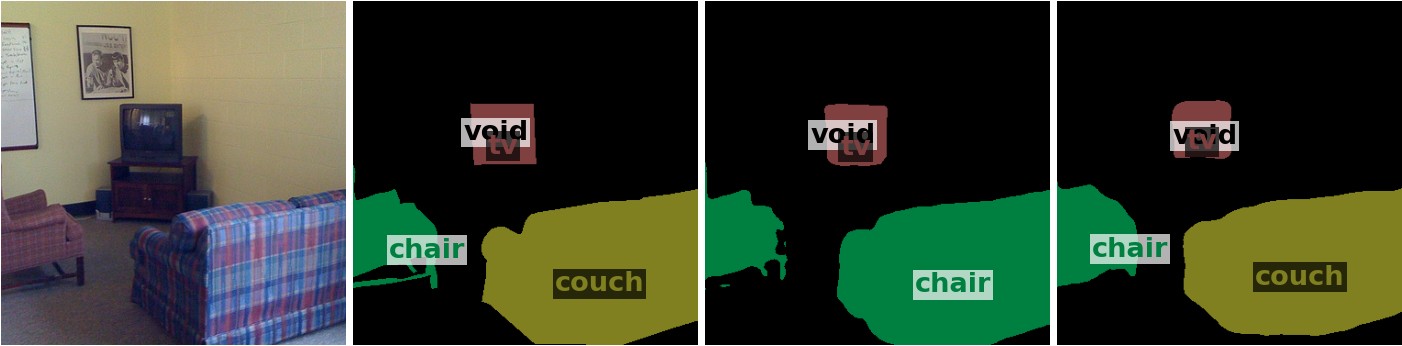}
\includegraphics[width=.48\linewidth]{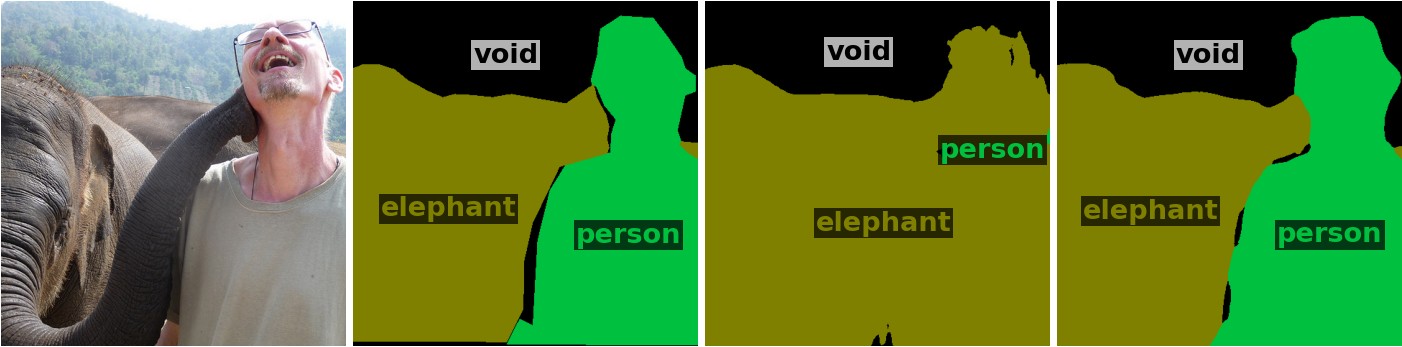}
\includegraphics[width=.48\linewidth]{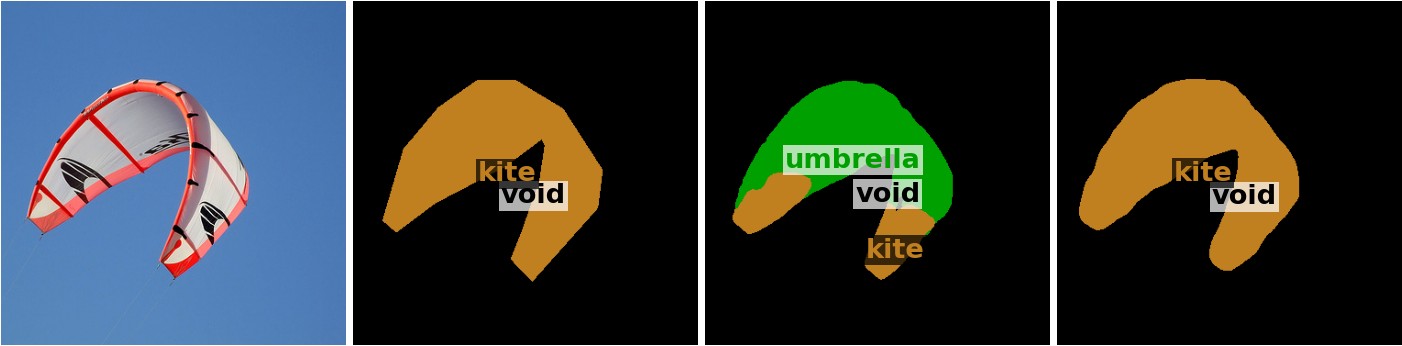}
\includegraphics[width=.48\linewidth]{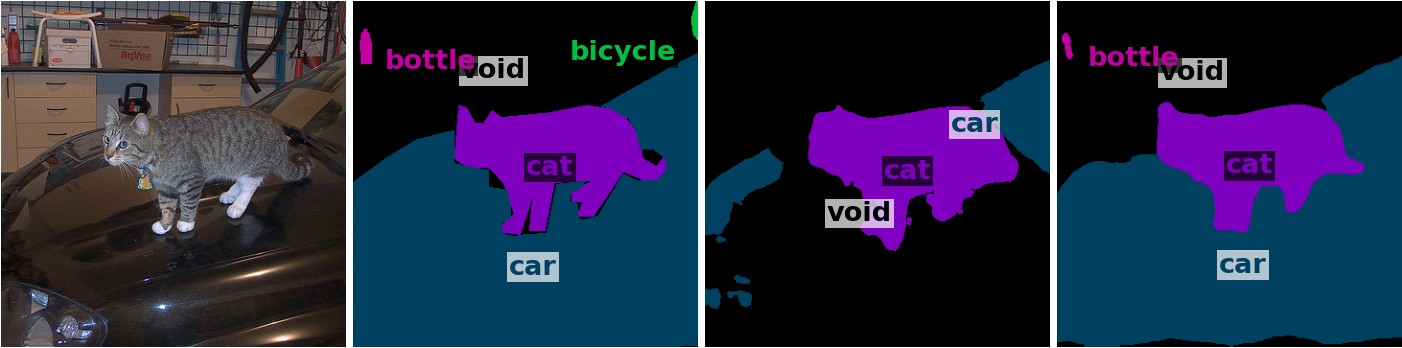}
\includegraphics[width=.48\linewidth]{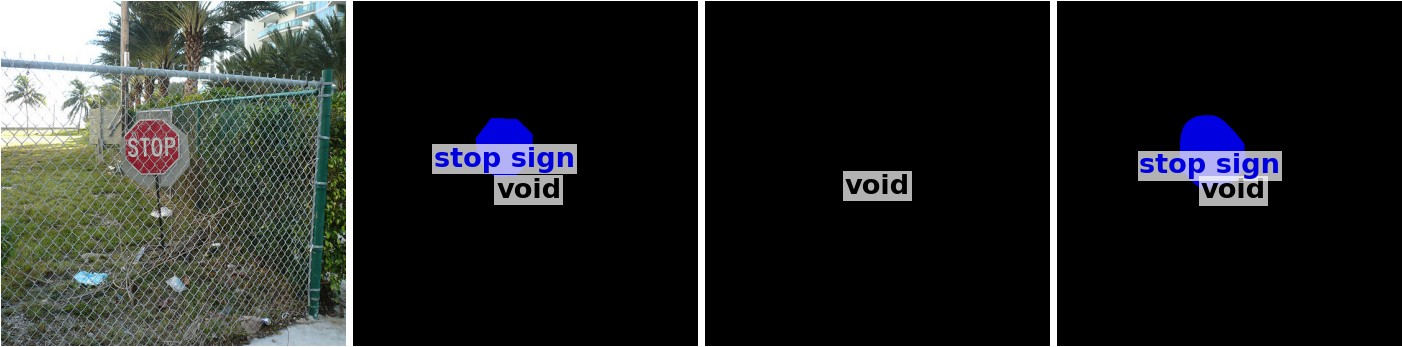}
\includegraphics[width=.48\linewidth]{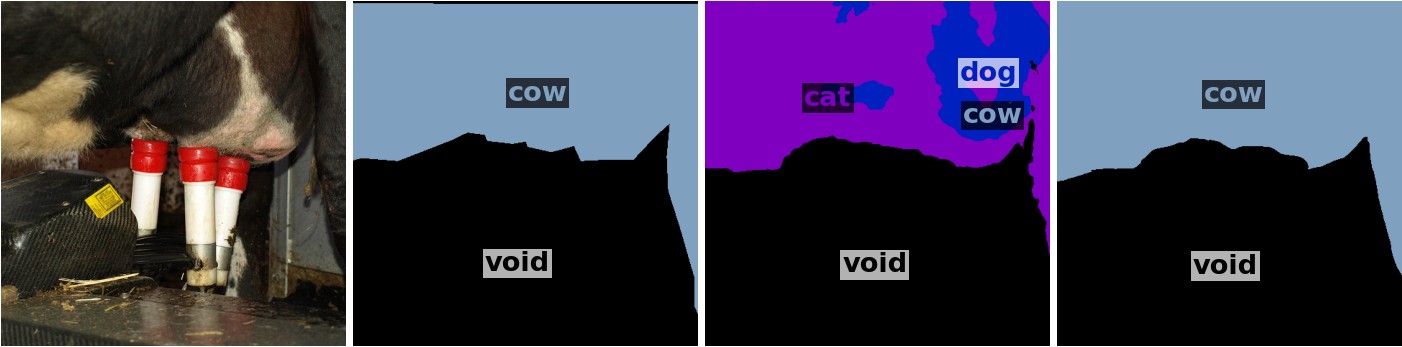}
\includegraphics[width=.48\linewidth]{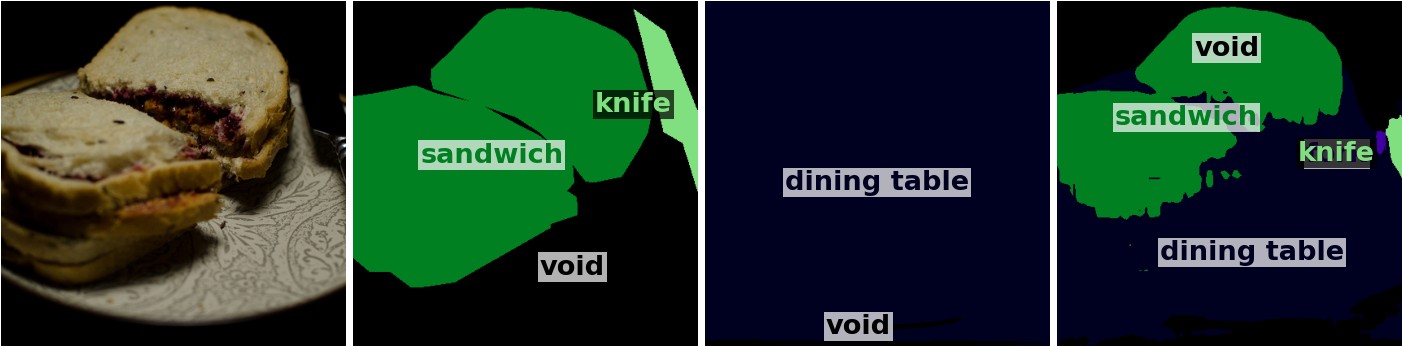}
\includegraphics[width=.48\linewidth]{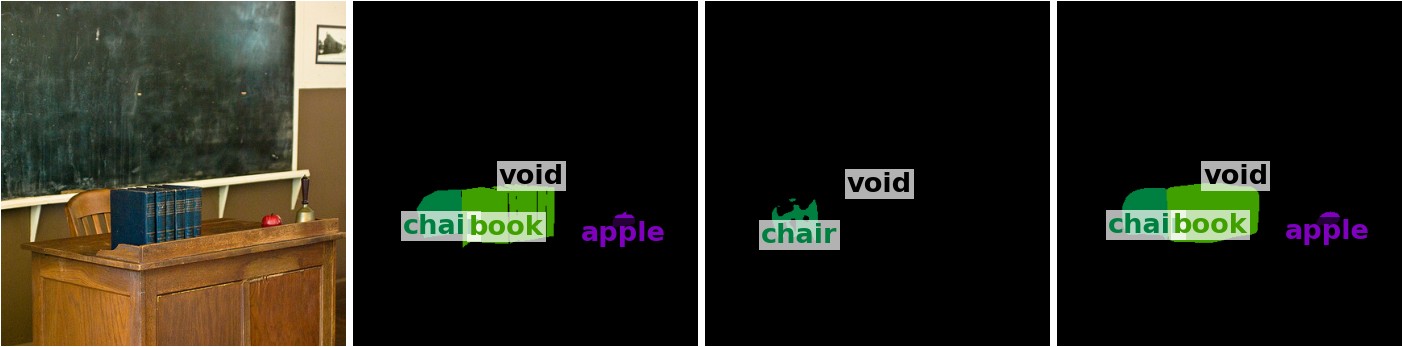}
\includegraphics[width=.48\linewidth]{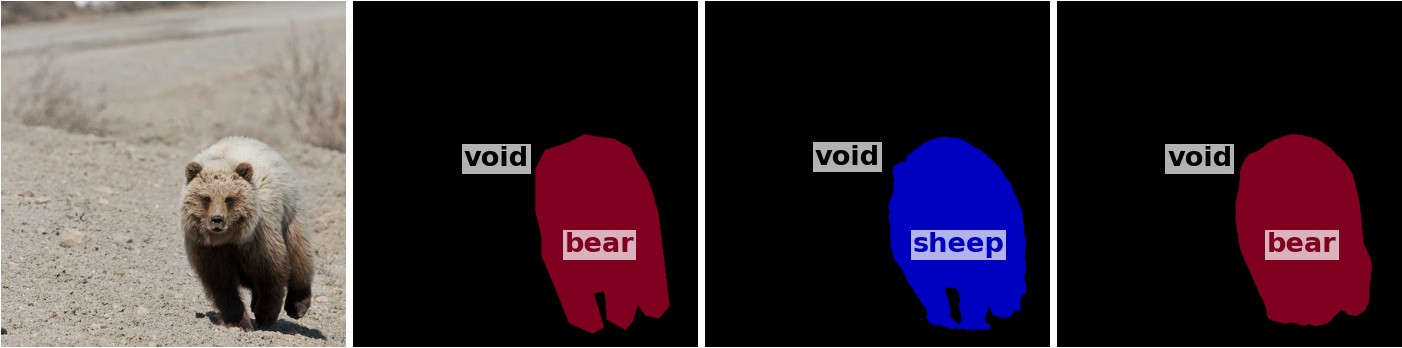}
\includegraphics[width=.48\linewidth]{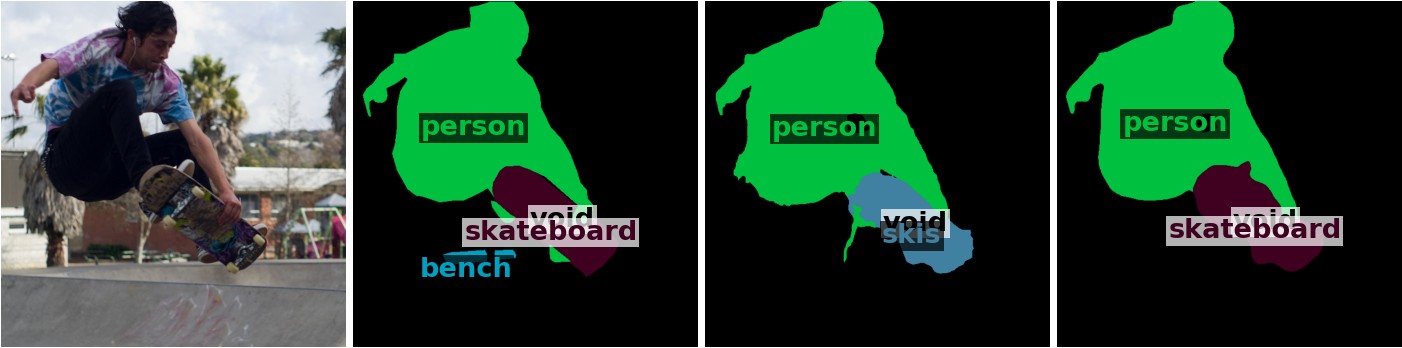}
\includegraphics[width=.48\linewidth]{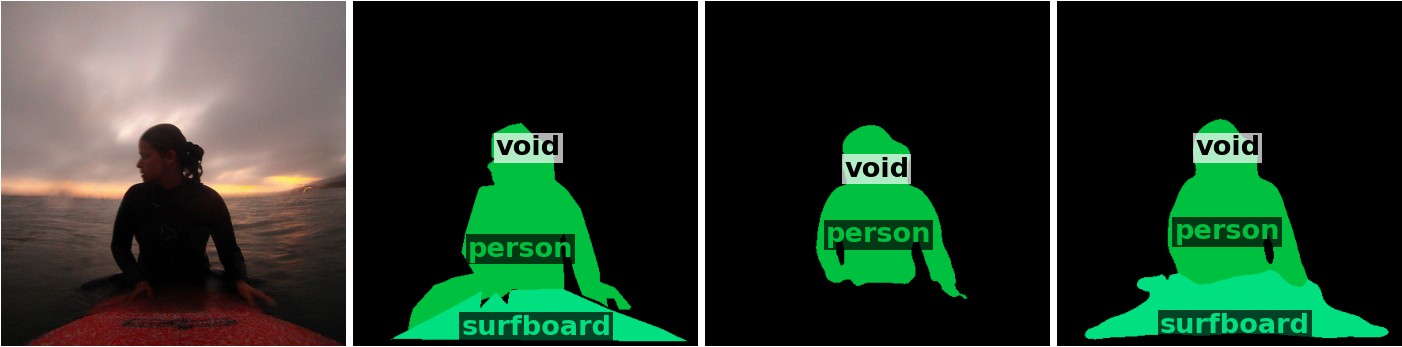}
\includegraphics[width=.48\linewidth]{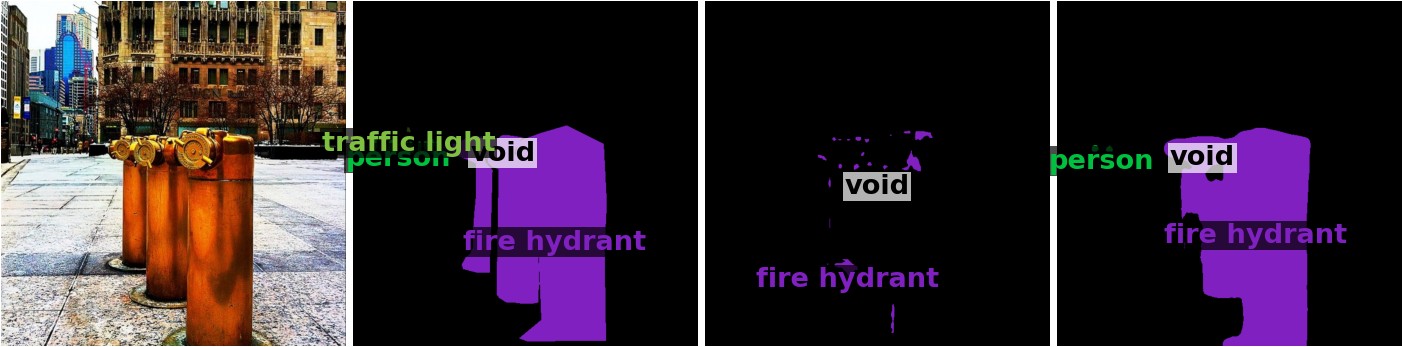}
\includegraphics[width=.48\linewidth]{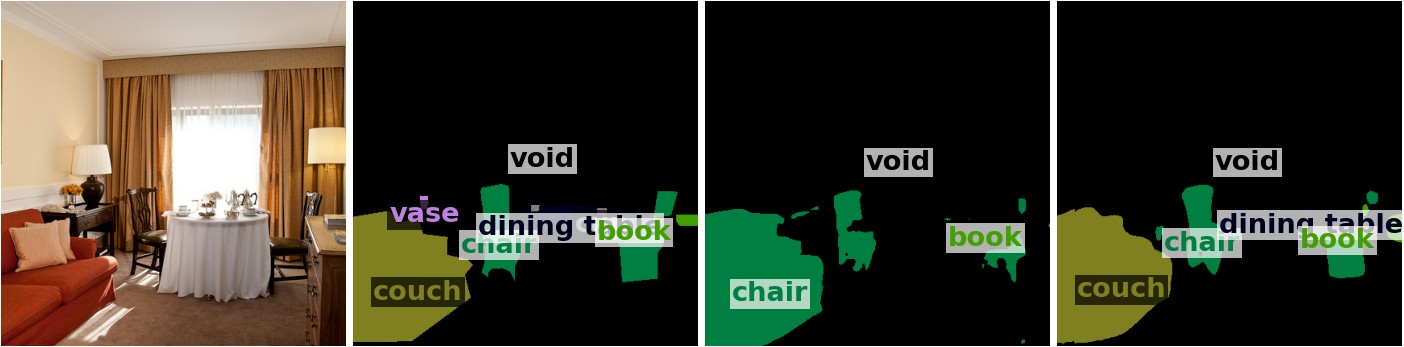}
\includegraphics[width=.48\linewidth]{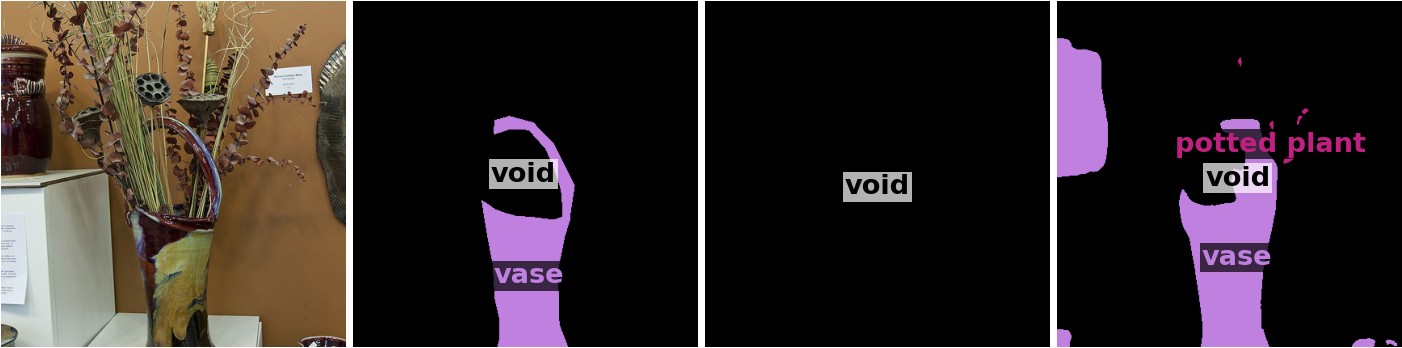}
\caption{\textbf{Example predictions on COCO} (232 labels) showing the improved semantic understanding of \ours. In particular, SemiVL better distinguishes classes with similar visual appearance such as different animals (bear, sheep, cat, and cow), food (cake, donut, sandwich, and apple), furniture (couch, chair, book, and vase), and sports gear (skateboard, skis, kite, umbrella, and surfboard).}
\label{fig:suppl_examples_coco}
\end{figure*}

\begin{figure*}
\footnotesize
\centering
\begin{tabularx}{.48\linewidth}{*{4}{Y}}
Image & G. Truth & UniMatch~\cite{yang2023revisiting} & \ours\ \\
\end{tabularx} 
\begin{tabularx}{.48\linewidth}{*{4}{Y}}
Image & G. Truth & UniMatch~\cite{yang2023revisiting} & \ours\ \\
\end{tabularx} \\
\includegraphics[width=.48\linewidth]{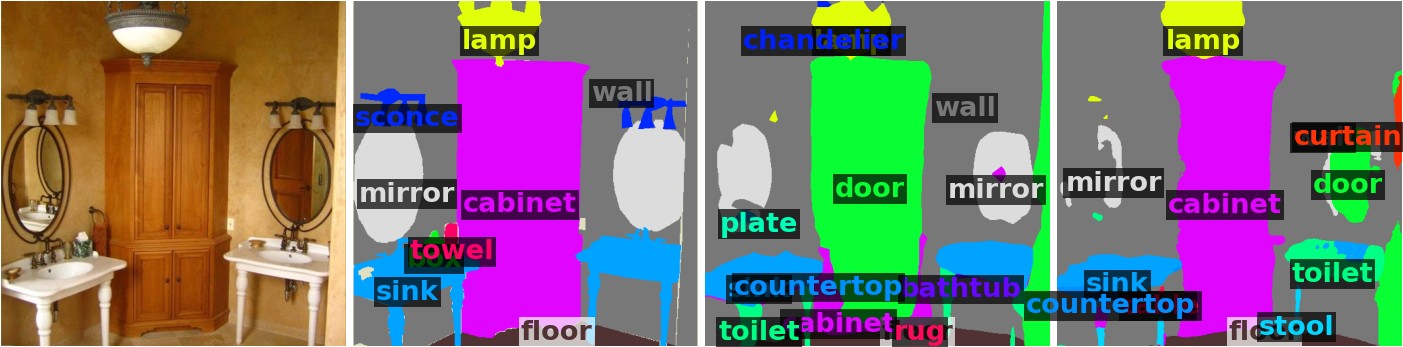}
\includegraphics[width=.48\linewidth]{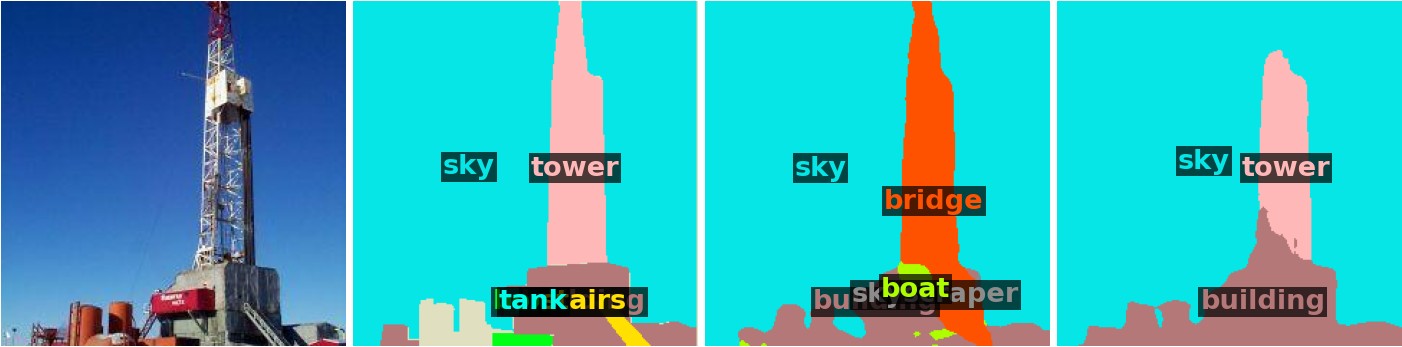}
\includegraphics[width=.48\linewidth]{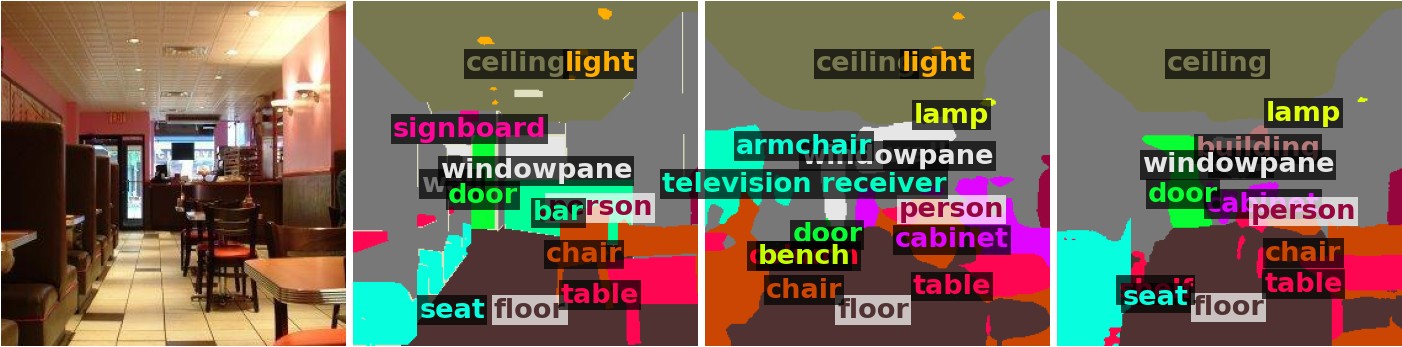}
\includegraphics[width=.48\linewidth]{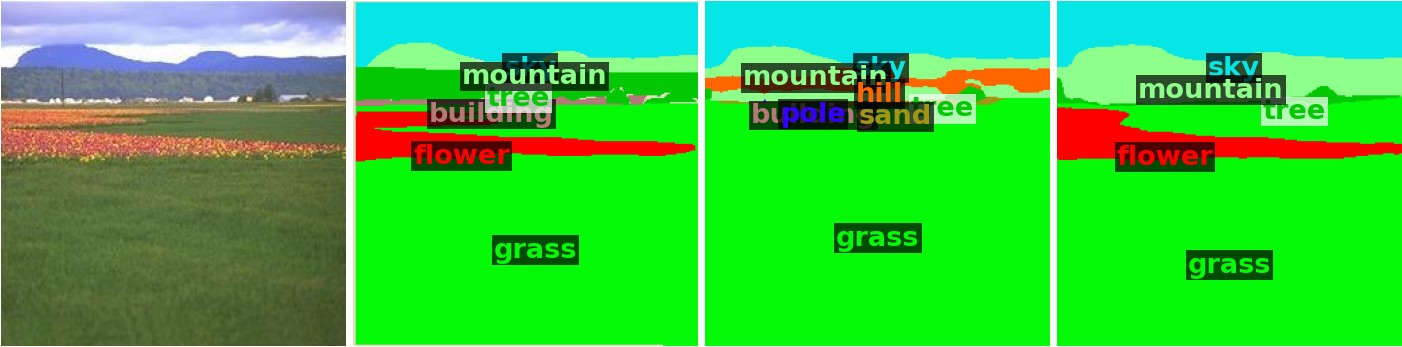}
\includegraphics[width=.48\linewidth]{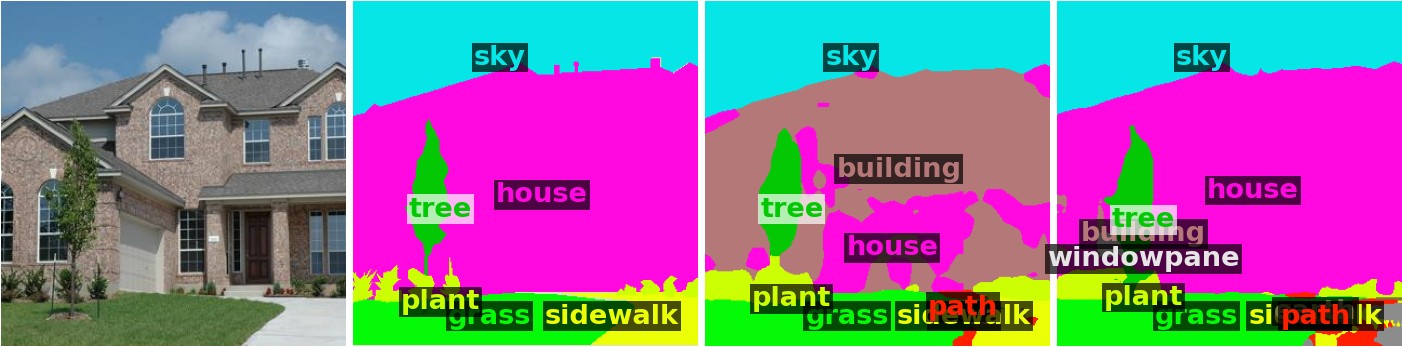}
\includegraphics[width=.48\linewidth]{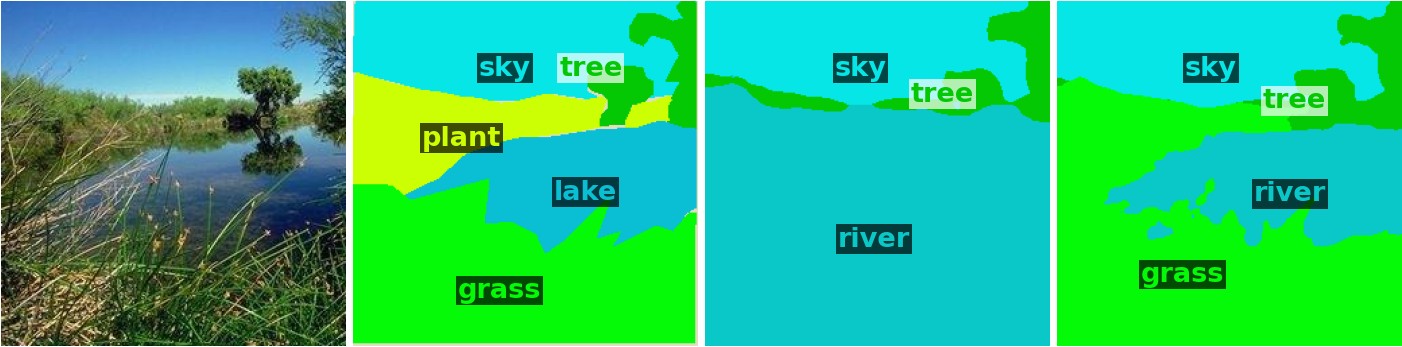}
\includegraphics[width=.48\linewidth]{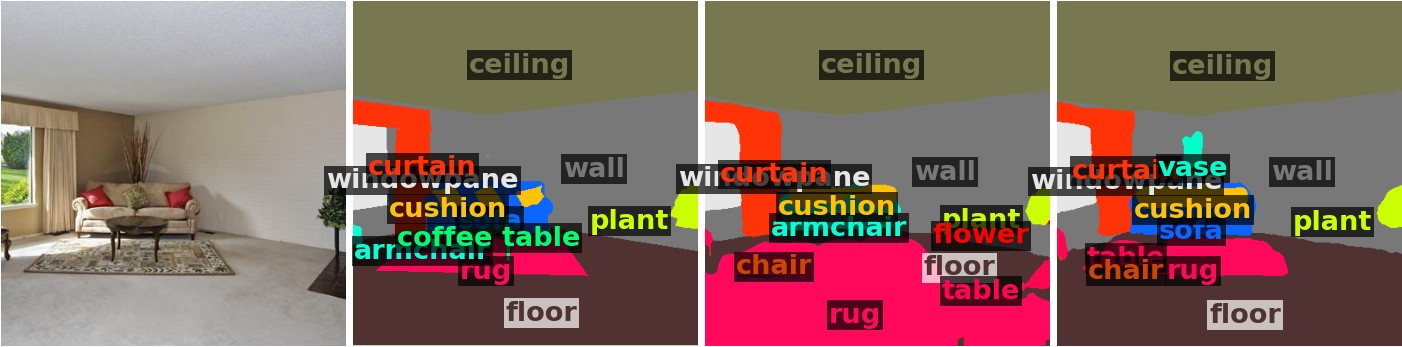}
\includegraphics[width=.48\linewidth]{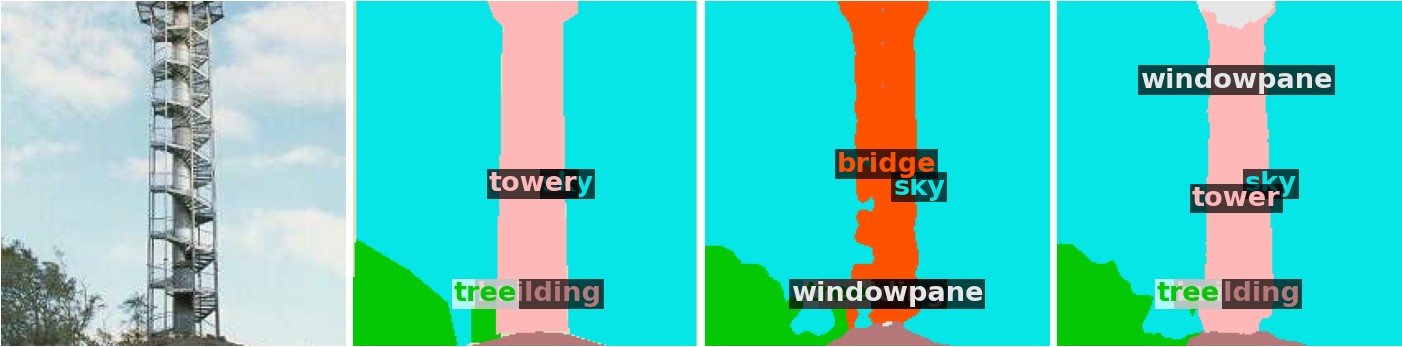}
\includegraphics[width=.48\linewidth]{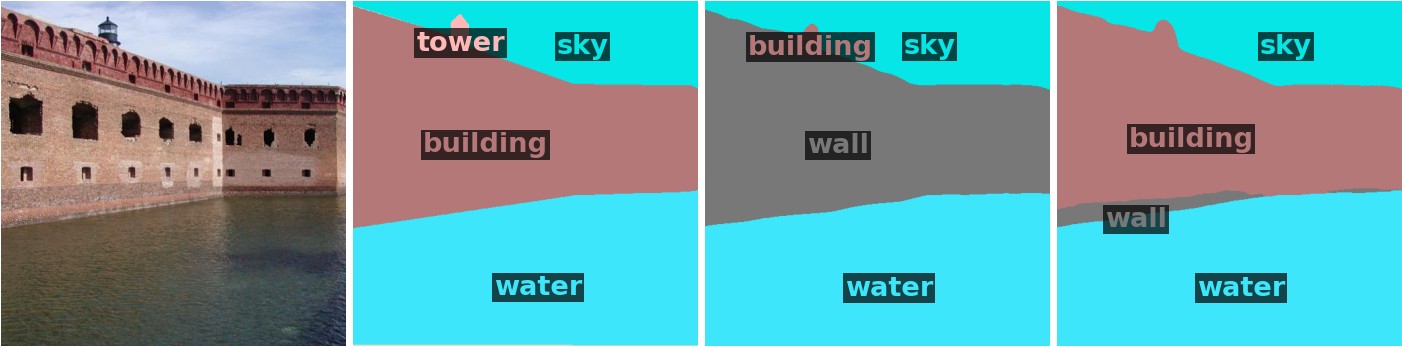}
\includegraphics[width=.48\linewidth]{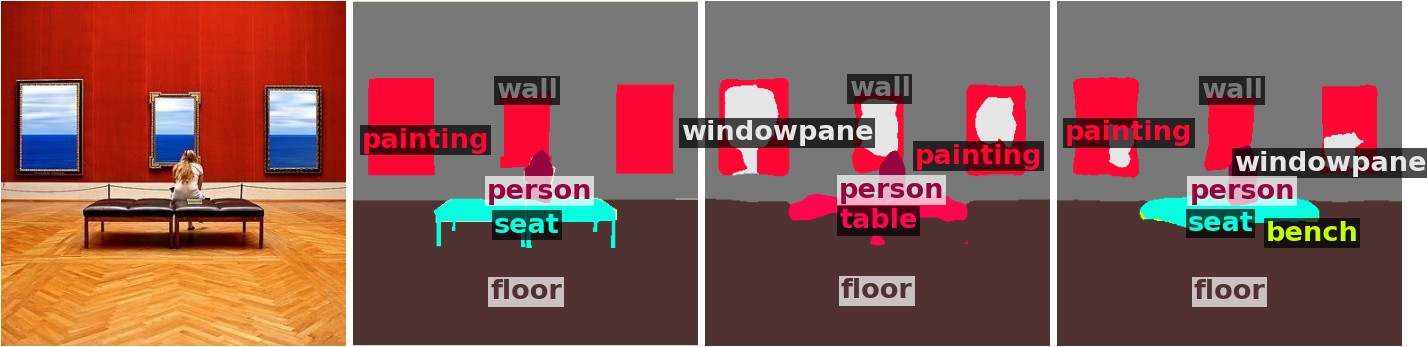}
\includegraphics[width=.48\linewidth]{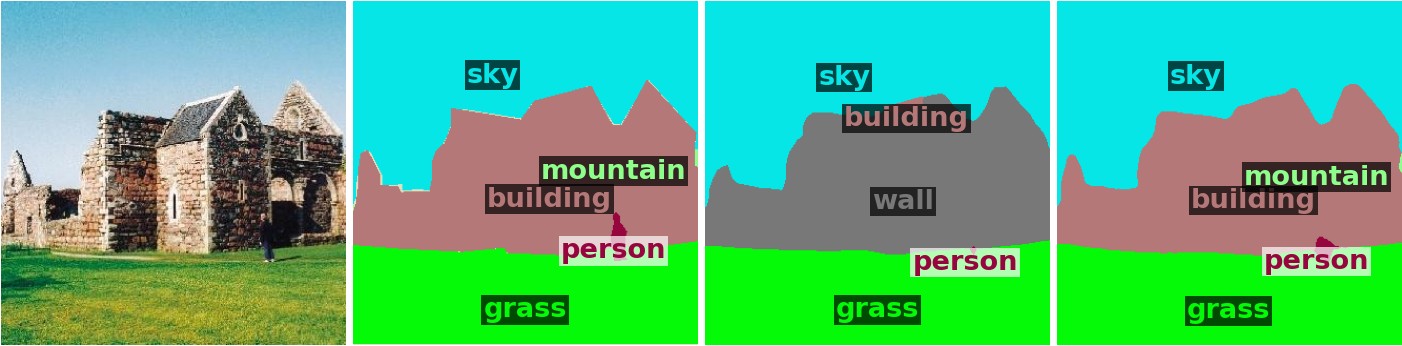}
\includegraphics[width=.48\linewidth]{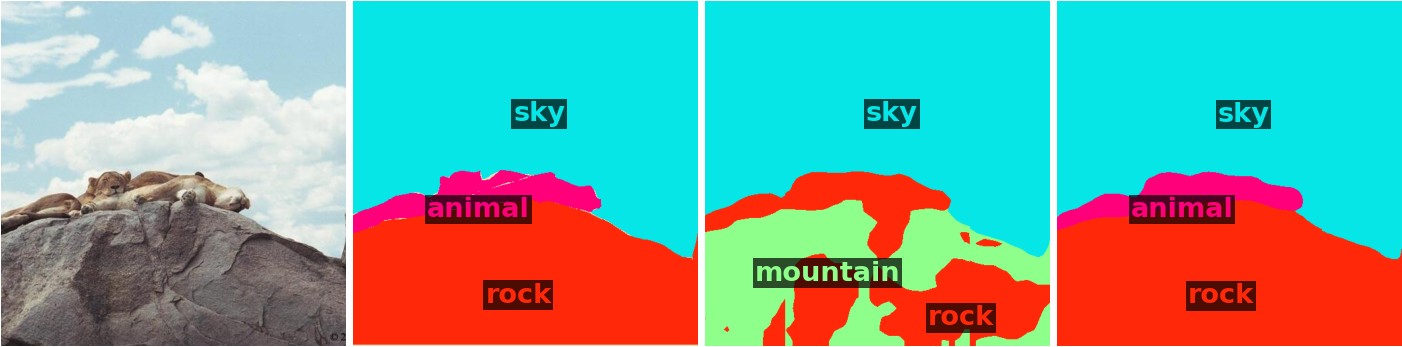}
\includegraphics[width=.48\linewidth]{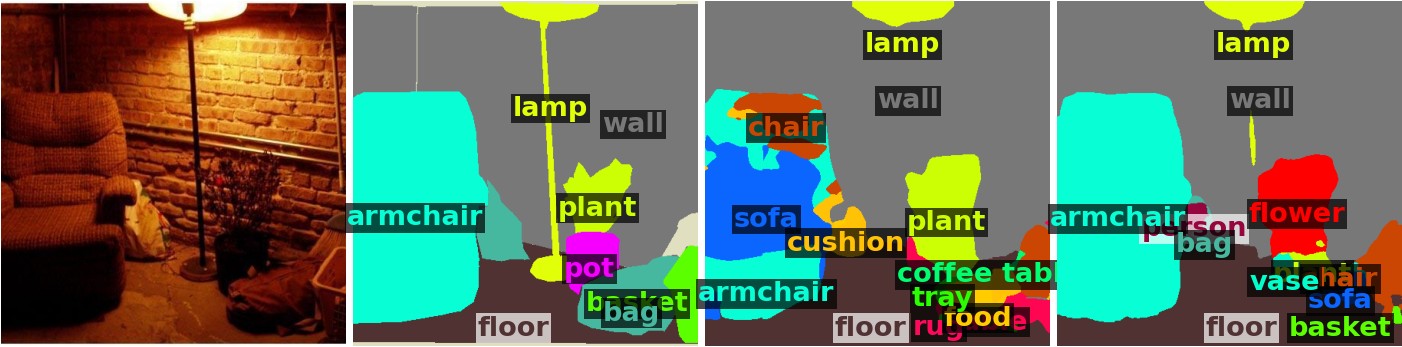}
\includegraphics[width=.48\linewidth]{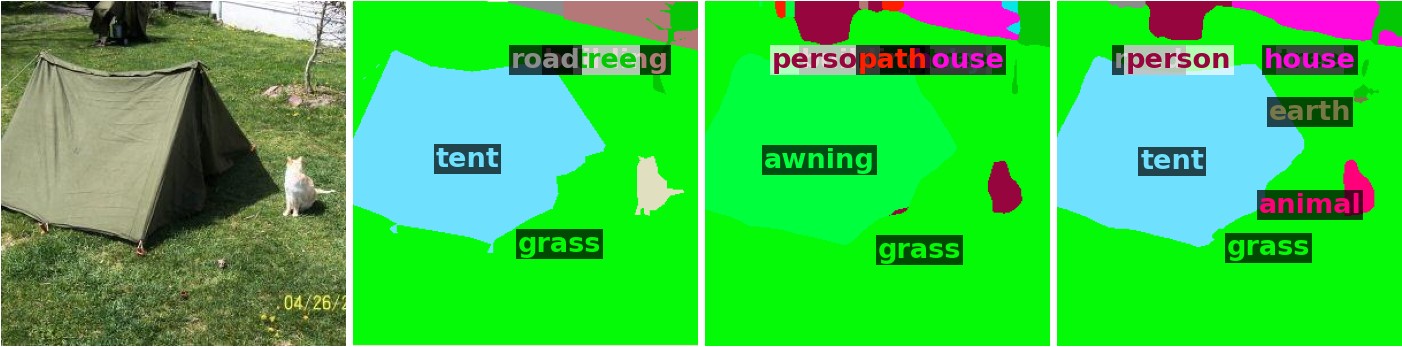}
\includegraphics[width=.48\linewidth]{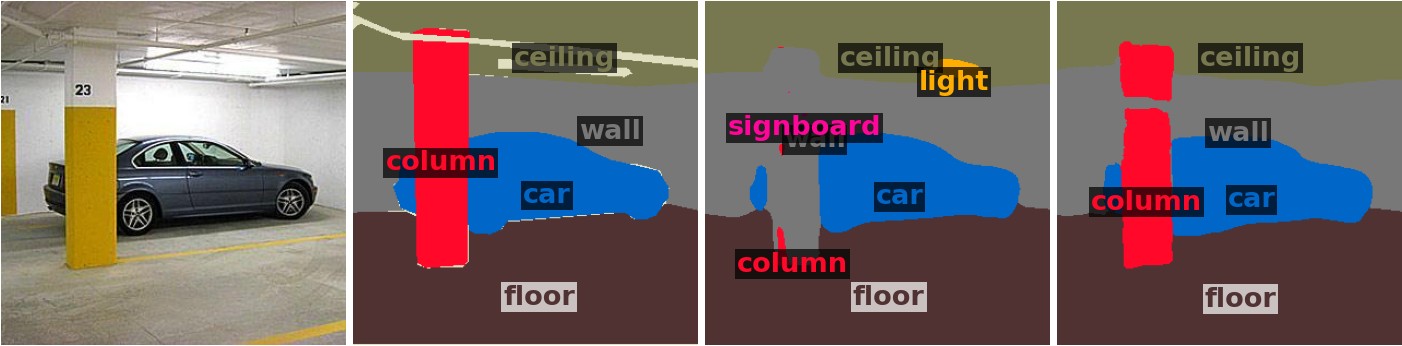}
\includegraphics[width=.48\linewidth]{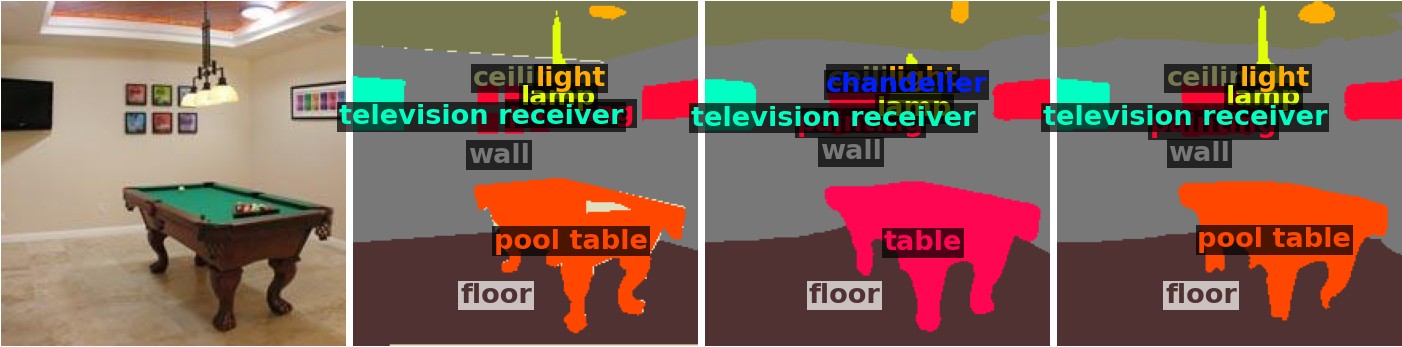}
\includegraphics[width=.48\linewidth]{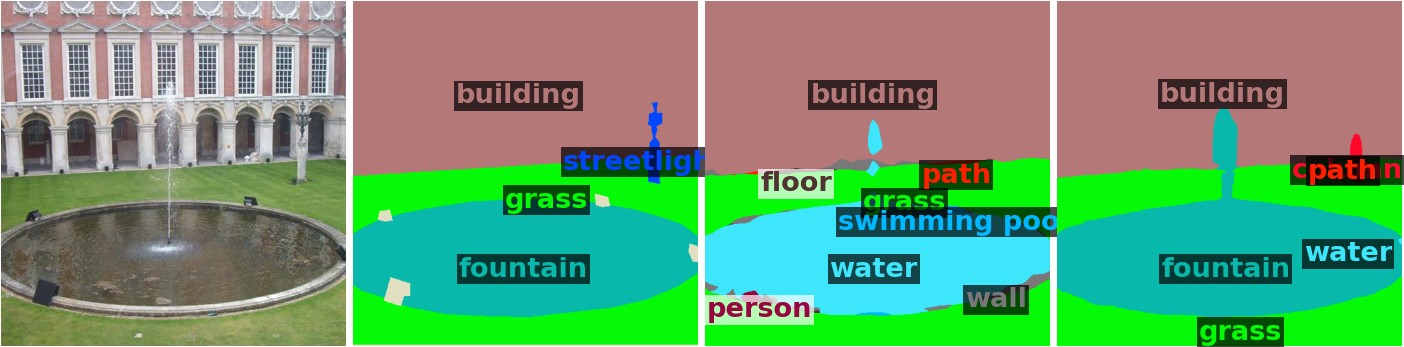}
\includegraphics[width=.48\linewidth]{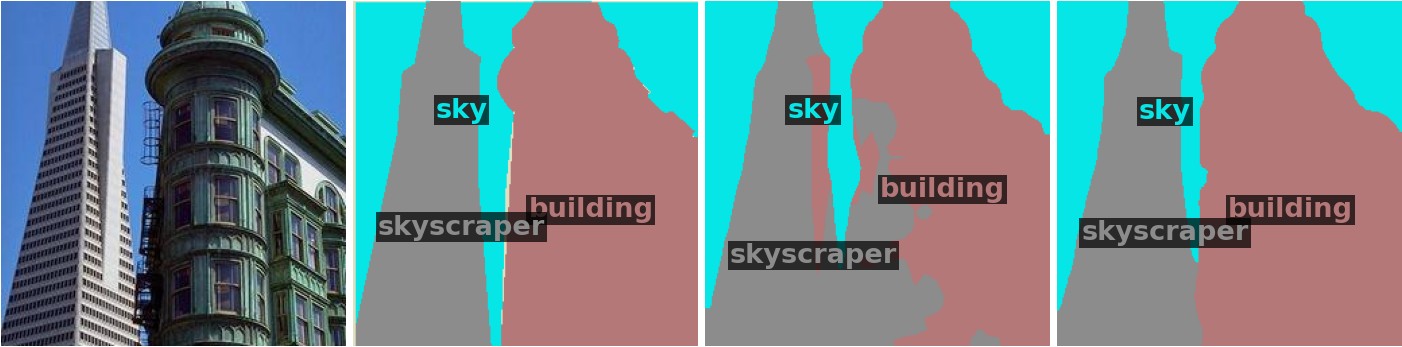}
\caption{\textbf{Example predictions on ADE20K} (158 labels) showing the improved semantic understanding of \ours. In particular, SemiVL better distinguishes classes with similar visual appearance such as different structures (tower, bridge, building, house, skyscraper, column, and wall), different furniture (cabinet, door, chair, seat, table, sofa, and pool table), and ground types (rock, mountain, water, fountain, floor, rug, grass, and river).}
\label{fig:suppl_examples_ade}
\end{figure*}

\begin{figure*}
\footnotesize
\centering
\begin{tabularx}{.48\linewidth}{*{4}{Y}}
Image & G. Truth & UniMatch~\cite{yang2023revisiting} & \ours\ \\
\end{tabularx} 
\begin{tabularx}{.48\linewidth}{*{4}{Y}}
Image & G. Truth & UniMatch~\cite{yang2023revisiting} & \ours\ \\
\end{tabularx} \\
\includegraphics[width=.48\linewidth]{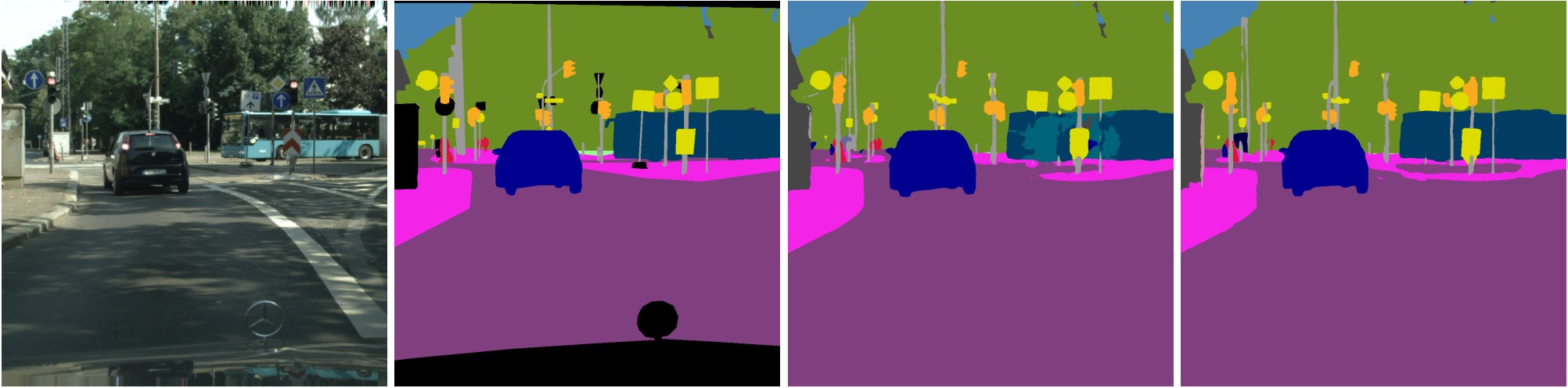}
\includegraphics[width=.48\linewidth]{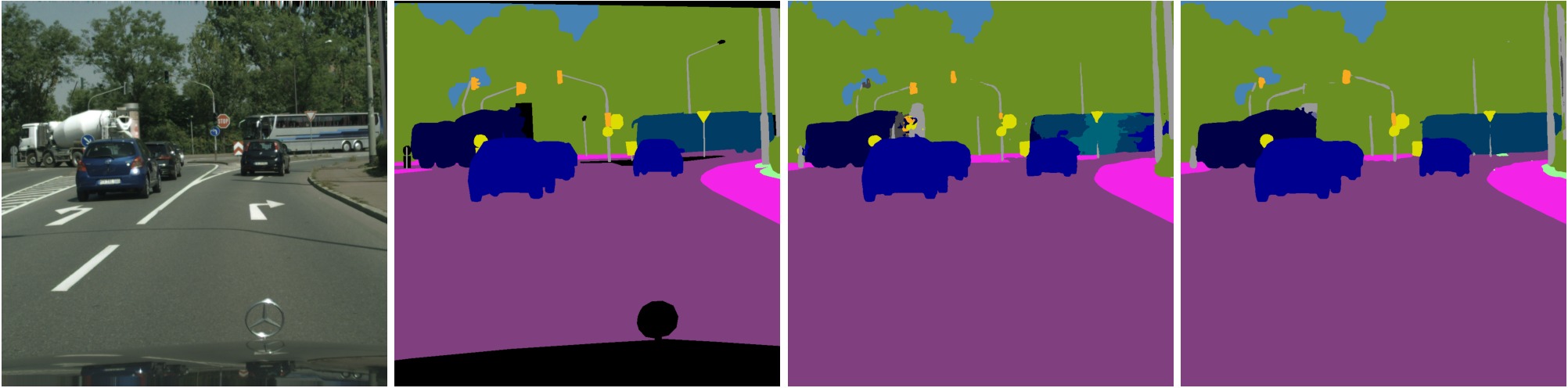}
\includegraphics[width=.48\linewidth]{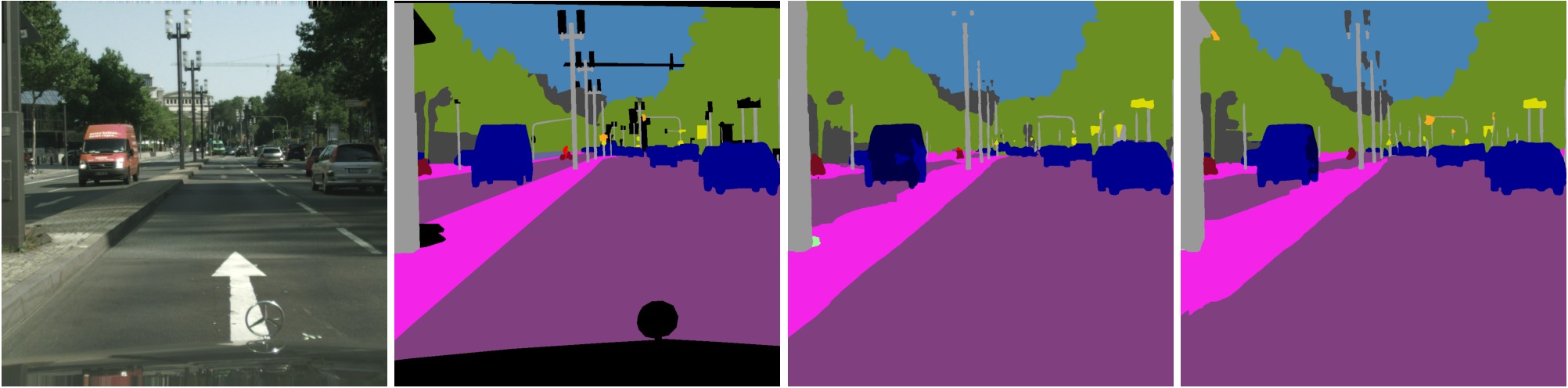}
\includegraphics[width=.48\linewidth]{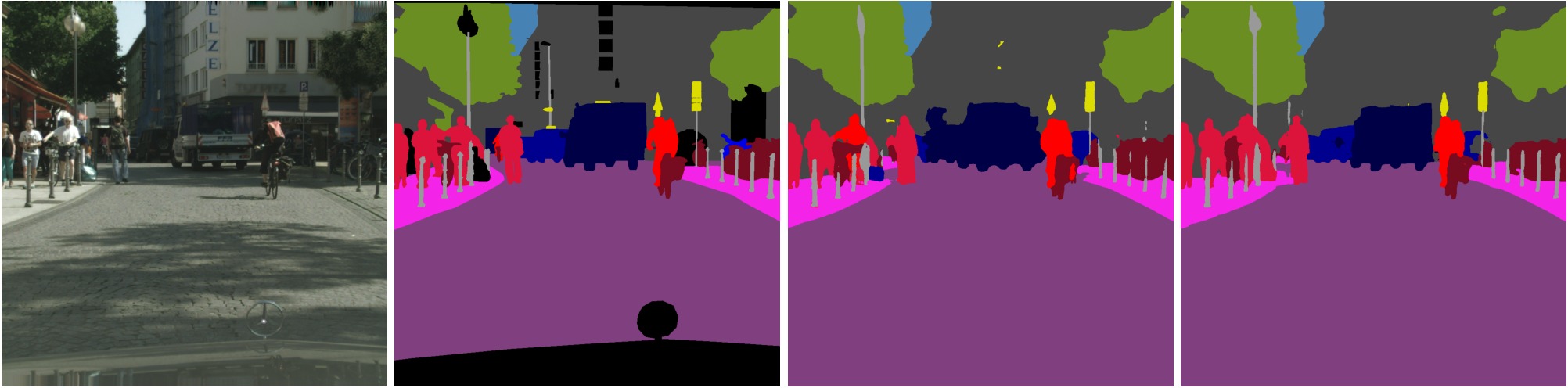}
\includegraphics[width=.48\linewidth]{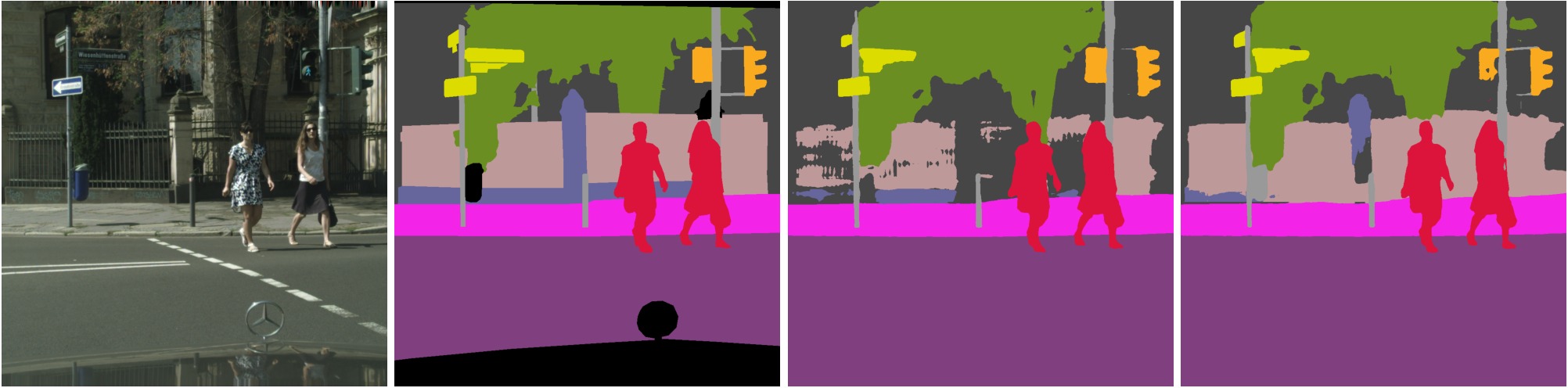}
\includegraphics[width=.48\linewidth]{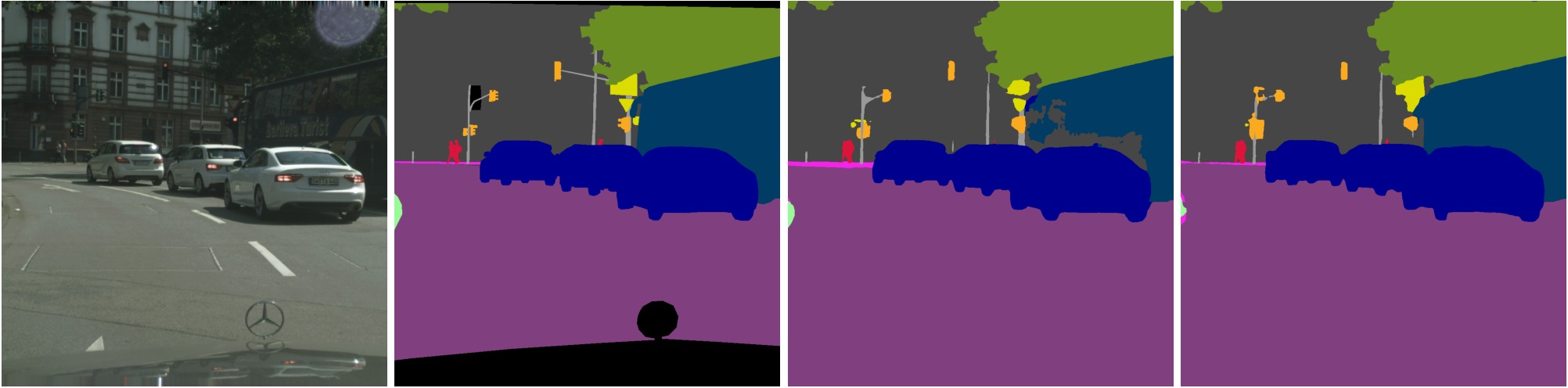}
\includegraphics[width=.48\linewidth]{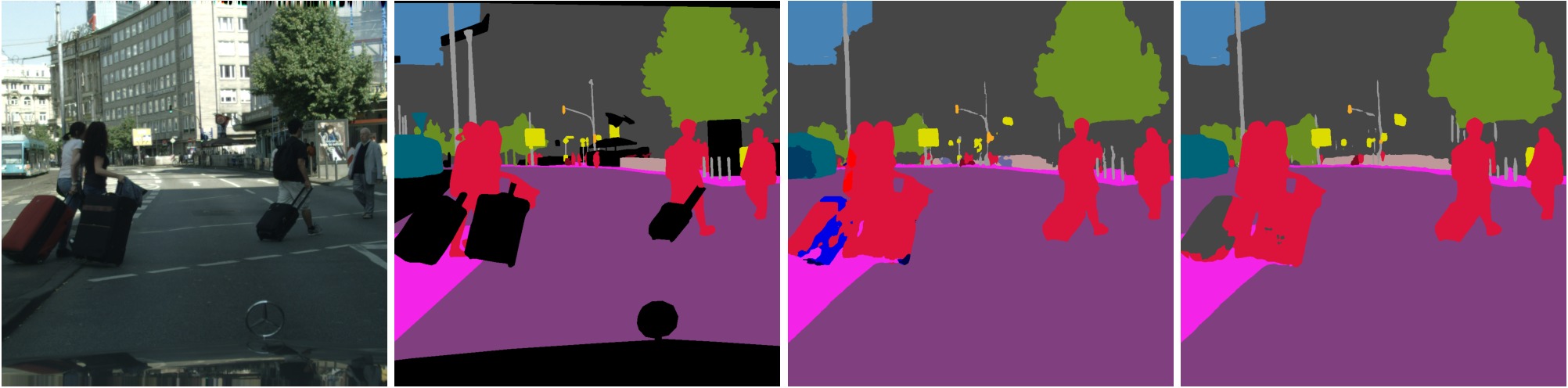}
\includegraphics[width=.48\linewidth]{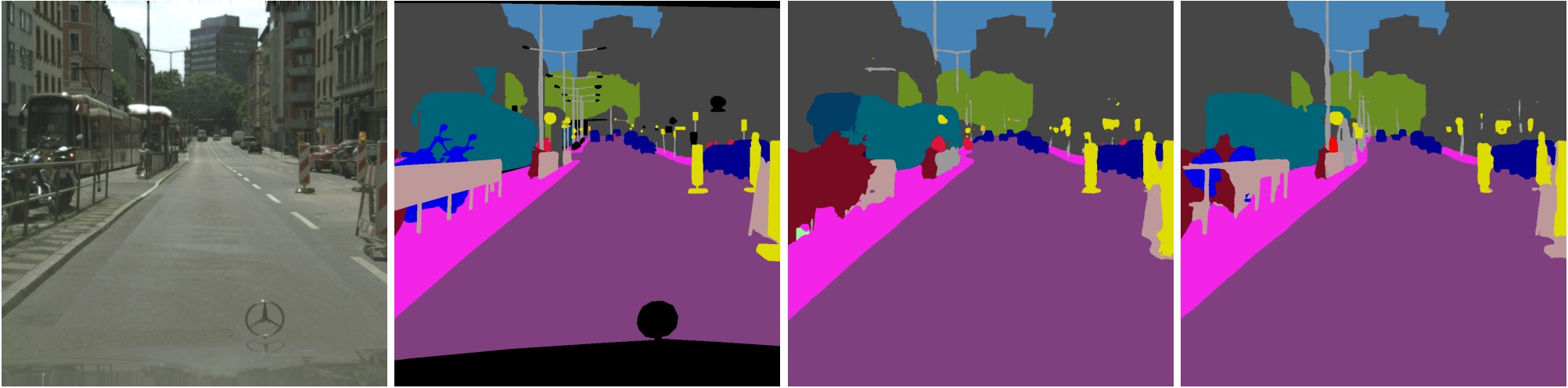}
\includegraphics[width=.48\linewidth]{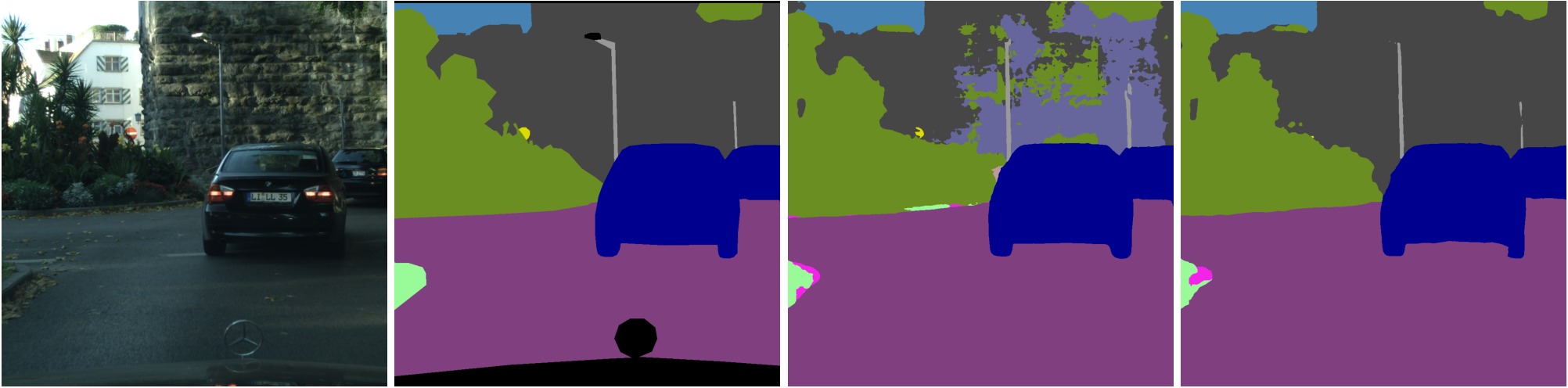}
\includegraphics[width=.48\linewidth]{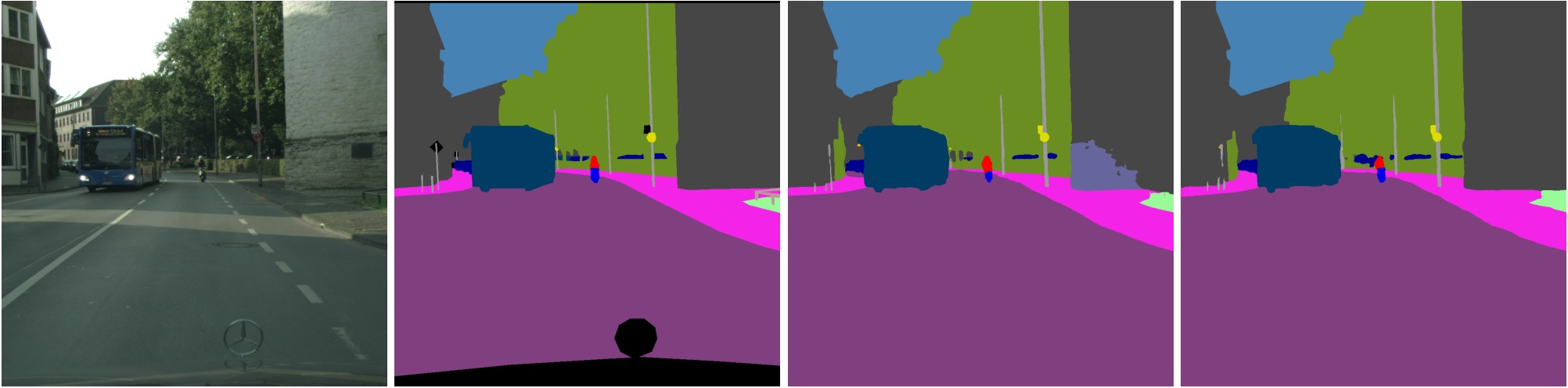}
\includegraphics[width=.48\linewidth]{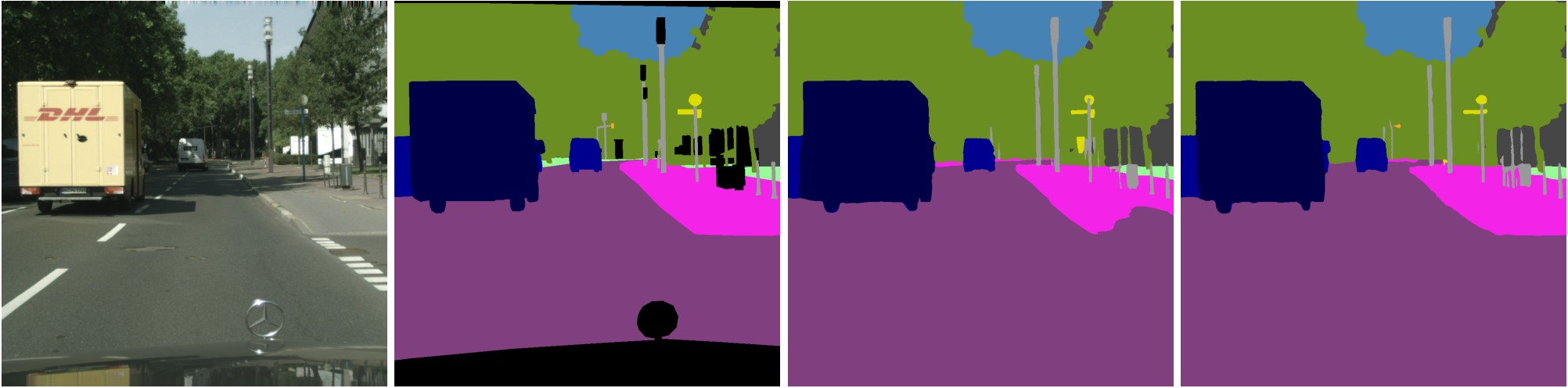}
\includegraphics[width=.48\linewidth]{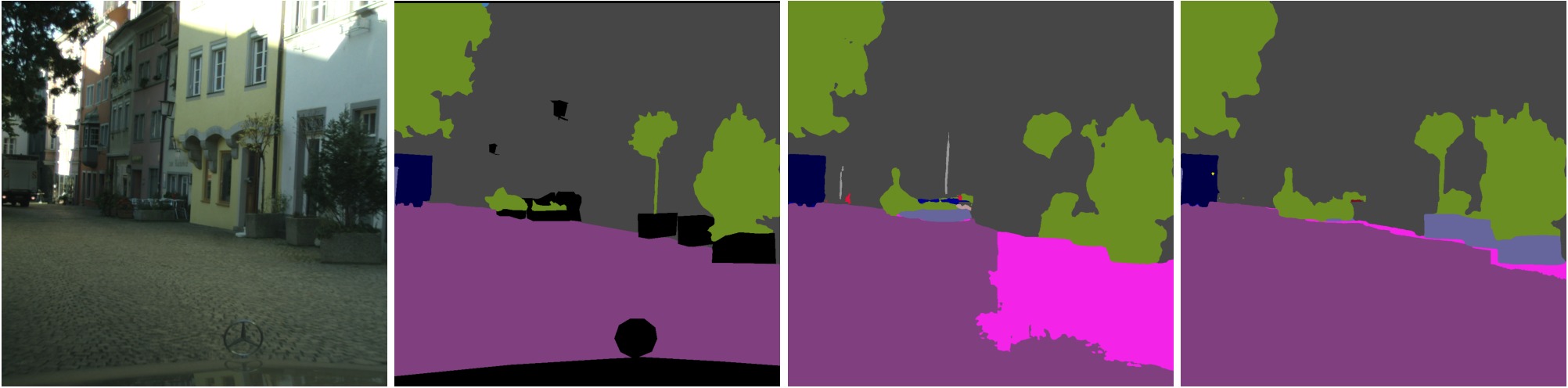}
\includegraphics[width=.48\linewidth]{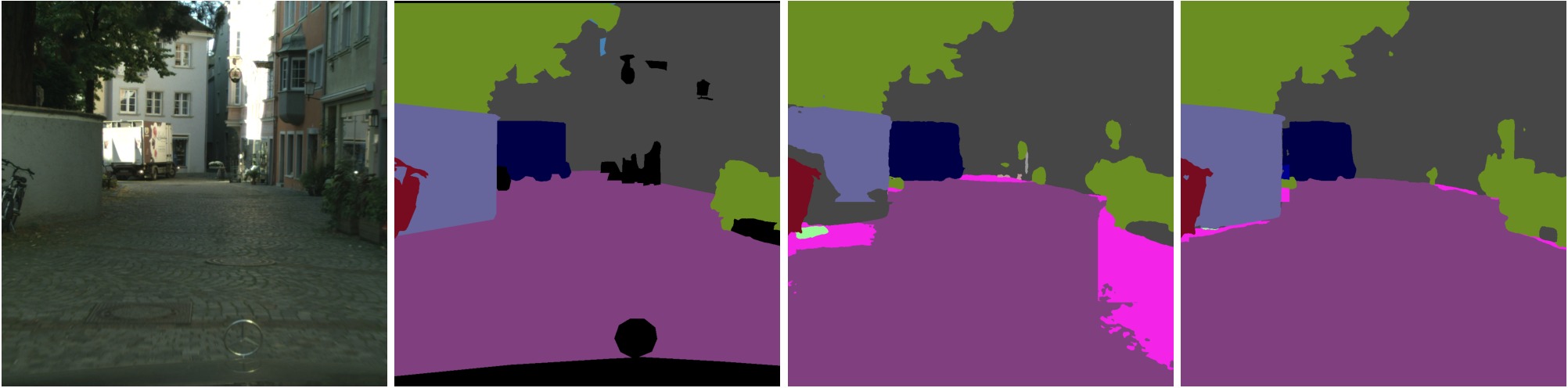}
\includegraphics[width=.48\linewidth]{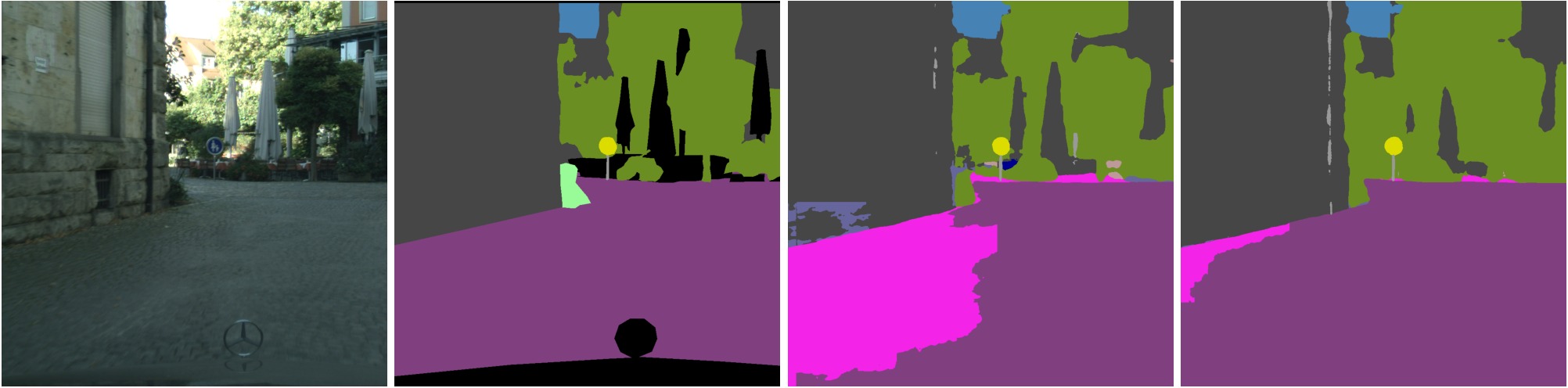}
\scriptsize%
\setlength\tabcolsep{1pt}%
{%
\newcolumntype{P}[1]{>{\centering\arraybackslash}p{#1}}
\begin{tabular}{@{}*{20}{P{0.09\columnwidth}}@{}}
     {\cellcolor[rgb]{0.5,0.25,0.5}}\textcolor{white}{road} 
     &{\cellcolor[rgb]{0.957,0.137,0.91}}sidew. 
     &{\cellcolor[rgb]{0.275,0.275,0.275}}\textcolor{white}{build.} 
     &{\cellcolor[rgb]{0.4,0.4,0.612}}\textcolor{white}{wall} 
     &{\cellcolor[rgb]{0.745,0.6,0.6}}fence 
     &{\cellcolor[rgb]{0.6,0.6,0.6}}pole 
     &{\cellcolor[rgb]{0.98,0.667,0.118}}tr. light
     &{\cellcolor[rgb]{0.863,0.863,0}}tr. sign 
     &{\cellcolor[rgb]{0.42,0.557,0.137}}veget. 
     &{\cellcolor[rgb]{0.596,0.984,0.596}}terrain 
     &{\cellcolor[rgb]{0.275,0.510,0.706}}sky
     &{\cellcolor[rgb]{0.863,0.078,0.235}}\textcolor{white}{person} 
     &{\cellcolor[rgb]{0.988,0.494,0.635}}\textcolor{black}{rider} 
     &{\cellcolor[rgb]{0,0,0.557}}\textcolor{white}{car} 
     &{\cellcolor[rgb]{0,0,0.275}}\textcolor{white}{truck} 
     &{\cellcolor[rgb]{0,0.235,0.392}}\textcolor{white}{bus}
     &{\cellcolor[rgb]{0,0.392,0.471}}\textcolor{white}{train} 
     &{\cellcolor[rgb]{0,0,0.902}}\textcolor{white}{m.bike} 
     & {\cellcolor[rgb]{0.467,0.043,0.125}}\textcolor{white}{bike}
     &{\cellcolor[rgb]{0,0,0}}\textcolor{white}{n/a.}
\end{tabular}
}%

\caption{\textbf{Example predictions on Cityscapes} (186 labels) showing the improved semantic understanding of \ours. In particular, SemiVL better distinguishes classes with similar visual appearance such as different vehicles (car, truck, bus, and train), different structures (fence, wall, and building), and ground types (road and sidewalk).}
\label{fig:suppl_examples_cityscapes}
\end{figure*}

We present an extended qualitative comparison of \ours\ with UniMatch~\cite{yang2023revisiting} in Fig.~\ref{fig:suppl_examples_voc} for Pascal VOC with 92 labels, in Fig.~\ref{fig:suppl_examples_coco} for COCO with 232 labels, in Fig.~\ref{fig:suppl_examples_ade} for ADE20K with 158 labels, and in Fig.~\ref{fig:suppl_examples_cityscapes} for Cityscapes with 186 labels. They consistently show that \ours\ better distinguishes classes with a similar visual appearance. For Pascal VOC, Fig.~\ref{fig:suppl_examples_voc} shows that \ours\ better distinguishes different animals (sheep, cat, dog, cow, and horse), furniture (chair, dining table, and sofa), and vehicles (bus, car, boat, airplane, and train). It also improves the distinction of foreground classes vs. the background class. For COCO, Fig.~\ref{fig:suppl_examples_coco} shows that \ours\ better recognizes different animals (bear, sheep, cat, and cow), food (cake, donut, sandwich, and apple), furniture (couch, chair, book, and vase), and sports gear (skateboard, skis, kite, umbrella, and surfboard). For ADE20K, Fig.~\ref{fig:suppl_examples_ade} shows that \ours\ better differentiates structures (tower, bridge, building, house, skyscraper, column, and wall), furniture (cabinet, door, chair, seat, table, sofa, and pool table), and ground types (rock, mountain, water, fountain, floor, rug, grass, and river). And for Cityscapes, Fig.~\ref{fig:suppl_examples_cityscapes} shows that \ours\ better distinguishes different vehicles (car, truck, bus, and train), different structures (fence, wall, and building), and ground types (road and sidewalk). To visualize them in a grid, all examples are the center crop of the full-sized predictions.

\begin{figure*}
\footnotesize
\centering
\textbf{Vision-Language Pre-Training (VL Pretr.) Improvements}\\
\begin{tabularx}{.82\linewidth}{*{8}{Y}}
Image & G. Truth & UniMatch$_\text{ViT}$ & +VL Pretr. & +SFT & +Lang.Dec. & +Guid. & +Cls.Def.\ \\
\end{tabularx}
\includegraphics[width=.82\linewidth]{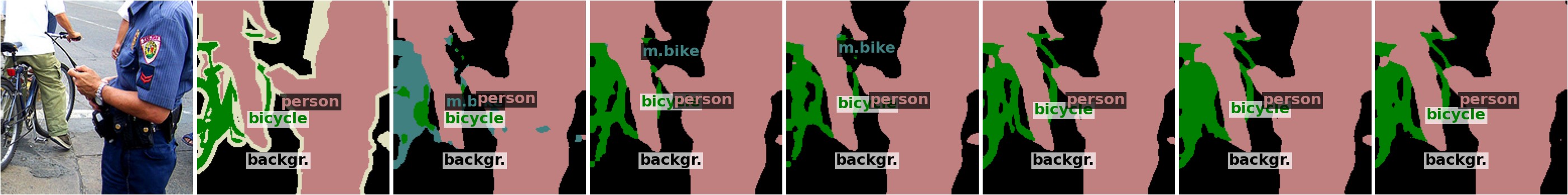}
\includegraphics[width=.82\linewidth]{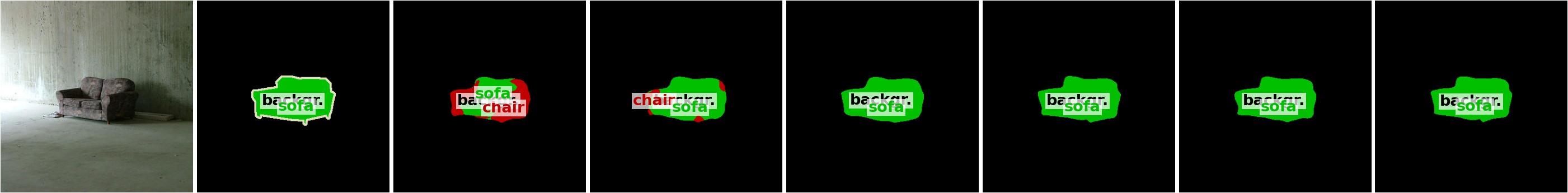}
\\\textbf{Spatial fine-tuning (SFT) Improvements}\\
\includegraphics[width=.82\linewidth]{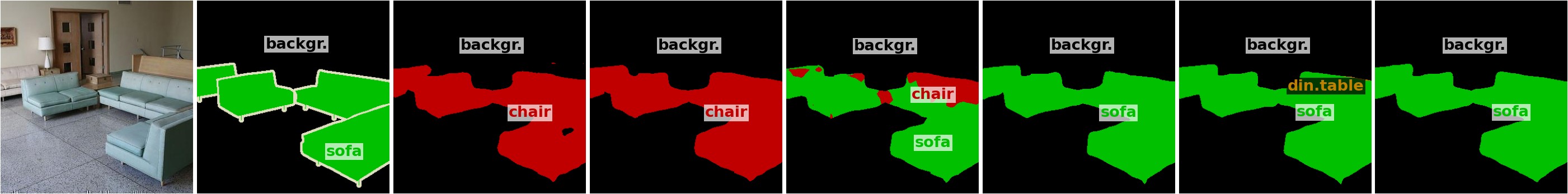}
\includegraphics[width=.82\linewidth]{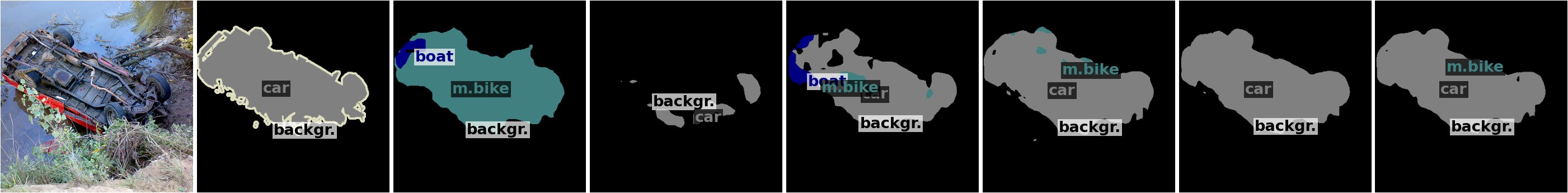}
\\\textbf{Language-Guided Decoder (Lang.Dec.) Improvements}\\
\begin{tabularx}{.82\linewidth}{*{8}{Y}}
Image & G. Truth & UniMatch$_\text{ViT}$ & +VL Pretr. & +SFT & +Lang.Dec. & +Guid. & +Cls.Def.\ \\
\end{tabularx}
\includegraphics[width=.82\linewidth]{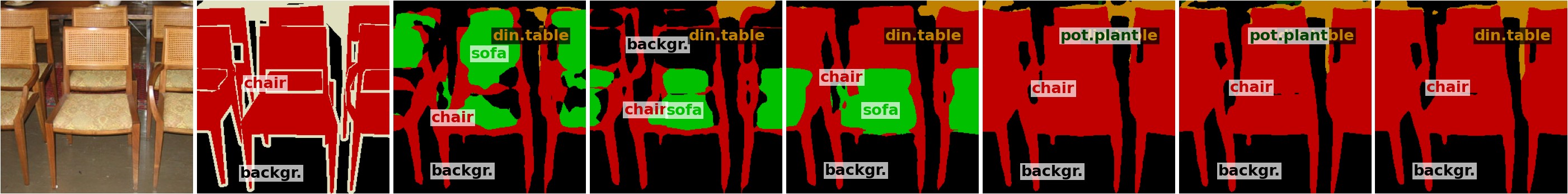}
\includegraphics[width=.82\linewidth]{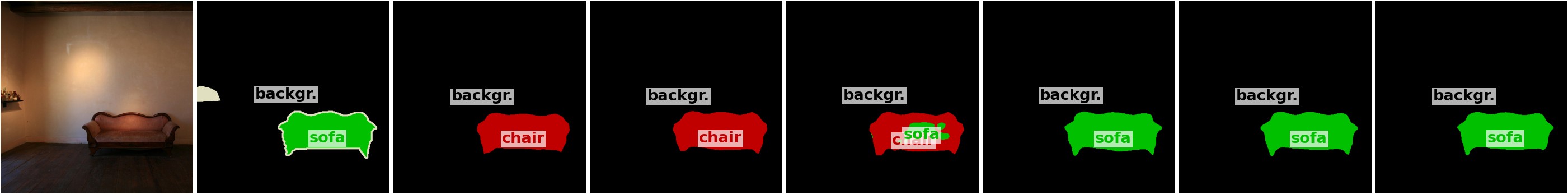}
\\\textbf{Dense CLIP Guidance (Guid.) Improvements}\\
\includegraphics[width=.82\linewidth]{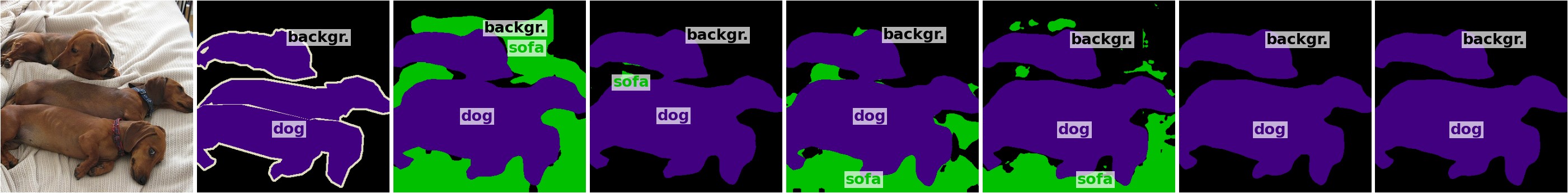}
\includegraphics[width=.82\linewidth]{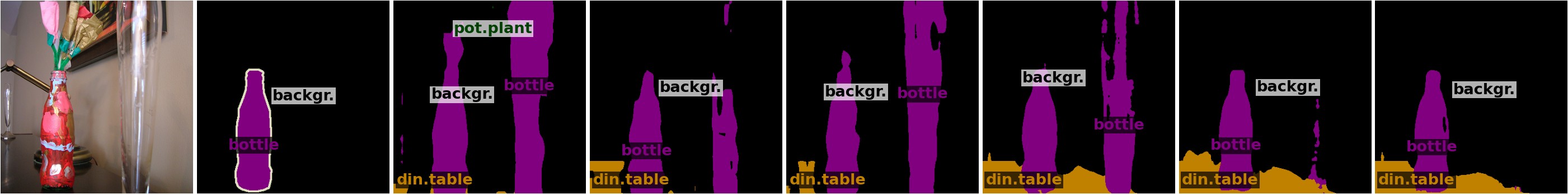}
\\\textbf{Class Definitions (Cls.Def.) Improvements}\\
\begin{tabularx}{.82\linewidth}{*{8}{Y}}
Image & G. Truth & UniMatch$_\text{ViT}$ & +VL Pretr. & +SFT & +Lang.Dec. & +Guid. & +Cls.Def.\ \\
\end{tabularx}
\includegraphics[width=.82\linewidth]{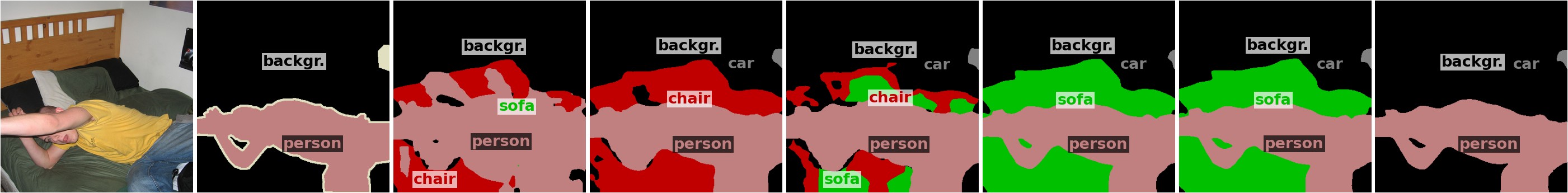}
\includegraphics[width=.82\linewidth]{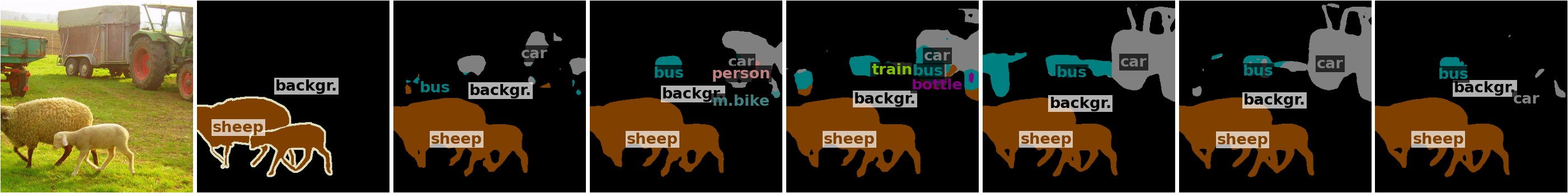}

\caption{\textbf{Example predictions of the ablation study} on VOC (92 labels). Each of SemiVL's components enhances its ability to distinguish semantically similar classes such as bicycle/motorbike, sofa/chair, car/motorbike, dining table/chair, bottle/vase(background), sofa/bed(background), and car/tractor(background).}
\label{fig:suppl_examples_ablation}
\end{figure*}

For deeper insights on the behavior of SemiVL's components, we visualize qualitative examples from the ablation study on VOC with 92 labels. Fig.~\ref{fig:suppl_examples_ablation} shows that each of SemiVL's components enhances its ability to distinguish semantically similar classes such as bicycle/motorbike, sofa/chair, car/motorbike, dining table/chair, bottle/vase(background), sofa/bed(background), and car/tractor(background). Specifically, VL pre-training and spatial fine-tuning exploit the rich semantic priors of CLIP, better recognizing the bicycle, sofa, and car. The language-guided decoder improves the spatial reasoning between object parts, enhancing the recognition of the chair/sofa seats from their context, i.e., the correctly classified chair legs and sofa armrests. The dense CLIP guidance counters drift issues appearing with spatial fine-tuning such as for the sofa and bottle. Finally, the class definitions help in the last two challenging rows as bed and tractor were defined as background in the class definitions (see Tab.~\ref{tab:class_def_voc_concepts}).

\end{document}